\documentclass[10pt,twocolumn,letterpaper]{article}

\usepackage{cvpr}

\usepackage{times}
\usepackage{epsfig}
\usepackage{graphicx}
\usepackage{amsmath}
\usepackage{amssymb}
\usepackage{enumitem}
\usepackage{verbatim}            
\usepackage{booktabs}
\usepackage{relsize}
\usepackage{scrextend}           
\usepackage{multirow}
\usepackage{array}               
\usepackage{makecell}            
\usepackage{subcaption}          
\usepackage{mathtools}           
\usepackage{siunitx}             
\usepackage{wrapfig}
\allowdisplaybreaks              
\usepackage{float}
\usepackage{cite}
\usepackage{caption}
\graphicspath{{fig/}}
\usepackage{bm}
\usepackage{paralist}

\newcommand{\example}{{e.g.}}
\newcommand{\thatIs}{{i.e.}}

\newcommand{\bmu}{\boldsymbol{\mu}_{ij}}
\newcommand{\bmux}{\mu_{ijx}}
\newcommand{\bmuy}{\mu_{ijy}}
\newcommand{\bmuF}{\boldsymbol{\mu}_{Kj}}
\newcommand{\bmuFx}{{\mu_{Kj}}_{x}}

\newcommand{\ground}{\mathbf{p}_j}
\newcommand{\groundx}{{p}_{jx}}

\newcommand{\bSigmaNoSub}{\boldsymbol{\Sigma}}
\newcommand{\bSigma}{\boldsymbol{\Sigma}_{ij}}
\newcommand{\bSigmaF}{\boldsymbol{\Sigma}_{Kj}}
\newcommand{\bSigmaFxx}{{\Sigma_{Kj}}_{xx}}
\newcommand{\bSigmaFyy}{{\Sigma_{Kj}}_{yy}}
\newcommand{\bSigmaFxy}{{\Sigma_{Kj}}_{xy}}
\newcommand{\bSigmaInv}{\bm{\Sigma}_{ij}^{-1}}
\newcommand{\bh}{\bm{H}_{ij}}
\newcommand{\bL}{\bm{L}_{ij}}
\newcommand{\bLT}{\bm{L}_{ij}^{T}}

\newcommand{\loss}{\mathcal{L}}
\newcommand{\visgd}{v_{j}}
\newcommand{\vispred}{\widehat{v}_{ij}}

\newcommand{\ltwodistance}{\ell_2}

\newcommand{\threehundredW}{$300$-W}
\newcommand{\threehundredLP}{$300$W-LP-$2$D}
\newcommand{\menpoTwoD}{Menpo $2$D}
\newcommand{\cofwSixtyEight}{COFW-$68$}
\newcommand{\aflwNineteen}{AFLW-$19$}
\newcommand{\wflwNinetyEight}{WFLW-$98$}
\newcommand{\ourdataset}{MERL-RAV}

\newcommand{\threehundredWHeading}{300-W}
\newcommand{\aflwNineteenHeading}{AFLW-19}
\newcommand{\ourdatasetHeading}{MERL-RAV}

\newcommand{\nmebox}{$\mathrm{NME_{\text{box}}}$}
\newcommand{\nmeboxvis}{$\mathrm{NME^{\text{vis}}_{\text{box}}}$}
\newcommand{\nmediag}{$\mathrm{NME_{\text{diag}}}$}
\newcommand{\nmeocular}{$\mathrm{NME_{\text{inter-ocular}}}$}
\newcommand{\aucbox}{$\mathrm{AUC}^{7}_{\text{box}}$}
\newcommand{\aucocular}{$\mathrm{AUC}^{10}_{\text{inter-ocular}}$}
\newcommand{\frocular}{$\mathrm{FR}^{10}_{\text{inter-ocular}}$}

\newcommand{\zerovec}{\bm{0}}

\newcommand{\first}[1]{$\textcolor{red}{\mathbf{#1}}$}
\newcommand{\second}[1]{$\textcolor{blue}{\mathbf{#1}}$}
\newcommand{\firstkey}[1]{\textcolor{red}{\textbf{#1}}}
\newcommand{\secondkey}[1]{\textcolor{blue}{\textbf{#1}}}

\newcommand{\z}{{\bf z}}

\newcommand{\I}{{\bf I}}

\renewcommand{\P}{{\mathcal{P}}}

\newcommand{\ben}{\begin{enumerate}}
\newcommand{\een}{\end{enumerate}}

\newcommand{\EE}{\mathbb{E}}

\newcommand{\cmt}[1]{}

\usepackage{pifont}
\newcommand{\cmark}{\checkmark}
\newcommand{\xmark}{\ding{53}}%

\usepackage{xcolor}

\newcommand{\myTopRule}{\Xhline{2\arrayrulewidth}}
\newcolumntype{t}{!{\vrule width 2\arrayrulewidth}}
\newcolumntype{m}{!{\vrule width 2.5\arrayrulewidth}}
\newcommand\scalemath[2]{\scalebox{#1}{\mbox{\ensuremath{\displaystyle #2}}}}

\usepackage{tikz} 
\usetikzlibrary{arrows}
\usetikzlibrary{arrows.meta}
\usetikzlibrary{shapes.arrows}
\usetikzlibrary{fadings, shadows}
\usetikzlibrary{decorations.text}
\definecolor{my_green}{rgb}{0.0, 0.9, 0.24}
\definecolor{my_green_2}{rgb}{0.0, 0.4, 0.0}
\definecolor{my_violet}{rgb}{0.79, 0.40, 1}
\definecolor{my_blue}{rgb}{0.2, 0.6, 1}
\definecolor{my_yellow}{rgb}{0.9, 0.8, 0.54}
\definecolor{my_yellow_2}{rgb}{1, 0.75, 0.}
\definecolor{my_red}{rgb}{1,0,0}
\definecolor{my_purple}{rgb}{0.27,0.8, 0.8}

\usepackage[pagebackref=true,breaklinks=true,letterpaper=true,colorlinks,bookmarks=false]{hyperref}

\cvprfinalcopy

\def\cvprPaperID{7427} 

\ifcvprfinal\fi
\begin{document}

\setlength{\abovedisplayskip}{0.075cm}%
\setlength{\belowdisplayskip}{0.075cm}%
\setlength{\abovedisplayshortskip}{0.075cm}%
\setlength{\belowdisplayshortskip}{0.075cm}%

\title{LUVLi Face Alignment: Estimating Landmarks'\\
Location, Uncertainty, and Visibility Likelihood}

\author{Abhinav Kumar\thanks{Equal Contributions} $^{,1}$, Tim K. Marks\footnotemark[1]  $^{,2}$, Wenxuan Mou\footnotemark[1] $^{,3}$,\\ Ye Wang$^{2}$, Michael Jones$^{2}$, Anoop Cherian$^{2}$, Toshiaki Koike-Akino$^{2}$, Xiaoming Liu$^{4}$, Chen Feng$^{5}$\\
{\tt\small abhinav3663@gmail.com, tmarks@merl.com, wenxuanmou@gmail.com, }\\
{\tt\small $[$ywang, mjones, cherian, koike$]$@merl.com, liuxm@cse.msu.edu, cfeng@nyu.edu }\\
{\footnotesize $^{1}$University of Utah, $^{2}$Mitsubishi Electric Research Labs (MERL),  $^{3}$University of Manchester, $^{4}$Michigan State University, $^{5}$New York University}
}

\maketitle
\thispagestyle{empty}

\begin{abstract}
Modern face alignment methods have become quite accurate at predicting the locations of facial landmarks, but they do not typically estimate the uncertainty of their predicted locations nor predict whether landmarks are visible. In this paper, we present a novel framework for jointly predicting landmark locations, associated uncertainties of these predicted locations, and landmark visibilities. We model these as mixed random variables and estimate them using a deep network trained with our proposed Location, Uncertainty, and Visibility Likelihood (LUVLi) loss. In addition, we release an entirely new labeling of a large face alignment dataset with over 19,000 face images in a full range of head poses. Each face is manually labeled with the ground-truth locations of 68 landmarks, with the additional information of whether each landmark is unoccluded, self-occluded (due to extreme head poses), or externally occluded. Not only does our joint estimation yield accurate estimates of the uncertainty of predicted landmark locations, but it also yields state-of-the-art estimates for the landmark locations themselves on multiple standard face alignment datasets. Our method's estimates of the uncertainty of predicted landmark locations could be used to automatically identify input images on which face alignment fails, which can be critical for downstream tasks.
\end{abstract}

\vspace{-0.5cm}
\section{Introduction}
    Modern methods for face alignment (facial landmark localization) perform quite well most of the time, but all of them fail some percentage of the time. Unfortunately, almost all of the state-of-the-art (SOTA) methods simply output predicted landmark locations, with no assessment of whether (or how much) downstream tasks should {\it trust} these landmark locations. This is concerning, as face alignment is a key pre-processing step in numerous safety-critical applications, including advanced driver assistance systems (ADAS), driver monitoring, and remote measurement of vital signs~\cite{nowara2018sparseppg}. As deep neural networks are notorious for producing overconfident predictions~\cite{guo2017calibration}, similar concerns have been raised for other neural network technologies~\cite{le2018uncertainty}, and they become even more acute in the era of adversarial machine learning where adversarial images may pose a great threat to a system~\cite{chen2019face}. However, previous work in face alignment (and landmark localization in general) has largely ignored the area of uncertainty estimation.
    
    To address this need, we propose a method to jointly estimate facial landmark locations and a parametric probability distribution representing the uncertainty of each estimated location. Our model also jointly estimates the visibility of landmarks, which predicts whether each landmark is occluded due to extreme head pose.

    \begin{figure}[!tb]
    	\centering
    	\begin{subfigure}{0.3\linewidth}
    		\includegraphics[height=\linewidth]{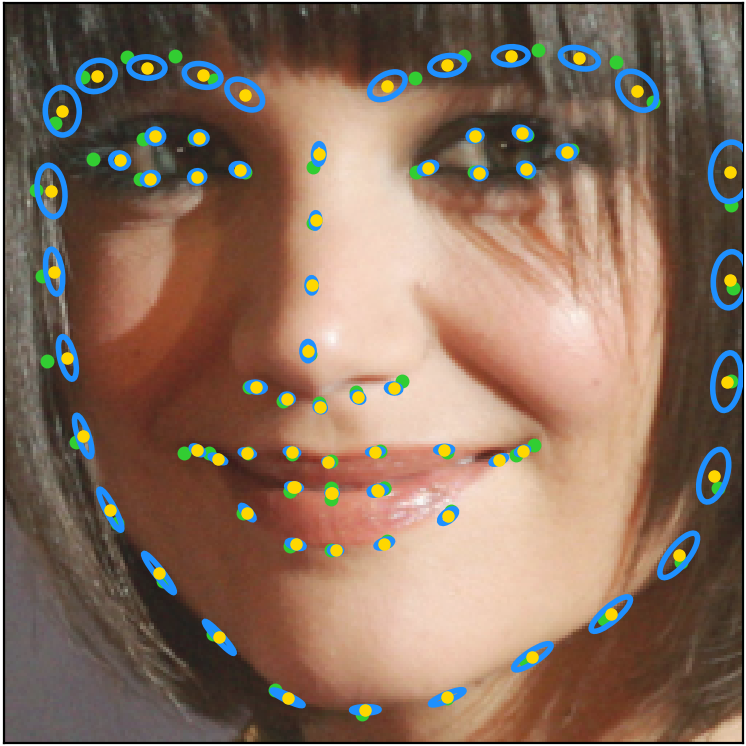}
    	\end{subfigure}
    	\hfill
    	\begin{subfigure}{0.3\linewidth}
    		\includegraphics[height=\linewidth]{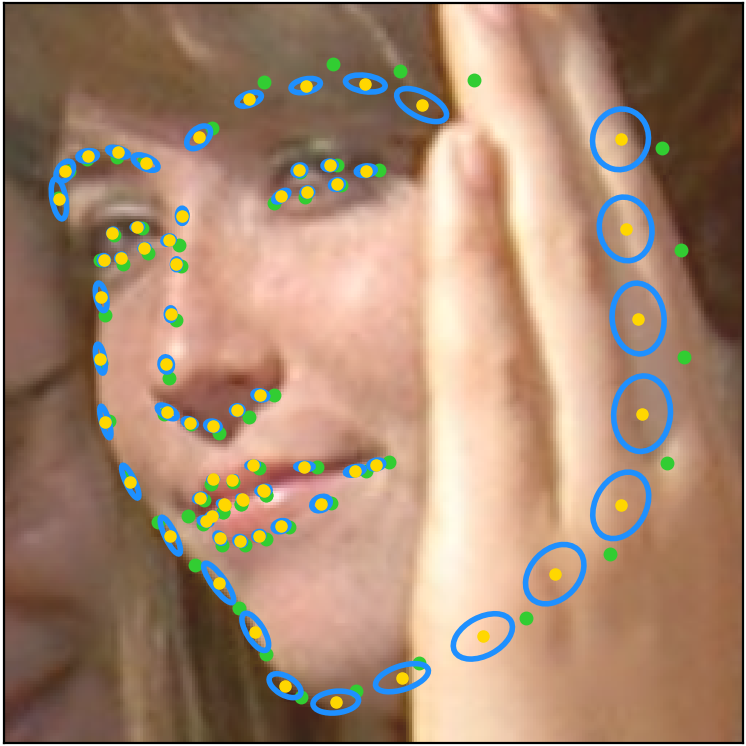}
    	\end{subfigure}%
    	\hfill
    	\begin{subfigure}{0.3\linewidth}
    		\includegraphics[height=\linewidth]{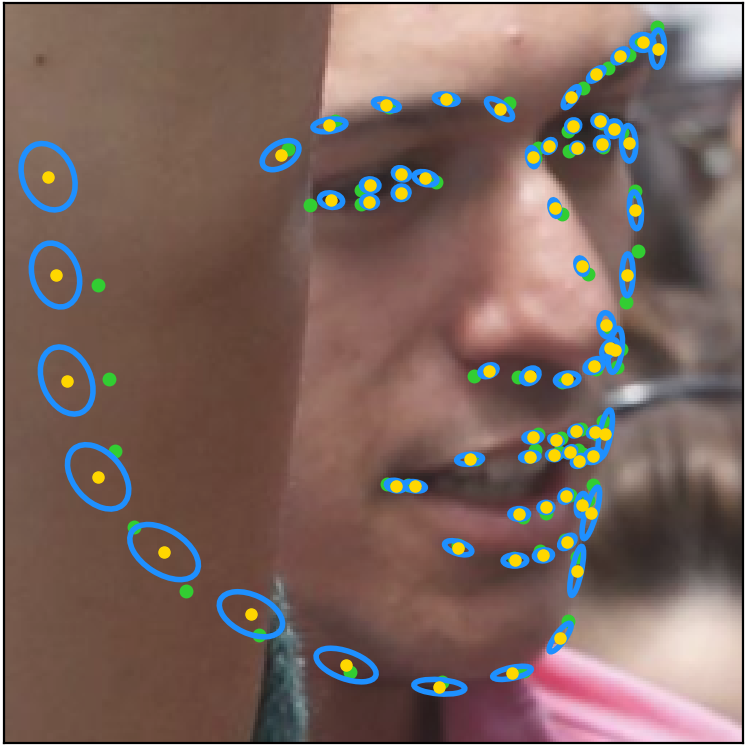}
    	\end{subfigure}%
    	\caption{Results of our joint face alignment and uncertainty estimation on three test images. Ground-truth (green) and predicted (yellow) landmark locations are shown. The estimated uncertainty of the predicted location of each landmark is shown in blue (Error ellipse for Mahalanobis distance 1). Landmarks that are occluded (\example, by the hand in center image) tend to have larger uncertainty. 
    	}
    	\vspace{-0.25cm}
    	\label{fig:fig1}
    \end{figure}
    
    We find that the choice of methods for calculating mean and covariance is crucial. Landmark locations are best obtained using heatmaps, rather than by direct regression. To estimate landmark locations in a differentiable manner using heatmaps, we do not select the location of the maximum (argmax) of each landmark's heatmap, but instead propose to use the spatial mean of the positive elements of each heatmap. Unlike landmark locations, uncertainty distribution parameters are best obtained by direct regression rather than from heatmaps. To estimate the uncertainty of the predicted locations, we add a Cholesky Estimator Network (CEN) branch to estimate the covariance matrix of a multivariate Gaussian or Laplacian probability distribution. To estimate visibility of each landmark, we add a Visibility Estimator Network (VEN). We combine these estimates using a joint loss function that we call the Location, Uncertainty and Visibility Likelihood (LUVLi) loss. Our primary goal in designing this model was to estimate uncertainty in landmark localization. In the process, not only does our method yields accurate uncertainty estimation, but it also produces SOTA landmark localization results on several face alignment datasets.

    Uncertainty can be broadly classified into two categories~\cite{kendall2017uncertainties}: \emph{epistemic} uncertainty is related to a lack of knowledge about the model that generated the observed data, and \emph{aleatoric} uncertainty
    is related to the noise inherent in the observations, \example, sensor or labelling noise. The ground-truth landmark locations marked on an image by human labelers would vary across multiple labelings of an image by different human labelers (or even by the same human labeler). Furthermore, this variation will itself vary across different images and landmarks (\example, it will vary more for occluded landmarks and poorly lit images). The goal of our method is to estimate this aleatoric uncertainty.
    
    The fact that each image only has one ground-truth labeled location per landmark makes estimating this uncertainty distribution difficult, but not impossible. To do so, we use a parametric model for the uncertainty distribution. We train a neural network to estimate the parameters of the model for each landmark of each input face image so as to maximize the likelihood under the model of the ground-truth location of that landmark (summed across all landmarks of all training faces).

    The main contributions of this work are as follows:
    \begin{compactitem}
        \item This is the first work to introduce the concept of parametric uncertainty estimation for face alignment.
        \item We propose an end-to-end trainable model for the joint estimation of landmark location, uncertainty, and visibility likelihood (LUVLi), modeled as a mixed random variable.
        \item We compare our model using multivariate Gaussian and multivariate Laplacian probability distributions. 
        \item Our algorithm yields accurate uncertainty estimation and state-of-the-art landmark localization results on several face alignment datasets.
        \item We are releasing a new dataset with manual labels of the locations of $68$ landmarks on over $19,\!000$ face images in a wide variety of poses, where each landmark is also labeled with one of three visibility categories.
    \end{compactitem}

\section{Related Work}
\subsection{Face Alignment}
\vspace{-0.15cm}

    Early methods for face alignment were based on Active Shape Models (ASM) and Active Appearance Models (AAM)~\cite{cootes1995active, cootes2001active,sauer2011accurate,sung2009adaptive,tzimiropolous2013aam} as well as their  variations~\cite{romdhani1999multiview,cootes2002viewAAM,Asthana2011ICCV3DposeNorm,hu2004multicamAAM, liu2008discriminative,liu2010video}. Subsequently, direct regression methods became popular due to their excellent performance. Of these, tree-based regression methods~\cite{ren2014face,kazemi2014one,cao2014face,cootes2012robust,tuzel2008learning} proved particularly fast, and the subsequent cascaded regression methods~\cite{dollar2010cascaded,xiong2013supervised,asthana2014incremental,tzimiropoulos2015POCR,tuzel2016robust} improved accuracy.
    
    Recent approaches~\cite{Zhang2014eccvCFAN, toshev2014deeppose, zhu2015face, yang2017stacked, bulat2017far, wu2018look, tang2019towards, wang2019adaptive} are all based on deep learning and can be classified into two sub-categories: direct regression~\cite{toshev2014deeppose, carreira2016human} and heatmap-based approaches. The SOTA deep methods, \example, stacked hourglass networks~\cite{yang2017stacked, bulat2017far} and densely connected U-nets (DU-Net)~\cite{tang2019towards}, use a cascade of deep networks, originally developed for human body $2$D pose estimation~\cite{newell2016stacked}. These models~\cite{newell2016stacked, bulat2017far, tang2018quantized, tang2019towards} are trained using the $\ltwodistance$ distance between the predicted heatmap for each landmark and a proxy ground-truth heatmap that is generated by placing a symmetric Gaussian distribution with small fixed variance at the ground-truth landmark location.~\cite{li2019rethinking} uses a larger variance for early hourglasses and a smaller variance for later hourglasses.~\cite{wang2019adaptive} employs different variations of MSE for different pixels of the proxy ground-truth heatmap. Recent works also infer facial boundary maps to improve alignment~\cite{wu2018look,wang2019adaptive}. In heatmap-based methods,  landmarks are estimated by the argmax of each predicted heatmap. Indirect inference through a predicted heatmap offers several advantages over direct prediction~\cite{Belagiannis17}.
    
    \textbf{Disadvantages of Heatmap-Based Approaches.} These heatmap-based methods have at least two disadvantages. First, since the goal of training is to mimic a proxy ground-truth heatmap containing a fixed symmetric Gaussian, the predicted heatmaps are poorly suited to uncertainty prediction~\cite{kdnuncertain, chen2019face}. Second, they suffer from quantization errors since the heatmap's argmax is only determined to the nearest pixel~\cite{luvizon20182d,nibali2018numerical,tai2019towards}. To achieve sub-pixel localization for body pose estimation,~\cite{luvizon20182d} replaces the argmax with a spatial mean over the softmax. Alternatively, for sub-pixel localization in videos, \cite{tai2019towards} samples two additional points adjacent to the max of the heatmap to estimate a local peak.  
    
    \textbf{Landmark Regression with Uncertainty.}
    We have only found two other methods that estimate uncertainty of landmark regression, both developed concurrently with our approach. The first method~\cite{kdnuncertain, chen2019face} estimates face alignment uncertainty using a non-parametric approach: a kernel density network obtained by convolving the heatmaps with a fixed symmetric Gaussian kernel. The second~\cite{gundavarapu2019structured} performs body pose estimation with uncertainty using direct regression method (no heatmaps) to directly predict the mean and precision matrix of a Gaussian distribution.

    \subsection{Uncertainty Estimation in Neural Networks} 
    \vspace{-0.2cm}\
    Uncertainty estimation broadly uses two types of approaches~\cite{le2018uncertainty}: sampling-based and sampling-free. Sampling-based methods include Bayesian neural networks~\cite{shridhar2019comprehensive}, Monte Carlo dropout~\cite{gal2016dropout}, and bootstrap ensembles~\cite{lakshminarayanan2017simple}. They rely on multiple evaluations of the input to estimate uncertainty~\cite{le2018uncertainty}, and bootstrap ensembles also need to store several sets of weights~\cite{ilg2018uncertainty}. Thus, sampling-based methods work for small $1$D regression problems but might not be feasible for higher-dimensional problems~\cite{ilg2018uncertainty}. 

    Sampling-free methods produce two outputs, one for the estimate and the other for the uncertainty, and optimize Gaussian log-likelihood (GLL) instead of classification and regression losses~\cite{kendall2017uncertainties, lakshminarayanan2017simple,le2018uncertainty}.~\cite{lakshminarayanan2017simple} combines the benefits of sampling-free and sampling-based methods. 
    
    Recent object detection methods have used uncertainty estimation~\cite{le2018uncertainty, jiang2018acquisition, levi2019evaluating, he2019bounding, miller2019benchmarking, harakeh2019bayesod,atoum2017monocular}. Sampling-free methods~\cite{le2018uncertainty, levi2019evaluating, he2019bounding} jointly estimate the four parameters of the bounding box using Gaussian log-likelihood~\cite{levi2019evaluating}, Laplacian log-likelihood~\cite{le2018uncertainty}, or both~\cite{he2019bounding}. However, these methods assume the four parameters of the bounding box are independent (assume a diagonal covariance matrix). Sampling-based approaches use Monte Carlo dropout~\cite{miller2019benchmarking} and network ensembles~\cite{lakshminarayanan2017simple} for object detection.
    Uncertainty estimation has also been applied to pixelwise depth regression~\cite{kendall2017uncertainties}, optical flow~\cite{ilg2018uncertainty}, pedestrian detection~\cite{neumann2018relaxed,bhattacharyya2018long,bertoni2019monoloco} and $3$D vehicle detection~\cite{feng2018towards}.

\section{Proposed Method}\label{sec:method}
\begin{figure}[!tb]
        \centering
        \input{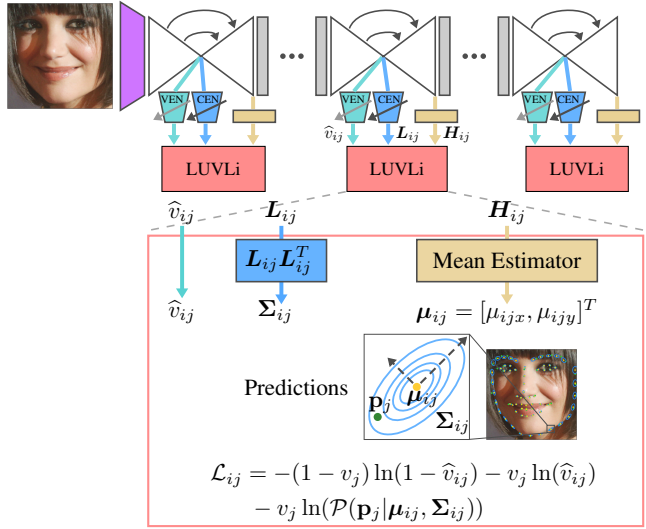}
        \vspace{-0.4cm}
        \caption{Overview of our LUVLi method. From each U-net of a DU-Net, we append a shared Cholesky Estimator Network (CEN) and Visibility Estimator Network (VEN) to the bottleneck layer and apply a mean estimator to the heatmap. The figure shows the joint estimation of location, uncertainty, and visibility of the landmarks performed for each U-net $i$ and landmark $j$. The landmark has ground-truth (labeled) location $\ground$ and visibility $\visgd \in \{0,1\}$.}
        \label{fig:overview}
        \vspace{-0.3cm}
    \end{figure}

\vspace{-0.15cm}
    Figure~\ref{fig:overview} shows an overview of our LUVLi Face Alignment. The input RGB face image is passed through a \mbox{DU-Net}~\cite{tang2019towards} architecture, to which we add three additional components branching from each U-net. The first new component is a \emph{mean estimator}, which computes the estimated location of each landmark as the weighted spatial mean of the positive elements of the corresponding heatmap. The second and the third new component, the \emph{Cholesky Estimator Network} (CEN) and the \emph{Visibility Estimator Network} (VEN), emerge from the bottleneck layer of each U-net. CEN and VEN weights are shared across all U-nets. The CEN estimates the Cholesky coefficients of the covariance matrix for each landmark location. The VEN estimates the probability of visibility of each landmark in the image, $1$ meaning visible and $0$ meaning not visible. For each U-net $i$ and each landmark $j$, the landmark's location estimate $\bmu$, estimated covariance matrix $\bSigma$, and estimated visibility $\vispred$ are tied together by the LUVLi loss function $\loss_{ij}$, which enables end-to-end optimization of the entire framework.
    
    Rather than the argmax of the heatmap, we choose a mean estimator for the heatmap that is differentiable and enables sub-pixel accuracy: the weighted spatial mean of the heatmap's positive elements. Unlike the non-parametric model of~\cite{kdnuncertain, chen2019face}, our uncertainty prediction method is parametric: we directly estimate the parameters of a single multivariate Laplacian or Gaussian distribution. Furthermore, our method does not constrain the Laplacian or Gaussian covariance matrix to be diagonal.

    \subsection{Mean Estimator}\label{sec:soft-argmax}
    \vspace{-0.15cm}
    Let $\bh(x,y)$ denote the value at pixel location $(x,y)$ of the $j{\text{th}}$ landmark's heatmap from the $i{\text{th}}$ U-net. 
    The landmark's location estimate $\bmu = [\bmux, \bmuy]^T$ is given by first post-processing the pixels of the heatmap $\bh$ with a function $\sigma$, then taking the weighted spatial mean of the result (See~\eqref{eq:mean} in the supplementary material). We considered three different functions for $\sigma$: the ReLU function (eliminates the negative values), the softmax function (makes the mean estimator a soft-argmax of the heatmap~\cite{chapelle2010gradient, yi2016lift, luvizon20182d, dong2018supervision}), and a temperature-controlled softmax function (which, depending on the temperature setting, provides a continuum of softmax functions that range from a ``hard'' argmax to the uniform distribution). The ablation studies (Section~\ref{sec:results_ablation}) show that choosing $\sigma$ to be the ReLU function yields the simplest and best mean estimator.

    \subsection{LUVLi Loss} \label{sec:loss}
    \vspace{-0.15cm}
        Occluded landmarks, \example, landmarks on the far side of a profile-pose face, are common in real data. To explicitly represent visibility, we model the probability distributions of landmark locations using mixed random variables. For each landmark $j$ in an image, we denote the ground-truth (labeled) visibility by the binary variable $\visgd \in \{0, 1\}$, where $1$ denotes {\em visible}, and the ground-truth location by $\ground$. By convention, if the landmark is not visible ($\visgd = 0$), then $\ground = \emptyset$, a special symbol indicating non-existence. Together, these variables are distributed according to an unknown distribution $p(\visgd, \ground)$. The marginal Bernoulli distribution $p(\visgd)$ captures the probability of visibility, $p(\ground|\visgd\!=\!1)$ denotes the distribution of the landmark location when it is visible, and $p(\ground|\visgd\!=\!0)\!=\!\mathbf{1}_{\emptyset}(\ground)$, where $\mathbf{1}_{\emptyset}$ denotes the PMF that assigns probability one to the symbol $\emptyset$.
    
         After each U-net $i$, we estimate the joint distribution of the visibility $v$ and location $\z$ of each landmark $j$ via
        \begin{equation}
            q(v, \z) = q_v(v) q_z(\z|v),
        \end{equation}
        where $q_v(v)$ is a Bernoulli distribution with
        \begin{equation}
            q_v(v=1) = \vispred, \qquad 
            q_v(v=0) = 1 - \vispred,
        \end{equation}
        where $\vispred$ is the predicted probability of visibility, and
        \begin{align}
            q_z(\z | v=1) &= \P(\z | \bmu,\bSigma),\\
            q_z(\z | v=0) &=    \emptyset,
        \end{align}
        where $\P$ denotes the likelihood of the landmark being at location $\z$ given the estimated mean $\bmu$ and covariance $\bSigma$.
        
        The LUVLi loss is the negative log-likelihood with respect to $q(v, \z)$, as given by
        \begin{align}
            \loss_{ij} =& - \ln q(\visgd, \ground) \nonumber \\
            =& - \ln q_v(\visgd) - \ln q_z(\ground | \visgd) \nonumber \\
            =& -(1 - \visgd) \ln (1 - \vispred) - \visgd \ln(\vispred) \nonumber \\
            &- \visgd \ln\bigl(\P(\ground |\bmu, \bSigma)\bigr), \label{eq:vloss}
        \end{align}
        and thus minimizing the loss is equivalent to maximum likelihood estimation.

        The terms of~\eqref{eq:vloss} are a binary cross entropy plus $\visgd$ times the negative log-likelihood of $\ground$ with respect to $\P$. This can be seen as an instance of multi-task learning~\cite{caruana1997multitask}, since we are predicting three things about each landmark: its location, uncertainty, and visibility. The first two terms on the right hand side of~\eqref{eq:vloss} can be seen as a classification loss for visibility, while the last term corresponds to a regression loss of location estimation.
        The sum of classification and regression losses is also widely used in object detection~\cite{jiao2019survey}.
        
        Minimization of negative log-likelihood also corresponds to minimizing KL-divergence, since
        \begin{align}
            &\mathbb{E} [ - \ln q(\visgd, \ground) ] = \mathbb{E} \left[ \ln \frac{p(\visgd, \ground)}{q(\visgd, \ground)} - \ln p(\visgd, \ground) \right] \\
            &= \mathrm D_{\mathrm{KL}}(p(\visgd, \ground) \| q(\visgd, \ground)) + \mathbb{E}[- \ln p(\visgd, \ground)],
        \end{align}
        where expectations are with respect to \mbox{$(\visgd, \ground) \sim p(\visgd, \ground)$}, and the entropy term $\mathbb{E}[- \ln p(\visgd, \ground)]$ is constant with respect to the estimate $q(\visgd, \ground)$.
        Further, since
        \begin{align}
            &\mathbb{E} [ - \ln q(\visgd, \ground) ] = \mathbb{E}_{\visgd \sim p(\visgd)} [ - \ln q(\visgd) ] \nonumber\\
            &\qquad + p_v \mathbb{E}_{\ground \sim p(\ground | \visgd=1)} [ - \ln \P(\ground | \bmu,\bSigma) ], \label{eq:comboKL-anal1}
        \end{align}
        where $p_v := p(\visgd = 1)$ for brevity, minimizing the negative log-likelihood (LUVLi loss) is also equivalent to minimizing the combination of KL-divergences given by
        \begin{equation}
            \!\! \! \mathrm D_{\mathrm{KL}}\!\bigl(p(\visgd) \| q(v) \!\bigr)
            \!+ p_v \mathrm D_{\mathrm{KL}}\!\bigl (p(\ground | \visgd\!\!=\!\!1) \| \P(\z | \bmu,\!\bSigma)\!\bigr)
            \label{eq:combo-KL-loss}
        \end{equation}

    \subsubsection{Models for Location Likelihood} \label{sec:loclikechoice}
    \vspace{-0.15cm}
        For the multivariate location distribution $\P$, we consider two different models: Gaussian and Laplacian.
        
        \textbf{Gaussian Likelihood.}
        The $2$D Gaussian likelihood is:
        \begin{equation}
            \P(\z|\bmu,\! \bSigma) \! = \!
            \frac{\exp \bigl(\!- \tfrac{1}{2}(\z\!-\!\!\bmu)^T\bSigmaInv(\z\!-\!\!\bmu) \bigr)}{2\pi\sqrt{\left| \bSigma \right|}}.
            \label{eq:gauss_likelihood}
        \end{equation}
        Substituting~\eqref{eq:gauss_likelihood} into~\eqref{eq:vloss}, we have
        \begin{align}
            \loss_{ij} &= -(1\!-\!\visgd)\ln(1\!-\!\vispred)\!-\visgd\ln(\vispred) +\!\visgd\!\underbrace{\frac{1}{2}\log|\bSigma|}_{T_1} \nonumber \\[-0.6cm]
            &\quad  +\visgd\!\underbrace{\frac{1}{2}(\ground\!-\bmu\!)^T\bSigmaInv(\!\ground-\!\bmu\!)}_{T_2\vspace{-0.3cm}}.
            \label{eq:gauss_likelihood_loss}
        \end{align}
        In~\eqref{eq:gauss_likelihood_loss}, $T_2$ is the squared Mahalanobis distance, while $T_1$ serves as a regularization or prior term that ensures that the Gaussian uncertainty distribution does not get too large.
        
        \textbf{Laplacian Likelihood.}
        We use a $2$D Laplacian likelihood~\cite{kotz2001asymmetric} given by:
        \begin{align}
            P(\z|\bmu,\! \bSigma) \! &= \!
            \frac{e^{-\sqrt{3(\z-\bmu)^T\bSigmaInv (\z-\bmu)}} }{\frac{2\pi}{3}\sqrt{\left| \bSigma \right|}}.
            \label{eq:lapl_simplified_likelihood}
        \end{align}
        Substituting~\eqref{eq:lapl_simplified_likelihood} in~\eqref{eq:vloss}, we have
        \begin{align}
            \loss_{ij}
            &= -(1\!-\!\visgd)\ln(1\!-\!\vispred)\!-\visgd\ln(\vispred) + \visgd\! \underbrace{\frac{1}{2}\!\log|\bSigma|}_{T_1}\!\nonumber \\[-0.6cm]
            &\quad  +\visgd\!\underbrace{\sqrt{3(\ground\!-\!\bmu)^T\bSigmaInv(\ground\!-\!\bmu)}}_{T_2}. 
            \label{eq:lapl_simplified_loss}
        \end{align}
        In~\eqref{eq:lapl_simplified_loss}, $T_2$ is a scaled Mahalanobis distance, while $T_1$ serves as a regularization or prior term that ensures that the Laplacian uncertainty distribution does not get too large. 
        
    Note that if $\bSigma$ is the identity matrix and if all landmarks are assumed to be visible, then~\eqref{eq:gauss_likelihood_loss} simply reduces to the squared $\ltwodistance$ distance, and~\eqref{eq:lapl_simplified_loss} just minimizes the $\ltwodistance$ distance.

    \subsection{Uncertainty and Visibility Estimation}\label{sec:CEN_VEN}
    \vspace{-0.15cm}
        Our proposed method uses heatmaps for estimating landmarks' locations, but not for estimating their uncertainty and visibility. We experimented with several methods for computing a covariance matrix directly from a heatmap, but none were accurate enough. We discuss this in Section~\ref{sec:300w}.
    
        \textbf{Cholesky Estimator Network (CEN).} We represent the uncertainty of each landmark location using a $2\!\times\!2$ covariance matrix $\bSigma$, which is symmetric positive definite. The three degrees of freedom of $\bSigma$ are captured by its Cholesky decomposition: a lower-triangular matrix $\bL$ such that $\bSigma\!=\!\bL\bLT$. To estimate the elements of $\bL$, we append a Cholesky Estimator Network (CEN) to the bottleneck of each U-net. The CEN is a fully connected linear layer whose input is the bottleneck of the U-net $(128\!\times\!4\!\times\!4\!=\!2,\!048$ dimensions$)$ and output is an $N_p\!\times\!3$-dimensional vector, where $N_p$ is the number of landmarks (\example, $68$). As the Cholesky decomposition $\bL$ of a covariance matrix must have positive diagonal elements, we pass the corresponding entries of the output through an ELU activation function~\cite{clevert2015fast}, to which we add a constant to ensure the output is always positive (asymptote is negative $x$-axis). 
        
        \textbf{Visibility Estimator Network (VEN).} To estimate the visibility of the landmark $v_e$, we add another fully connected linear layer whose input is the bottleneck of the U-net $(128\!\times\!4\!\times\!4\!=\!2,\!048$ dimensions$)$ and output is an  $N_p$-dimensional vector. This is passed through a sigmoid activation so the predicted visibility $\vispred$ is between $0$ and $1$. 
        
        The addition of these two fully connected layers only slightly increases the size of the original model. The loss for a single U-net is the averaged $\loss_{ij}$ across all the landmarks $j= 1,...,N_p$\,, and the total loss $\loss$ for each input image is a weighted sum of the losses of all $K$ of the U-nets:
        \begin{equation}
            \loss = \sum_{i=1}^K\lambda_i\loss_i\,, \quad \text{where} \quad \loss_i = \frac{1}{N_p}\sum_{j=1}^{N_p}\loss_{ij}\,.
            \label{eq:loss_all}
        \end{equation}
        At test time, each landmark's mean and Cholesky coefficients are derived from the $K{\text{th}}$ (final) U-net. The covariance matrix is calculated from the Cholesky coefficients.

\vspace{-0.2cm}
\section{New Dataset: \ourdatasetHeading}
\vspace{-0.15cm}
    To promote future research in face alignment with uncertainty, we now introduce a new dataset with entirely new, manual labels of over $19,\!000$ face images from the AFLW~\cite{koestinger2011annotated} dataset. In addition to landmark locations, every landmark is labeled with one of three visibility classes. We call the new dataset {\em MERL Reannotation of AFLW with Visibility} (\ourdataset).

    \textbf{Visibility Classification.}
    Each landmark of every face is classified as either \emph{unoccluded}, \emph{self-occluded}, or \emph{externally occluded}, as illustrated in Figure~\ref{fig:sample}. \emph{Unoccluded} denotes landmarks that can be seen directly in the image, with no obstructions. \emph{Self-occluded} denotes landmarks that are occluded because of extreme head pose---they are occluded by another part of the face (\example, landmarks on the far side of a profile-view face). \emph{Externally occluded} denotes landmarks that are occluded by hair or an intervening object such as a cap, hand, microphone, or goggles. Human labelers are generally very bad at localizing self-occluded landmarks, so we do not provide ground-truth locations for these. We do provide ground-truth (labeled) locations for both unoccluded and externally occluded landmarks.

    \textbf{Relationship to Visibility in LUVLi.} 
    In Section~\ref{sec:method}, visible landmarks ($\visgd=1$) are landmarks for which ground-truth location information is available, while invisible landmarks ($\visgd=0$) are landmarks for which no ground-truth location information is available ($\ground = \emptyset$). Thus, invisible ($\visgd=0$) in the model is equivalent to the \emph{self-occluded} landmarks in our dataset. In contrast, both \emph{unoccluded} and \emph{externally occluded} landmarks are considered visible ($\visgd=1$) in our model. We choose this because human labelers are generally good at estimating the locations of externally occluded landmarks but poor at estimating the locations of self-occluded landmarks.

    \textbf{Existing Datasets.}
    The most commonly used publicly available datasets for evaluation of $2$D face alignment are summarized in Table~\ref{tab:dataset}. The \threehundredW~dataset~\cite{sagonas2013300, sagonas2016300, sagonas2013semi} uses a $68$-landmark system that was originally used for Multi-PIE~\cite{gross2010multi}. \menpoTwoD~\cite{zafeiriou2017menpo, trigeorgis2016mnemonic, deng2019menpo} makes a hard distinction (denoted F/P) between nearly frontal faces (F) and profile faces (P). \menpoTwoD~uses the same landmarks as \threehundredW~for frontal faces, but for profile faces it uses a different set of $39$ landmarks that do not all correspond to the $68$ landmarks in the frontal images. \threehundredLP~\cite{zhu2016face,bulat2017far} is a synthetic dataset created by automatically reposing \threehundredW~faces, so it has a large number of labels, but they are noisy. The $3$D model locations of self-occluded landmarks are projected onto the visible part of the face as if the face were transparent (denoted by T). The WFLW~\cite{wu2018look} and AFLW-$68$~\cite{qian2019aggregation} datasets do not identify which landmarks are self-occluded, but instead label self-occluded landmarks as if they were located on the visible boundary of the noseless face. 
    
    \begin{table}[!t]
        \caption{Overview of face alignment datasets. [Key: \\Self Occ= Self-Occlusions, Ext Occ= External Occlusions]}
        \label{tab:dataset}
        \setlength{\tabcolsep}{0.075cm}
        \centering
        \footnotesize
        \begin{tabular}{lcccccc}
            \myTopRule
            \textbf{Dataset} & \textbf{\#train}  & \textbf{\#test} & \textbf{\#marks}  & \textbf{Profile} & \textbf{Self}  & \textbf{Ext}\\
            &&&  & \textbf{Images} & \textbf{Occ} & \textbf{Occ} \\
            \myTopRule
            COFW~\cite{burgos2013robust}                                                  & $1,\!345$  & $507$              & $29$          & \xmark & \xmark & \cmark \\
            \cofwSixtyEight~\cite{ghiasi2015occlusion}                                    & -          & $507$              & $68$          & \xmark & \xmark & \cmark \\
            \threehundredW~\cite{sagonas2013300, sagonas2016300, sagonas2013semi}         & $3,\!837$  & $600$              & $68$          & \xmark & \xmark & \xmark \\
            \menpoTwoD~\cite{zafeiriou2017menpo, trigeorgis2016mnemonic, deng2019menpo}   & $7,\!564$  & $7,\!281$          & $68/39$       & \cmark & F/P    & \xmark \\ 
            \threehundredLP~\cite{zhu2016face}                                            & $61,\!225$ & -                  & $68$          & \cmark & T      & \xmark \\
            WFLW~\cite{wu2018look}                                                        & $7,\!500$  & $2,\!500$          & $\mathbf{98}$ & \cmark & \xmark & \xmark \\
            AFLW~\cite{koestinger2011annotated}                                           & $20,\!000$ & $\mathbf{4,\!386}$ & $21$          & \cmark & \cmark & \xmark \\
            \aflwNineteen~\cite{zhu2016unconstrained}                                     & $20,\!000$ & $\mathbf{4,\!386}$ & $19$          & \cmark & \xmark & \xmark \\
            AFLW-$68$~\cite{qian2019aggregation}                                          & $20,\!000$ & $\mathbf{4,\!386}$ & $68$          & \cmark & \xmark & \xmark \\
            \bottomrule
            \ourdataset~(Ours)                                                            & $15,\!449$ & $3,\!865$          & $68$          & \cmark & \cmark & \cmark \\
            \myTopRule
        \end{tabular}
        \vspace{-0.7cm}
    \end{table}

    \textbf{Differences from Existing Datasets.}
    Our \ourdataset~dataset is the only one that labels every landmark using both types of occlusion (self-occlusion and external occlusion). Only one other dataset, AFLW, indicates which individual landmarks are self-occluded, but it has far fewer landmarks and does not label external occlusions. COFW and \cofwSixtyEight~indicate which landmarks are externally occluded but do not have self-occlusions. \menpoTwoD~categorizes faces as frontal or profile, but landmarks of the two classes are incompatible. Unlike \menpoTwoD, our dataset smoothly transitions from frontal to profile, with gradually more and more landmarks labeled as self-occluded.  
    
    Our dataset uses the widely adopted $68$ landmarks used by \threehundredW, to allow for evaluation and cross-dataset comparison.  Since it uses images from AFLW, our dataset has pose variation up to $\pm 120^\circ$ yaw and $\pm 90^\circ$  pitch. Focusing on yaw, we group the images into five pose classes: frontal, left and right half-profile, and left and right profile. The train/test split is in the ratio of $4\!:\!1$. Table~\ref{tab:dataset_statistics} provides the statistics of our \ourdataset~dataset. A sample image from the dataset is shown in Figure~\ref{fig:sample}. In the figure, unoccluded landmarks are green, externally occluded landmarks are red, and self-occluded landmarks are indicated by black circles in the face schematic on the right.
  
    \begin{table}[!tb]
        \begin{minipage}[b]{0.5\linewidth}
            \footnotesize
            \setlength{\tabcolsep}{0.075cm}
            \begin{tabular}{cccc}
                \myTopRule
                \textbf{Pose} & \textbf{Side} & \textbf{\#Train} & \textbf{\#Test}\\
                \myTopRule
                Frontal       & -             & $8,\!778$           & $2,\!195$\\
                \hline
                Half-         & Left half     & $1,\!180$           & $295$\\
                Profile       & Right half    & $1,\!221$           & $306$\\
                \hline
                \multirow{2}{*}{Profile} 
                              & Left          & $2,\!080$           & $521$\\
                              & Right         & $2,\!190$           & $548$\\
                \myTopRule
                Total         & -             & $15,\!449$          & $3,\!865$\\
                \myTopRule
            \end{tabular}
            \captionof{table}{Statistics of our new dataset for face alignment.}
            \label{tab:dataset_statistics}
        \end{minipage}
        \hfill
        \begin{minipage}[b]{0.44\linewidth}
            \centering
            \includegraphics[trim=0 0 0 50, clip, width=1.85cm]{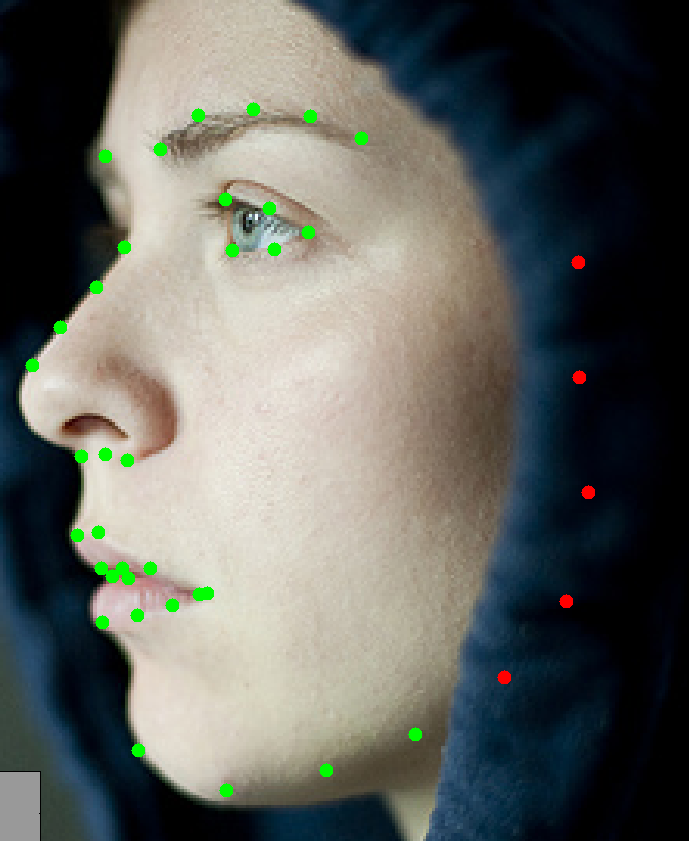}
            \includegraphics[trim=0 0 0 0, clip, width=1.5cm]{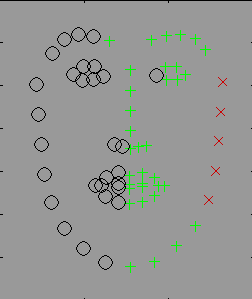}
            \captionof{figure}{\textcolor{green}{\textbf{Unoccluded}}, \textcolor{my_red}{\textbf{externally occluded}}, and self-occluded landmarks.}
            \label{fig:sample}
        \end{minipage}
        \vspace{-0.5cm}
    \end{table}

\section{Experiments}\label{sec:exp}
\vspace{-0.15cm}
    Our experiments use the datasets \threehundredW~\cite{sagonas2013300, sagonas2016300, sagonas2013semi}, \threehundredLP~\cite{zhu2016face}, \menpoTwoD~\cite{zafeiriou2017menpo, trigeorgis2016mnemonic, deng2019menpo}, \cofwSixtyEight~\cite{burgos2013robust, ghiasi2015occlusion}, \aflwNineteen~\cite{koestinger2011annotated}, WFLW~\cite{wu2018look}, and our \ourdataset~dataset. Training and testing protocols are described in the supplementary material. On a $12$ GB GeForce GTX Titan-X GPU, the inference time per image is $17$ ms.

    \textbf{Evaluation Metrics.}
    We use the standard metrics NME, AUC, and FR~\cite{tang2019towards, chen2019face, wang2019adaptive}. In each table, we report results using the same  metric adopted in respective baselines.
        
        \textit{Normalized Mean Error (NME).} The NME is defined as:
        \begin{equation}
            \mathrm{NME~(\%)} = \frac{1}{N_p}\sum_{j=1}^{N_p}\visgd \frac{{\|\ground - \bmuF\|}_2}{d} \times 100,
            \label{eq:nme}
        \end{equation}
        where $\visgd$, $\ground$ and $\bmuF$ respectively denote the visibility, ground-truth and predicted location of landmark $j$ from the $K$th (final) U-net. The factor of $v_j$ is there because we cannot compute an error value for points without ground-truth location labels. Several variations of the normalizing term $d$ are used.  \nmebox~\cite{zafeiriou2017menpo, bulat2017far, chen2019face} sets $d$ to the geometric mean of the width and height of the ground-truth bounding box $\bigl(\sqrt{w_\text{bbox} \cdot h_\text{bbox}}\,\bigr)$, while \nmeocular~\cite{sagonas2013300, kowalski2017deep, tang2019towards} sets $d$ to the distance between the outer corners of the two eyes. If a ground-truth box is not provided, the tight bounding box of the landmarks is used~\cite{bulat2017far, chen2019face}. \nmediag~\cite{wu2018look, sun2019high} sets $d$ as the diagonal of the bounding box. 
    
        \textit{Area Under the Curve (AUC).} To compute the AUC, the cumulative distribution of the fraction of test images whose $\mathrm{NME~(\%)}$ is less than or equal to the value on the horizontal axis is first plotted. The AUC for a test set is then computed as the area under that curve, up to the cutoff NME value.

        \textit{Failure Rate (FR).} FR refers to the percentage of images in the test set whose NME is larger than a certain threshold.

    \subsection{\threehundredWHeading~Face Alignment}
    \label{sec:300w}
    \vspace{-0.15cm}
        We train on the \threehundredW~\cite{sagonas2013300, sagonas2016300, sagonas2013semi}, and test on \threehundredW, \menpoTwoD~\cite{zafeiriou2017menpo, trigeorgis2016mnemonic, deng2019menpo}, and \cofwSixtyEight~\cite{burgos2013robust, ghiasi2015occlusion}. Some of the models are pre-trained on the \threehundredLP~\cite{zhu2016face}.

        \textbf{Data Splits and Evaluation Metrics.}\label{sec_split}
        There are two commonly used train/test splits for \threehundredW; we evaluate our method on both. \textit{Split $1$:} The train set contains $3,\!148$ images and full test set has $689$ images~\cite{tang2019towards}. \textit{Split $2$:} The train set includes $3,\!837$ images and test set has $600$ images~\cite{chen2019face}. The model trained on Split $2$ is additionally evaluated on the $6,\!679$ near-frontal training images of \menpoTwoD~and $507$ test images of \cofwSixtyEight~\cite{chen2019face}. For Split $1$, we use \nmeocular~\cite{tang2019towards, sun2019high, wang2019adaptive}. For Split $2$, we use \nmebox~and $\mathrm{AUC}_{\text{box}}$ with  $7\%$ cutoff~\cite{bulat2017far, chen2019face}.

        \begin{table}[!tb]
            \caption{\nmeocular~on \threehundredW~Common, Challenge, and Full datasets (Split $1$). [Key: \firstkey{Best}, \secondkey{Second best}]}
            \label{tab:nme_split1}
            \centering
            \footnotesize
            \begin{tabular}{tcmccct}
                \myTopRule
                 & \multicolumn{3}{ct}{\nmeocular~$(\%) (\downarrow)$}\\[0.05cm]
                 & Common & Challenge & Full \\
                \myTopRule
                SAN~\cite{dong2018style}                   & $3.34$        & $6.60$        & $3.98$\\
                AVS~\cite{qian2019aggregation}             & $3.21$        & $6.49$        & $3.86$\\
                DAN~\cite{kowalski2017deep}                & $3.19$        & $5.24$        & $3.59$\\
                LAB (w/B)~\cite{wu2018look}                & $2.98$        & $5.19$        & $3.49$\\
                Teacher~\cite{dong2019teacher}             & $2.91$        & $5.91$        & $3.49$\\
                DU-Net (Public code)~\cite{tang2019towards}& $2.97$        & $5.53$        & $3.47$\\
                DeCaFa (More data)~\cite{dapogny2019decafa}& $2.93$        & $5.26$        & $3.39$\\
                HR-Net~\cite{sun2019high}                  & $2.87$        & $5.15$        & $3.32$\\
                HG-HSLE~\cite{zou2019learning}             & $2.85$        & \second{5.03} & $3.28$\\
                AWing~\cite{wang2019adaptive}              & \first{2.72}  & \first{4.52}  & \first{3.07}\\
                \hline
                LUVLi (Ours)                               & \second{2.76} & $5.16$        & \second{3.23}\\
                \myTopRule
            \end{tabular}
            \vspace{-0.1cm}
        \end{table} 
        
        \begin{table}[!tb]
            \caption{\nmebox~and \aucbox~comparisons on \threehundredW~Test (Split $2$), \menpoTwoD~and \cofwSixtyEight~datasets. \\ {}
            [Key: \firstkey{Best}, \secondkey{Second best}, * = Pretrained on \threehundredLP]}
            \label{tab:nme_split2}
            \centering
            \footnotesize
            \setlength{\tabcolsep}{0.075cm}
            \begin{tabular}{tp{2cm} m p{0.85cm}p{0.85cm}p{0.85cm}mp{0.85cm}p{0.85cm}p{0.85cm}t}
                \myTopRule
                \addlinespace[0.01cm]
                 & \multicolumn{3}{cm}{\nmebox~$(\%)$ ($\downarrow$)}& \multicolumn{3}{ct}{\aucbox~$(\%)$ ($\uparrow$)}\\[0.05cm]
                                 & \threehundredW & Menpo & COFW & \threehundredW & Menpo & COFW\\ 
                \myTopRule
                SAN*~\cite{dong2018style}~in~\cite{chen2019face} & $2.86$        & $2.95$       & $3.50$        & $59.7$        & $61.9$        & $51.9$\\ 
                $2$D-FAN*~\cite{bulat2017far}                    & $2.32$        & $2.16$       & $2.95$        & $66.5$        & $69.0$        & $57.5$\\
                KDN~\cite{kdnuncertain}                          & $2.49$        & $2.26$       & -             & $67.3$        & $68.2$        & - \\
                Softlabel*~\cite{chen2019face}                   & $2.32$        & $2.27$       & $2.92$        & $66.6$        & $67.4$        & $57.9$ \\
                KDN*~\cite{chen2019face}                         & \second{2.21} & \first{2.01} & \second{2.73} & \second{68.3} & \second{71.1} & $60.1$\\
                \hline
                LUVLi (Ours)                                     & $2.24$        & $2.18$       & $2.75$        & \second{68.3} & $70.1$        & \second{60.8}\\
                LUVLi* (Ours)                                    & \first{2.10}  & \second{2.04}& \first{2.57}  & \first{70.2}  & \first{71.9}  & \first{63.4}\\
                \myTopRule
            \end{tabular}
            \vspace{-0.3cm}
        \end{table}
        
        \textbf{Results: Localization and Cross-Dataset Evaluation.} 
        The face alignment results for \threehundredW~Split $1$ and Split $2$ are summarized in Table~\ref{tab:nme_split1} and~\ref{tab:nme_split2}, respectively. Table~\ref{tab:nme_split2} also shows the results of our model (trained on Split $2$) on the Menpo and \cofwSixtyEight~datasets, as in~\cite{bulat2017far, chen2019face}. The results in Table~\ref{tab:nme_split1} show that our LUVLi landmark localization is competitive with the SOTA methods on Split $1$, usually one of the best two. Table~\ref{tab:nme_split2} shows that LUVLi significantly outperforms the SOTA on Split $2$, performing best on $5$ out of the $6$ cases ($3$ datasets $\times$ $2$ metrics). This is particularly impressive on \threehundredW~Split $2$, because even though most of the other methods are pre-trained on the \threehundredLP~dataset (as was our best method, LUVLi*), our method without pre-training still outperforms the SOTA in $2$ of $6$ cases. Our method performs particularly well in the cross-dataset evaluation on the more challenging \cofwSixtyEight~dataset, which has multiple externally occluded landmarks.

        \begin{figure}[!tb]
            \centering
        	\begin{subfigure}{0.325\linewidth}
        		\includegraphics[width=\linewidth]{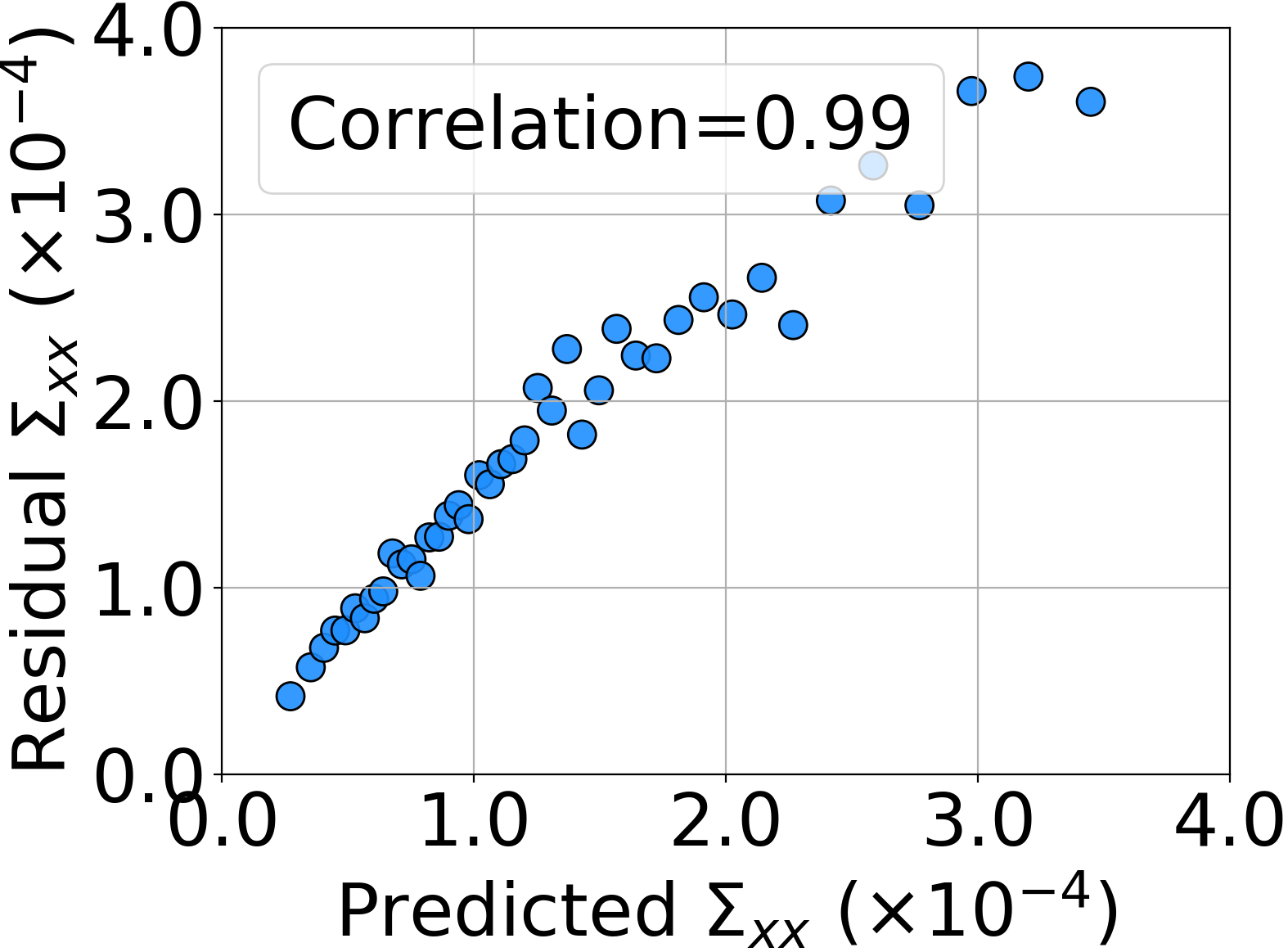}
        		\vspace{-5mm}
        		\captionof{figure}{variance of $x$}
        		\label{fig:res_vs_pred_x}
        	\end{subfigure}%
        	\hfill
        	\begin{subfigure}{0.325\linewidth}
        		\includegraphics[width=\linewidth]{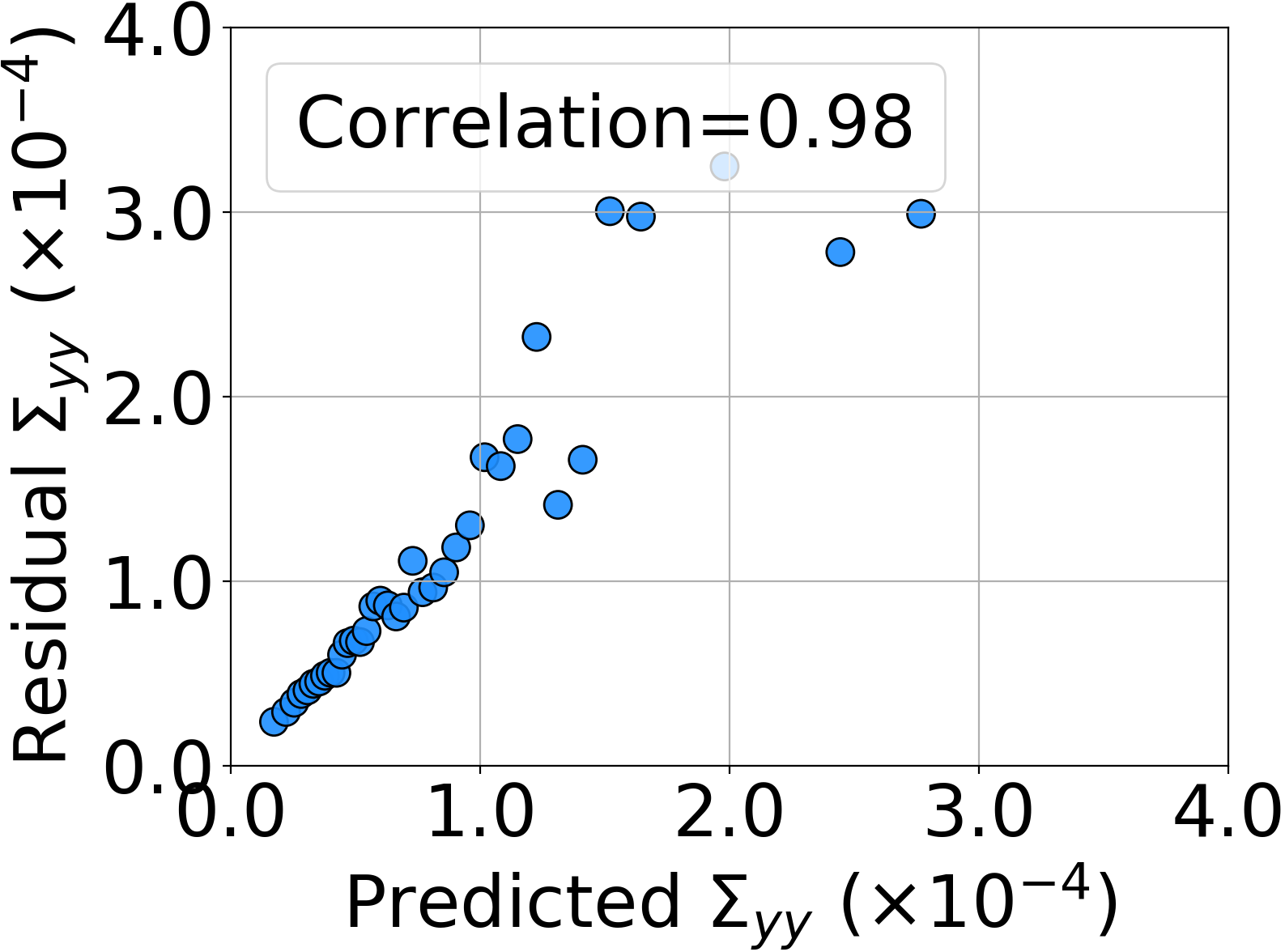}
        			\vspace{-5mm}
        		\captionof{figure}{variance of $y$}
        		\label{fig:res_vs_pred_y}
        	\end{subfigure}
        	\hfill
        	\begin{subfigure}{0.325\linewidth}
        		\includegraphics[width=\linewidth]{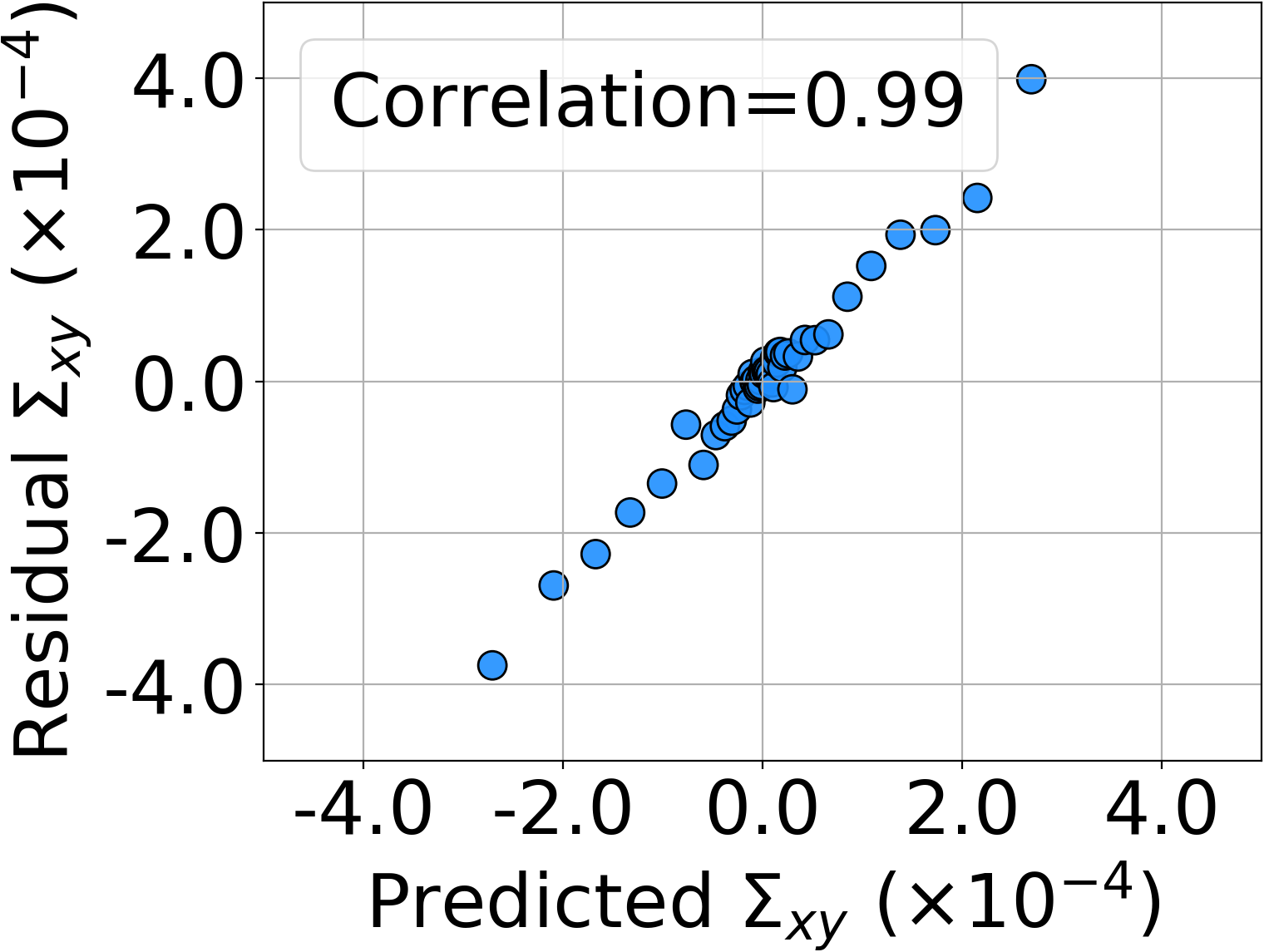}
        			\vspace{-5mm}
        		\captionof{figure}{covariance of $x,y$}
        		\label{fig:res_vs_pred_xy}
        	\end{subfigure}
        	\vspace{-0.3cm}
        	\caption{Mean squared residual error vs.~predicted covariance matrix for all landmarks in \threehundredW~Test (Split $2$).}
        	\label{fig:res_vs_pred}
    	    \vspace{-0.3cm}
        \end{figure}

        \textbf{Accuracy of Predicted Uncertainty.}
        To evaluate the accuracy of the predicted uncertainty covariance matrix, 
        $
        \scalemath{0.7}{
            \bSigmaF = 
            \begin{bmatrix}
            \bSigmaFxx  & \bSigmaFxy\\
            \bSigmaFxy  & \bSigmaFyy
            \end{bmatrix}},
        $
        we compare all three unique terms of this prediction with the statistics of the {\it residuals} ($2$D error between the ground-truth location $\ground$ and the predicted location $\bmuF$) of all landmarks in the test set. We explain how we do this for $\bSigmaFxx$ in Figure~\ref{fig:res_vs_pred_x}. First, we bin every landmark of every test image according to the value of the predicted variance in the $x$-direction $\bigl( \bSigmaFxx \bigr)$. Each bin is represented by one point in the scatter plot. Averaging $\bSigmaFxx$ across the $N_{\text{bin}} = 734$ landmark points within each bin gives a single predicted  $\bSigmaFxx$ value (horizontal axis). We next compute the residuals in the $x$-direction of all landmarks in the bin, and calculate the average of the squared residuals to obtain $\Sigma_{xx} = \EE(\groundx\!-\!\bmuFx)^2$ for the bin. This mean squared residual error, $\Sigma_{xx}$, is plotted on the vertical axis. If our predicted uncertainties are accurate, this residual error, $\Sigma_{xx}$, should be roughly equal to the predicted uncertainty variance in the $x$-direction (horizontal axis).
        
        Figure~\ref{fig:res_vs_pred} shows that all three terms of our method's predicted covariance matrices are highly predictive of the actual uncertainty: the mean squared residuals (error) are strongly proportional to the predicted covariance values, as evidenced by Pearson correlation coefficients of $0.98$ and $0.99$. However, decreasing $N_{\text{bin}}$ from 734 (plotted in Figure~\ref{fig:res_vs_pred}) to just 36 makes the correlation coefficients decrease to $0.84, 0.80, 0.72$.
        Thus, the predicted uncertainties are excellent after averaging but may yet have room to improve.

        \textbf{Uncertainty is Larger for Occluded Landmarks.} 
        The \cofwSixtyEight~\cite{ghiasi2015occlusion} test set annotates which landmarks are externally occluded. Similar to~\cite{chen2019face}, we use this to test uncertainty predictions of our model, where the square root of the determinant of the uncertainty covariance is a scalar measure of predicted uncertainty. We report the error, \nmebox, and average predicted uncertainty, $|\bSigmaF|^{1/2}$, in Table~\ref{tab:nme_ext_normal_cofw}. We do not use any occlusion annotation from the dataset during training. Like~\cite{chen2019face}, we find that our model's predicted uncertainty is much larger for externally occluded landmarks than for unoccluded landmarks. Furthermore, our method's location estimates are more accurate (smaller \nmebox) than those of~\cite{chen2019face} for both occluded and unoccluded landmarks.
        \begin{table}[!tb]
            \caption{\nmebox~and uncertainty $\bigl(|\bSigmaF|^{1/2} \bigr)$ on unoccluded and externally occluded landmarks of \cofwSixtyEight~dataset. [Key: \firstkey{Best}]}
            \label{tab:nme_ext_normal_cofw}
            \centering
            \footnotesize
            \setlength{\tabcolsep}{0.075cm}
            \begin{tabular}{|cmccmcc|}
                \myTopRule
                \addlinespace[0.01cm]
                    & \multicolumn{2}{cm}{Unoccluded} & \multicolumn{2}{c|}{Externally Occluded}\\
                    & \nmebox & $|\bSigmaNoSub|^{1/2}$ & \nmebox & $|\bSigmaNoSub|^{1/2}$\\
                \myTopRule
                Softlabel~\cite{chen2019face} & $2.30$ & $5.99$ & $5.01$ & $7.32$ \\
                KDN~\cite{chen2019face}       & $2.34$ & $1.63$ & $4.03$ & $11.62$ \\
                \hline
                LUVLi (Ours)                  & \first{2.15} & $9.31$ & \first{4.00} & $32.49$\\
                \myTopRule
            \end{tabular}
            \vspace{-0.2cm}
        \end{table}
        
\begin{figure}[t!]
            \centering
            \includegraphics[height=0.35\linewidth]{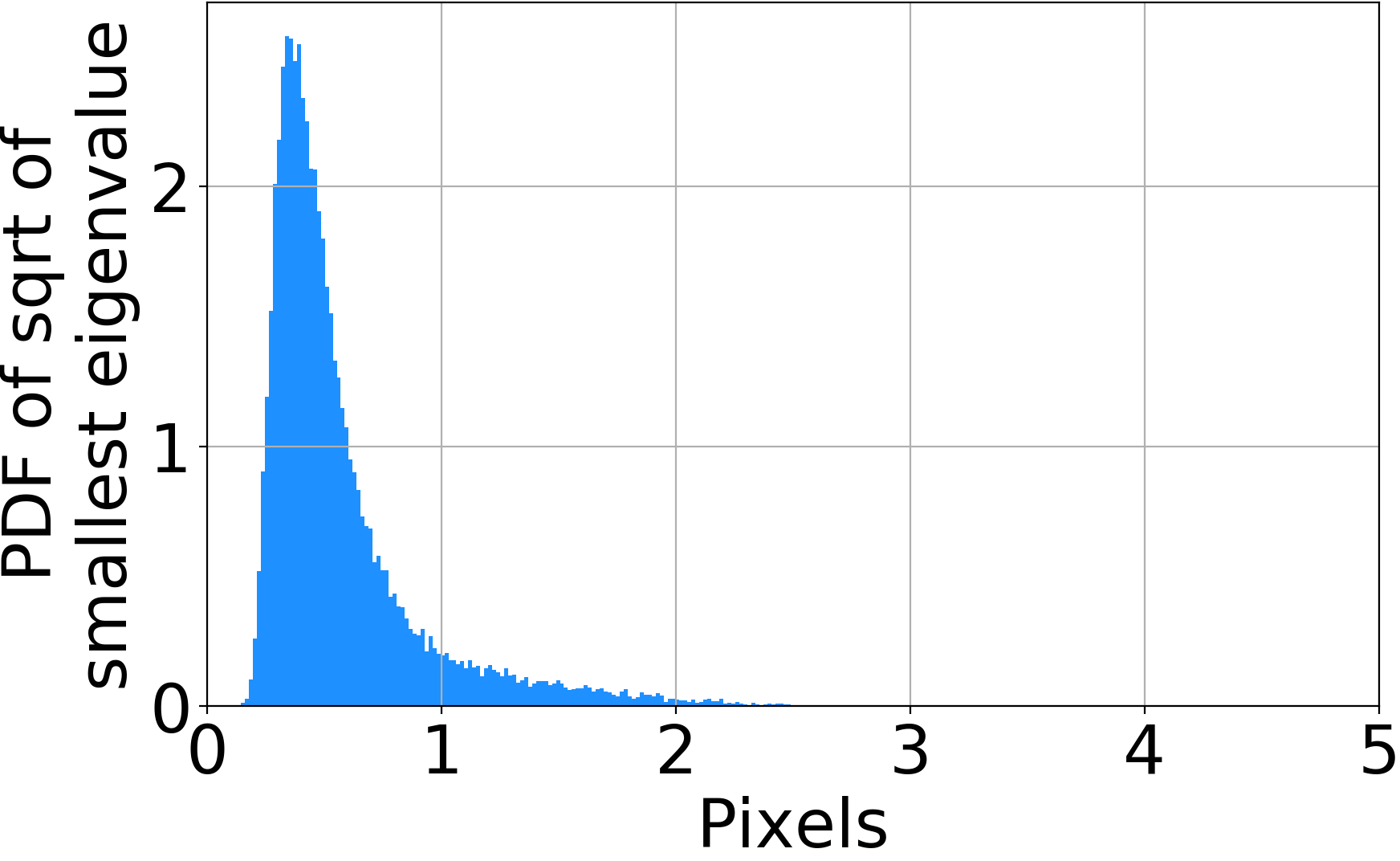}
            \vspace{-0.3cm}
            \caption{Histogram of the smallest eigenvalue of $\bSigmaF$.}
            \label{fig:hist_covar_small_eigen_value}
            \vspace{-0.5cm}
        \end{figure}

        \textbf{Heatmaps vs.~Direct Regression for Uncertainty.}
        We tried multiple approaches to estimate the uncertainty distribution from heatmaps, but none of these worked nearly as well as our direct regression using the CEN. We believe this is because in current heatmap-based networks, the resolution of the heatmap ($64 \times 64$) is too low for accurate uncertainty estimation. This is demonstrated in Figure~\ref{fig:hist_covar_small_eigen_value}, which shows a histogram over all landmarks in \threehundredW~Test (Split $2$) of LUVLi's predicted covariance in the narrowest direction of the covariance ellipse (the smallest eigenvalue of the predicted covariance matrix). The figure shows that in most cases, the uncertainty ellipses are less wide than one heatmap pixel, which explains why heatmap-based methods are not able to accurately capture such small uncertainties.
        
        \begin{table}[!tb]
            \caption{NME and AUC on the \aflwNineteen~dataset (previous results are quoted from~\cite{sun2019high, chen2019face}).
            [Key: \firstkey{Best}, \secondkey{Second best}]}
            \label{tab:nme_auc_aflw19}
            \centering
            \footnotesize
            \setlength{\tabcolsep}{0.1cm}
            \begin{tabular}{tcmccmccm}
                \myTopRule
                \addlinespace[0.01cm]
                & \multicolumn{2}{ct}{\nmediag} & \nmebox & \aucbox \\ 
                & Full & Frontal & Full& Full\\[0.05cm]
                \myTopRule
                CFSS~\cite{zhu2015face}        & $3.92$        & $2.68$        & -             & - \\
                CCL~\cite{zhu2016unconstrained}& $2.72$        & $2.17$        & -             & -\\
                DAC-CSR~\cite{feng2017dynamic} & $2.27$        & $1.81$        & -             & -\\
                LLL~\cite{robinson2019iccv}    & $1.97$        & -             & -             & - \\
                SAN~\cite{dong2018style}       & $1.91$        & $1.85$        & $4.04$        & $54.0$\\
                DSRN~\cite{miao2018direct}     & $1.86$        & -             & -             &- \\
                LAB (w/o B) \cite{wu2018look}  & $1.85$        & $1.62$        & -             &- \\
                HR-Net~\cite{sun2019high}      & \second{1.57} & \second{1.46} & -             &- \\
                Wing~\cite{feng2018wing}       & -             & -             & $3.56$        & $53.5$\\
                KDN~\cite{chen2019face}        & -             & -             & \second{2.80} & \second{60.3}\\
                \hline
                LUVLi (Ours)                  & \first{1.39}   & \first{1.19}  &\first{2.28}   & \first{68.0}\\  
                \myTopRule
            \end{tabular}
            \vspace{-0.1cm}
        \end{table}

    \subsection{\aflwNineteenHeading~Face Alignment}
    \vspace{-0.15cm}
    On \aflwNineteen, we train on $20,\!000$ images, and test on two sets: the \emph{AFLW-Full} set ($4,\!386$ test images) and the \emph{AFLW-Frontal} set ($1,\!314$ test images), as in~\cite{zhu2016unconstrained, wu2018look, sun2019high}. Table~\ref{tab:nme_auc_aflw19} compares our method's localization performance ~with other methods that only train on \aflwNineteen~(without training on any $68$-landmark dataset). Our proposed method outperforms not only the other uncertainty-based method KDN~\cite{chen2019face}, but also all previous SOTA methods, by a significant margin on both AFLW-Full and AFLW-Frontal.

    \subsection{WFLW Face Alignment}
    \vspace{-0.2cm}
    Landmark localization results for WFLW are shown in Table~\ref{tab:nme_wflw_short}. More detailed results on WFLW are in the supplementary material. Compared to the SOTA methods, LUVLi yields the second best performance on all metrics. Furthermore, while the other methods only predict landmark locations, LUVLi also estimates the prediction uncertainties.
        \begin{table}[!tb]
            \caption{WFLW-All dataset results for \nmeocular, \aucocular, and \frocular. [Key: \firstkey{Best}, \secondkey{Second best}]}
            \label{tab:nme_wflw_short}
            \centering
            \footnotesize
            \setlength{\tabcolsep}{0.1cm}
            \begin{tabular}{tcmctctct}
                \myTopRule
                 & NME(\%) ($\downarrow$) & AUC$^{10}$ ($\uparrow$)& FR$^{10}$(\%) ($\downarrow$)\\
                \myTopRule
                CFSS~\cite{zhu2015face}                & $9.07$        & $0.366$       & $20.56$\\
                DVLN~\cite{wu2017leveraging}           & $10.84$       & $0.456$       & $10.84$\\
                LAB (w/B)~\cite{wu2018look}            & $5.27$        & $0.532$       & $7.56$\\
                Wing~\cite{feng2018wing}               & $5.11$        & $0.554$       & $6.00$\\
                DeCaFa (w/DA)~\cite{dapogny2019decafa} & $4.62$        & $0.563$       & $4.84$\\
                AVS~\cite{qian2019aggregation}         & $4.39$        & \first{0.591} & $4.08$\\
                AWing~\cite{wang2019adaptive}          & \first{4.36}  & $0.572$       & \first{2.84}\\
                \hline
                LUVLi (Ours)                           & \second{4.37} & \second{0.577}& \second{3.12}\\ 
                \myTopRule
            \end{tabular}
            \vspace{-0.1cm}
        \end{table}

    \subsection{\ourdatasetHeading~Face Alignment}
    \vspace{-0.1cm}
        
        \textbf{Results of Landmark Localization.}
        Results for all head poses on our \ourdataset~dataset are shown in Table~\ref{tab:nme_aflw_ours}.
        \begin{table}[!tb]
        	\caption{\nmebox~and \aucbox~comparisons on \ourdataset~dataset. [Key: \firstkey{Best}]}
        	\label{tab:nme_aflw_ours}
        	\vspace{-3pt}
        	\setlength{\tabcolsep}{0.075cm}
        	\centering
            \footnotesize
            \begin{tabular}{tcmcmc|c|c|cm}
        		\myTopRule
        		Metric $(\%)$ & Method &	All	& Frontal & Half-Profile& Profile\\
                \myTopRule
        		\multirow{2}{*}{\nmebox ($\downarrow$)}
                &DU-Net~\cite{tang2019towards} & $1.99$       & $1.89$       & $2.50$       & $1.92$       \\
        		&LUVLi (Ours)                  & \first{1.61} & \first{1.74} & \first{1.79} & \first{1.25} \\
        		\myTopRule
        		\multirow{2}{*}{\aucbox  ($\uparrow$)}
        		&DU-Net~\cite{tang2019towards} & $71.80$       & $73.25$       & $64.78$       & $72.79$       \\
        		&LUVLi (Ours)                  & \first{77.08} & \first{75.33} & \first{74.69} & \first{82.10} \\
        		\myTopRule
        	\end{tabular}
            \vspace{-0.1cm}
        \end{table}
        
        \textbf{Results for All Visibility Classes.}
        We analyze LUVLi's performance on all test images for all three types of landmarks in Table~\ref{tab:analysis}. The first row is the mean value of the predicted visibility, $\widehat{v}_{j}$, for each type of landmark. Accuracy (Visible) tests the accuracy of predicting that landmarks are visible when $\widehat{v}_{j} > 0.5$. The last two rows show the scalar measure of uncertainty, $|\bSigmaF|^{1/2}$, both unnormalized and normalized by the face box size $\bigl( |\bSigmaNoSub|_{box}^{0.5} \bigr)$ similar to \nmebox. Similar to  results on \cofwSixtyEight~in Table~\ref{tab:nme_ext_normal_cofw}, the model predicts higher uncertainty for locations of externally occluded landmarks than for unoccluded landmarks. 

        \begin{table}[!tb]
            \caption{\ourdataset~results on three types of landmarks.} 
            \label{tab:analysis}
            \vspace{-3pt}
            \centering
            \footnotesize
            \setlength{\tabcolsep}{0.1cm}
            \begin{tabular}{|cmc|c|c|}
                \myTopRule
                \addlinespace[0.01cm]
                    & Self-Occluded & Unoccluded & Externally Occluded\\
                \myTopRule
                Mean $\widehat{v}_{j}$                       & $0.13$ & $0.98$ & $0.98$\\
                Accuracy (Visible)                           & $0.88$ & $0.99$ & $0.99$\\
                \nmebox                                      & -      & $1.60$ & $3.53$\\
                $|\bSigmaNoSub|^{0.5}$                       & -      & $9.28$ & $34.41$\\
                $|\bSigmaNoSub|_{box}^{0.5} (\times 10^{-4})$& -      & $1.87$ & $7.00$\\[0.05cm]
                \myTopRule
            \end{tabular}
            \vspace{-0.5cm}
        \end{table}

    \subsection{Ablation Studies}\label{sec:results_ablation}
    \vspace{-0.2cm}
    Table~\ref{tab:abl_1} compares modifications of our approach on Split $2$. Table~\ref{tab:abl_1} shows that computing the loss only on the last U-net performs worse than computing loss on all U-nets, perhaps because of the vanishing gradient problem~\cite{wei2016convolutional}. Moreover, LUVLi's log-likelihood loss without visibility outperforms using MSE loss on the landmark locations (which is equivalent to setting all $\bSigma = \mathbf{I}$). We also find that the loss with Laplacian likelihood~\eqref{eq:lapl_simplified_loss} outperforms the one with Gaussian likelihood~\eqref{eq:gauss_likelihood_loss}. Training from scratch is slightly inferior to first training the base DU-Net architecture before fine-tuning the full LUVLi network, consistent with previous observations that the model does not have strongly supervised pixel-wise gradients through the heatmap during training~\cite{nibali2018numerical}. Regarding the method for estimating the mean, using heatmaps is more effective than direct regression (Direct) from each U-net bottleneck, consistent with previous observations that neural networks have difficulty predicting continuous real values~\cite{Belagiannis17, nibali2018numerical}. As described in Section~\ref{sec:soft-argmax}, in addition to ReLU, we compared two other functions for $\sigma$: softmax, and a temperature-scaled softmax ($\tau$-softmax). Results for temperature-scaled softmax and ReLU are essentially tied, but the former is more complicated and requires tuning a temperature parameter, so we chose ReLU for our LUVLi model. Finally, reducing the number of U-nets from 8 to 4 increases test speed by about $2\!\times$ with minimal decrease in performance.

    \begin{table}[!tb]
        \caption{Ablation studies using our method trained on \threehundredLP~and then fine-tuned on \threehundredW~(Split $2$).}
        \label{tab:abl_1}
        \centering
        \footnotesize
        \setlength{\tabcolsep}{0.05cm}
        \begin{tabular}{tc|lmccmcct}
            \myTopRule
            \addlinespace[0.01cm]
            \multicolumn{2}{tcm}{\textbf{Change from LUVLi model:}} & \multicolumn{2}{cm}{\nmebox~$(\%)$} & \multicolumn{2}{ct}{\aucbox~$(\%)$}\\[0.05cm]
            Changed & From $\rightarrow$ To & \threehundredW & Menpo & \threehundredW & Menpo\\ 
            \myTopRule
            Supervision& All HGs $\rightarrow$ Last HG    & $2.32$          & $2.16$            & $67.7$          & $70.8$ \\
            \hline
            \multirow{4}{*}{Loss}
            & LUVLi$\rightarrow$MSE                       & $2.25$          & $2.10$            & $68.0$          & $71.0$ \\
            & Lap+vis$\rightarrow$Gauss+No-vis            & $2.15$          & $2.07$            & $69.6$          & $71.6$ \\
            & Lap+vis $\rightarrow$ Gauss+vis             & $2.13$          & $2.05$            & $69.8$          & $71.8$ \\
            & Lap+vis $\rightarrow$ Lap+No-vis            & $2.10$          & $2.05$            & $70.1$          & $71.8$ \\
            \hline 
            \multirow{2}{*}{Initialization}
            & LP-$2$D wts$\rightarrow$\threehundredW~wts  & $2.24$          & $2.18$            & $68.3$          & $70.1$ \\
            & LP-$2$D wts$\rightarrow$ Scratch            & $2.32$          & $2.26$            & $67.2$          & $69.4$ \\
            \hline 
            \multirow{3}{*}{\shortstack{Mean\\Estimator}} 
            & Heatmap $\rightarrow$ Direct                & $4.32$          & $3.99$            & $41.3$          & $47.5$ \\
            & ReLU $\rightarrow$ softmax                  & $2.37$          & $2.19$            & $66.4$          & $69.8$ \\
            & ReLU $\rightarrow \tau$-softmax             & $2.10$          & $2.04$            & $70.1$          & $71.8$ \\
            \hline
            No of HG & $8\rightarrow4$                    & $2.14$          & $2.07$            & $69.5$          & $71.5$ \\
            \hline
            {---} & \bf LUVLi (our best model)            & $\mathbf{2.10}$ & $\mathbf{2.04}$ & $\mathbf{70.2}$ & $\mathbf{71.9}$ \\
            \myTopRule
        \end{tabular}
        \vspace{-0.3cm}
    \end{table}

\vspace{-0.15cm}
\section{Conclusions}
\vspace{-0.15cm}
    In  this  paper, we present LUVLi, a novel end-to-end trainable framework for jointly estimating facial landmark locations, uncertainty, and visibility.  This joint estimation not only provides accurate uncertainty predictions, but also yields state-of-the-art estimates of the landmark locations on several datasets. We show that the predicted uncertainty distinguishes between unoccluded and externally occluded landmarks without any supervision for that task. In addition, the model achieves sub-pixel accuracy by taking the spatial mean of the ReLU'ed heatmap, rather than the arg max. We also introduce a new dataset containing manual labels of over $19,\!000$ face images with $68$ landmarks, which also labels every landmark with one of three visibility classes. Although our implementation is based on the DU-Net architecture, our framework is general enough to be applied to a variety of architectures for simultaneous estimation of landmark location, uncertainty, and visibility.

{\small
\bibliographystyle{ieee_fullname}
\bibliography{bibliography}
}

\clearpage
\appendix 
\twocolumn[\centering \section*{\Large LUVLi Face Alignment: Estimating Landmarks'\\ Location, Uncertainty, and Visibility Likelihood\\[12pt] Supplementary Material\\[18pt]}]

\renewcommand{\thesection}{A\arabic{section}}

\section{Implementation Details}
    Images are cropped using the detector bounding boxes provided by the dataset and resized to $256 \times 256$. Images with no detector bounding box are initialized by adding $5$\% uniform noise to the location of each edge of the tight bounding box around the landmarks, as in~\cite{bulat2017far}.
    
    \textbf{Training.}
    We modified the PyTorch~\cite{paszke2019pytorch} code for DU-Net~\cite{tang2019towards}, keeping the number of U-nets $K=8$ as in~\cite{tang2019towards}. Unless otherwise stated, we use the $2$D Laplacian likelihood~\eqref{eq:lapl_simplified_likelihood} as our landmark location likelihood, and therefore we use~\eqref{eq:lapl_simplified_loss} as our final loss function. All U-nets have equal weights $\lambda_i = 1$ in~\eqref{eq:loss_all}. For all datasets, visibility $\visgd = 1$ is assigned to unoccluded landmarks (those that are not labeled as occluded) and to landmarks that are labeled as externally occluded. Visibility $\visgd\!=\!0$ is assigned to landmarks that are labeled as self-occluded and landmarks whose labels are missing. 
    
    Training images for \threehundredW~Split $1$ are augmented randomly using scaling $(0.75-1.25)$, rotation $(-\ang{30},-\ang{30})$ and color jittering $(0.6,1.4)$ as in~\cite{tang2019towards}, while those from \threehundredW~Split $2$, \aflwNineteen, \wflwNinetyEight~and \ourdataset~datasets are augmented randomly using scaling $(0.8-1.2)$, rotation $(-\ang{50},\ang{50})$, color jittering $(0.6,1.4)$, and random occlusion, as in~\cite{bulat2017far}.
    
    The RMSprop optimizer is used as in~\cite{bulat2017far, tang2019towards}, with batch size $24$. Training from scratch takes $100$ epochs and starts with learning rate $2.5 \times 10^{-4}$, which is divided by $5$, $2$, and $2$ at epochs $30$, $60$, and $90$ respectively~\cite{tang2019towards}. When we initialize from  pretrained weights, we finetune for $50$ epochs using the LUVLi loss: $20$ with learning rate $10^{-4}$, followed by $30$ with learning rate $2\!\times\!10^{-5}$. We consider the model saved in the last epoch as our final model. 
    
    \textbf{Testing.}
    Whereas heatmap based methods~\cite{bulat2017far, tang2019towards, sun2019high} adjust their pixel output with a quarter-pixel offset in the direction from the highest response to the second highest response, we use the spatial mean as the landmark location without carrying out any adjustment nor shifting the heatmap even by a quarter of a pixel. We do not need to implement a sub-pixel shift, because our spatial mean over the ReLUed heatmaps already performs sub-pixel location prediction.

    \textbf{Spatial Mean}
    The spatial mean $\bmu$ of each of the heatmap is defined as 
        \begin{equation}
            \bmu = 
            \begin{bmatrix}
            \bmux\\ 
            \bmuy
            \end{bmatrix}
            = \dfrac{\mathlarger{\mathlarger{\sum}}\limits_{x,y} \sigma \bigl( \bh(x,y) \bigr)\begin{bmatrix}
            x\\
            y
            \end{bmatrix}}{\mathlarger{\mathlarger{\sum}}\limits_{x,y} \sigma \bigl( \bh(x,y) \bigr)}
            \label{eq:mean}
        \end{equation}
    where $\sigma\bigl(\bh(x,y)\bigr)$ denotes the output of post-processing the heatmap pixel with a function $\sigma$.

\section{Additional Experiments and Results}     
    We now provide additional results evaluating our system's performance in terms of both localization and uncertainty estimation.

    \subsection{System Trained on \threehundredWHeading}
        
        \subsubsection{Training}
        For Split~$1$, we initialized using the pre-trained DU-Net model available from the authors of~\cite{tang2019towards}, then fine-tuned on the \threehundredW~training set (Split~$1$) using our proposed architecture and LUVLi loss. For Split $2$, for the experiments in which we pre-trained on \threehundredLP, we pre-trained from scratch on \threehundredLP~using heatmaps (using the original DU-Net architecture and loss). We then fine-tuned on the \threehundredW~training set (Split~$2$) using our proposed architecture and LUVLi loss. 
        
        \subsubsection{Comparison with KDN~\cite{kdnuncertain}}
\begin{figure}[!htb]
            \centering
            \includegraphics[width=\linewidth]{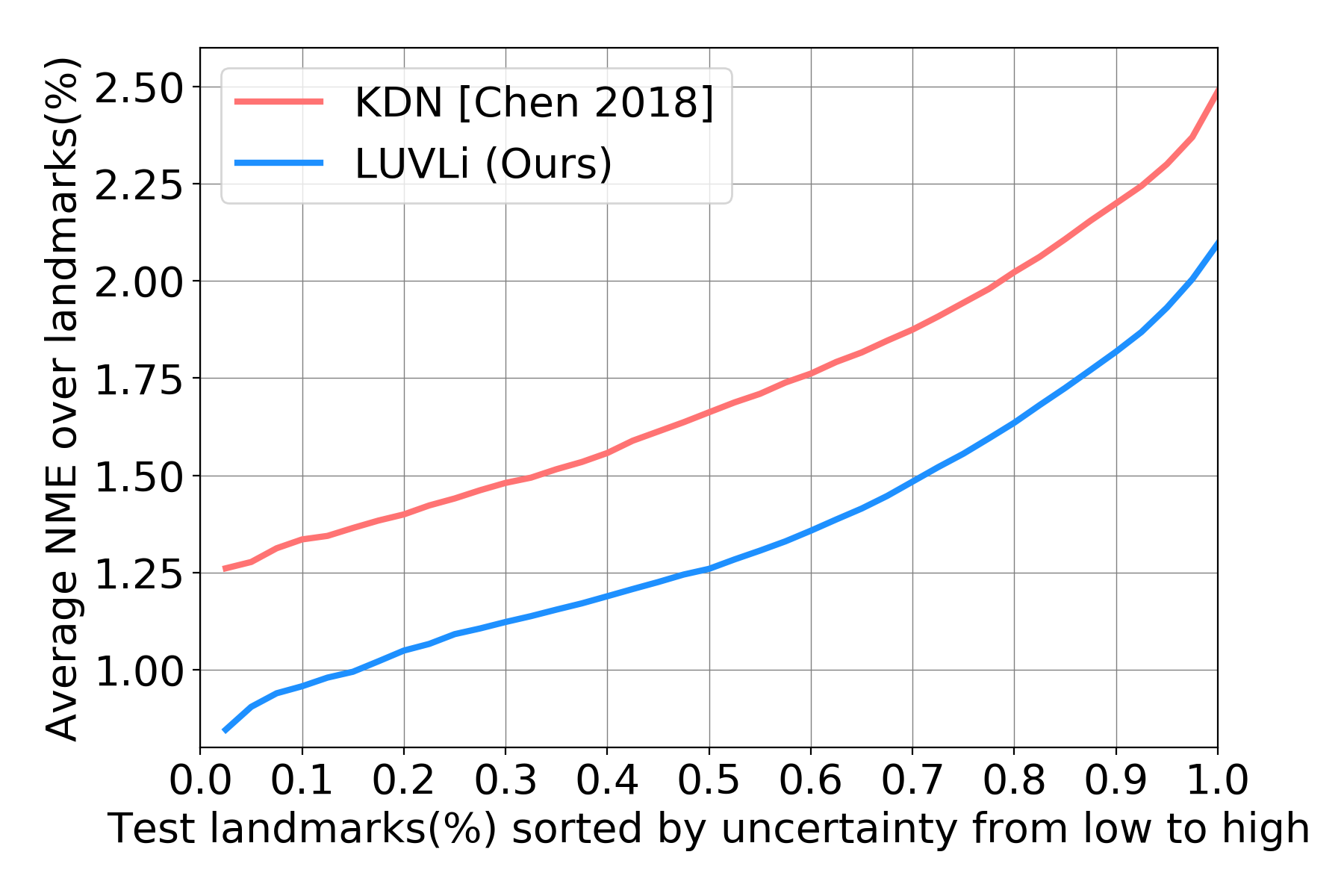}
            \caption{Average NME vs sorted uncertainty, averaged across landmarks in an image.}
            \label{fig:lisha_plot}
        \end{figure}

        To compare directly with Chen et al.~\cite{kdnuncertain}, in Figure~\ref{fig:lisha_plot} we plot normalized mean error (NME) vs. predicted uncertainty (rank, from smallest to largest), as in Figure 1 of~\cite{kdnuncertain}. (We obtained the predicted uncertainty and NME data of~\cite{kdnuncertain} from the authors.) The figure shows that for our method as well as for~\cite{kdnuncertain}, there is a strong trend that higher predicted uncertainties correspond to larger location errors. However, the errors of our method are significantly smaller than the errors produced by~\cite{kdnuncertain}.
        
        \subsubsection{Verifying Predicted Uncertainty Distributions}
        \begin{figure}[!htb]
            \centering
            \includegraphics[width=\linewidth]{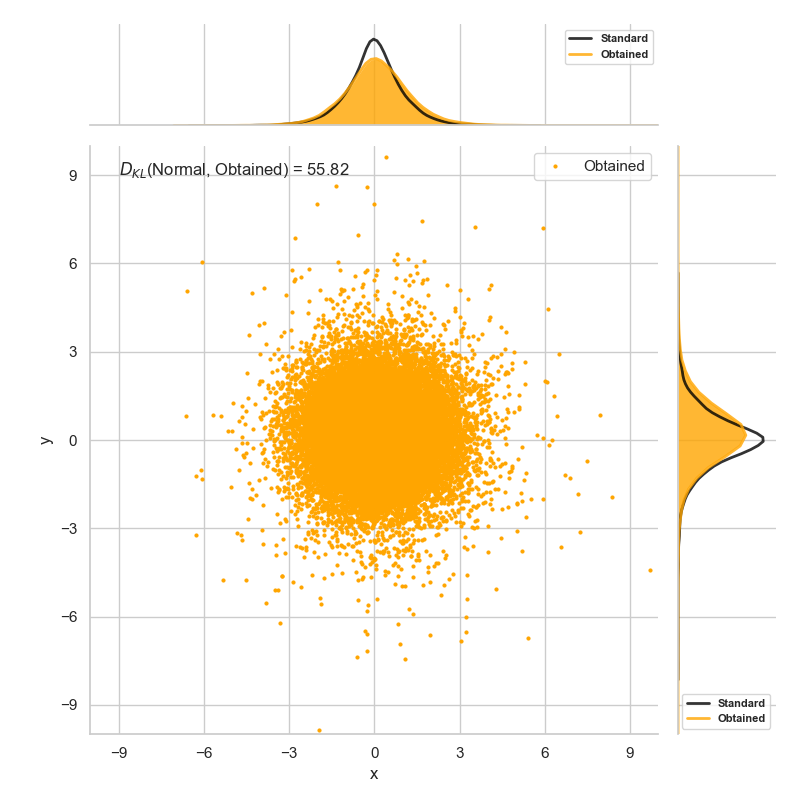}
        	\caption{Scatter plot of transformed ground-truth locations, $\ground' = \bSigmaF^{-0.5}(\ground - \bmuF)$, for \threehundredW~Test (Split $2$). The histograms (orange) of their $x$ and $y$ coordinates are very close to the the marginal pdf (black curves) of the Standard Laplacian distribution $P(\z'|\zerovec, \I)$.}
            \label{fig:transformed}    
        \end{figure}
        For every image, for each landmark $j$, our network predicts a mean $\bmuF$ and a covariance matrix $\bSigmaF$. We can view this as our network predicting that a human labeler of that image will effectively select the landmark location $\ground$ for that image from the Laplacian distribution from~\eqref{eq:lapl_simplified_likelihood} with mean $\bmuF$ and covariance $\bSigmaF$:
        \begin{equation}
            \ground \sim P(\z|\bmuF,\! \bSigmaF) =
            \frac{e^{-\sqrt{3(\z-\bmuF)^T\bSigmaF^{-1} (\z-\bmuF)}} }{\frac{2\pi}{3}\sqrt{\left| \bSigmaF \right|}}.
            \label{eq:lapl_prediction}
        \end{equation}
        If we had multiple labels (\example, ground-truth landmark locations from multiple human labelers) for a single landmark in one image, then it would be straightforward to evaluate how well our method's predicted probability distribution matches the distribution of labeled landmark locations. Unfortunately, face alignment datasets only have a single ground-truth location for each landmark in each image. This makes it difficult, but not impossible, to evaluate how well the human labels for images in the test set fit our method's predicted uncertainty distributions. We propose the following method for verifying the predicted probability distributions.
        
        Suppose we transform the ground-truth location of a landmark, $\ground$, using the predicted mean and covariance for that landmark as follows:
        \begin{equation}
            \ground' = \bSigmaF^{-0.5}(\ground - \bmuF).
            \label{eq:transformed_gt}
        \end{equation}
        If our method's predictions are correct, then from~\eqref{eq:lapl_prediction},  \mbox{$\ground \sim P(\z|\bmuF,\! \bSigmaF)$}. Hence, $\ground'$ is drawn from the transformed distribution $P(\z')$, where $\z' = \bSigmaF^{-0.5}(\z - \bmuF)$:
        \begin{equation}
            \ground' \sim P(\z'|\zerovec, \I) =
            \frac{e^{-\sqrt{3\z'^T\z'}} }{{2\pi}/{3}}.
            \label{eq:lapl_prediction_transformed}
        \end{equation}
        After this simple transformation (transforming the labeled ground-truth location $\ground$ of each landmark using its predicted mean and covariance), we have transformed our network's prediction about $\ground$ into a prediction about $\ground'$ that is much easier to evaluate, because the distribution in~\eqref{eq:lapl_prediction_transformed} is simply a standard $2$D Laplacian distribution---it no longer depends on the predicted mean and covariance.
        
        Thus, our method predicts that after the transformation~\eqref{eq:transformed_gt}, every ground-truth landmark location $\ground'$ is drawn from the same standard $2$D Laplacian distribution~\eqref{eq:lapl_prediction_transformed}. Now that we have an entire population of transformed labels that our model predicts are all drawn from the same distribution, it is easy to verify whether the labels fit our model's predictions. Figure~\ref{fig:transformed} shows a scatter plot of the transformed locations, $\ground'$, for all landmarks in all test images of \threehundredW~(Split $2$). We plot the histogram of the marginalized landmark locations ($x$- or $y$-coordinate of $\ground'$) in orange above and to the right of the plot, and overlay the marginal pdf of the standard Laplacian~\eqref{eq:lapl_prediction_transformed} in black.  The excellent match between the transformed landmark locations and the standard Laplacian distribution indicates that our model's predicted uncertainty distributions are quite accurate. Since Kullback-Leibler (KL) divergence  is invariant to affine transformations like the one in~\eqref{eq:transformed_gt}, we can evaluate the KL-divergence (printed at the top of the scatterplot) between the standard $2$D Laplacian distribution~\eqref{eq:lapl_prediction_transformed} and the distribution of the transformed landmark locations (using their $2$D histograms) as a numerical measure of how well the predictions of our model fit the distribution of labeled locations.
        
        \subsubsection{Relationship to Variation Among Human\\Labelers on Multi-PIE}
\begin{figure}[!htb]
            \centering
            \includegraphics[width=1\linewidth]{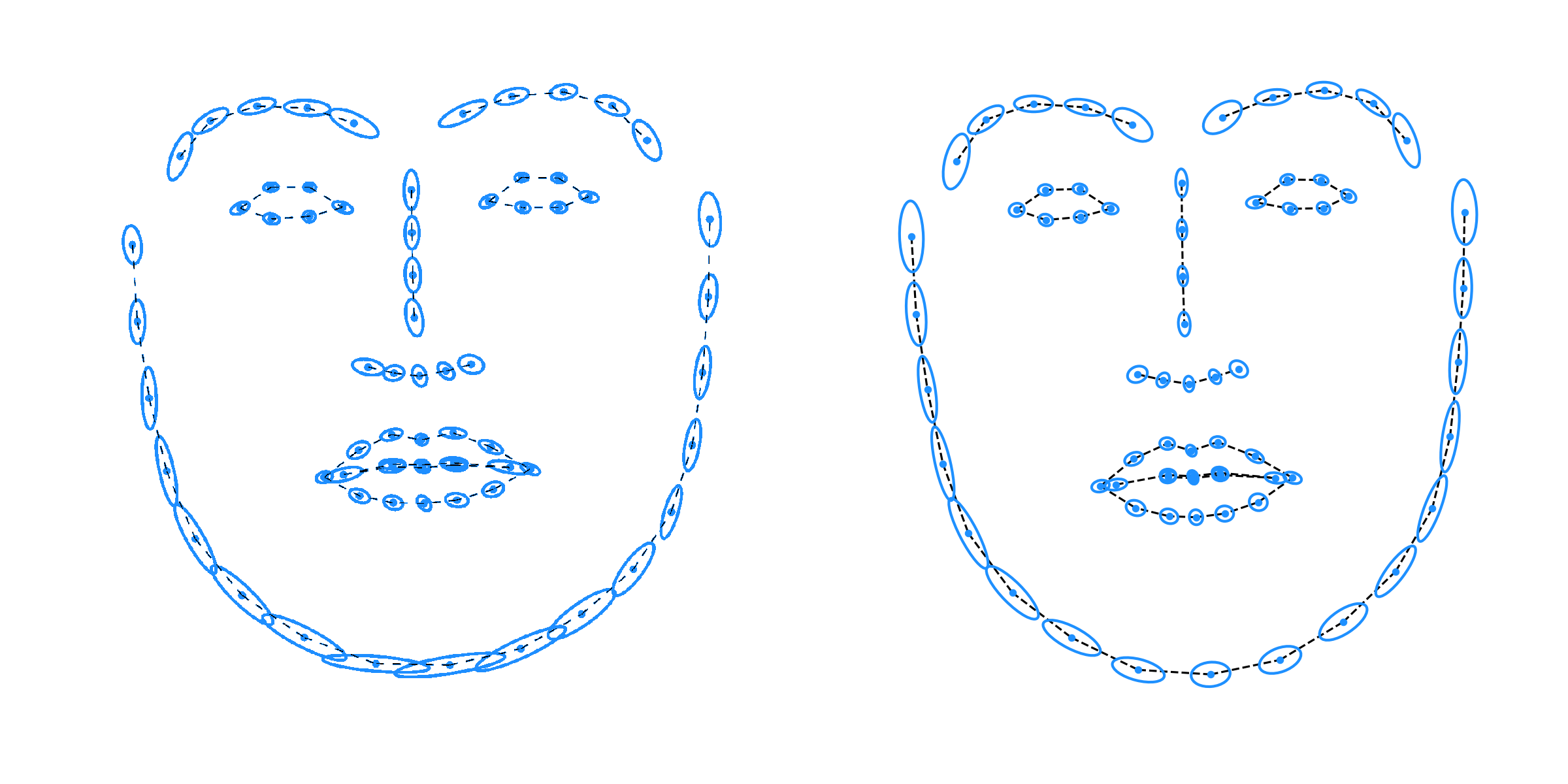}
        	\caption{Variation across three human labelers~\cite{sagonas2016300} ({\em left}) versus uncertainties computed by our proposed method on frontal images of Multi-PIE dataset ({\em right}).} 
            \label{fig:multi_pie}    
        \end{figure}

        We test our Split $2$ model on $812$ frontal face images of all subjects from the Multi-PIE dataset~\cite{gross2010multi}, then compute the mean of the uncertainty ellipses predicted by our model across all $812$ images. To compute the mean, we first normalize the location of each landmark using the inter-ocular distance, as in~\cite{sagonas2016300}, and also normalize the covariance matrix by the square of the inter-ocular distance. We then take the average of the normalized locations across all faces to obtain the mean landmark location. The covariance matrices are averaged across all faces using the log-mean-exponential technique. The mean location and covariance matrix of each landmark (averaged across all faces) is then used to plot the results which are shown on the right in Figure~\ref{fig:multi_pie}.
        
        We compare our model predictions with Figure 5 of~\cite{sagonas2016300}, shown on the left of Figure~\ref{fig:multi_pie}. To create that figure,~\cite{sagonas2016300} tasked three different human labelers with annotating the same frontal face images from the Multi-PIE database of 80 different subjects in frontal pose with neutral expression. For each landmark, they plotted the the covariance of the label locations across the three labelers using an ellipse. Note the similarity between our model's predicted uncertainties (on the right of Figure~\ref{fig:multi_pie} and the covariance across human labelers (on the left of Figure~\ref{fig:multi_pie}), especially around the eyes, nose, and mouth. Around the outside edge of the face, note that our model predicts that label locations will vary primarily in the direction parallel to the edge, which is precisely the pattern observed across human labelers.
        
        \subsubsection{Sample Uncertainty Ellipses on Multi-PIE}
\begin{figure}[!htb]
            \centering
            \includegraphics[width=0.7\linewidth]{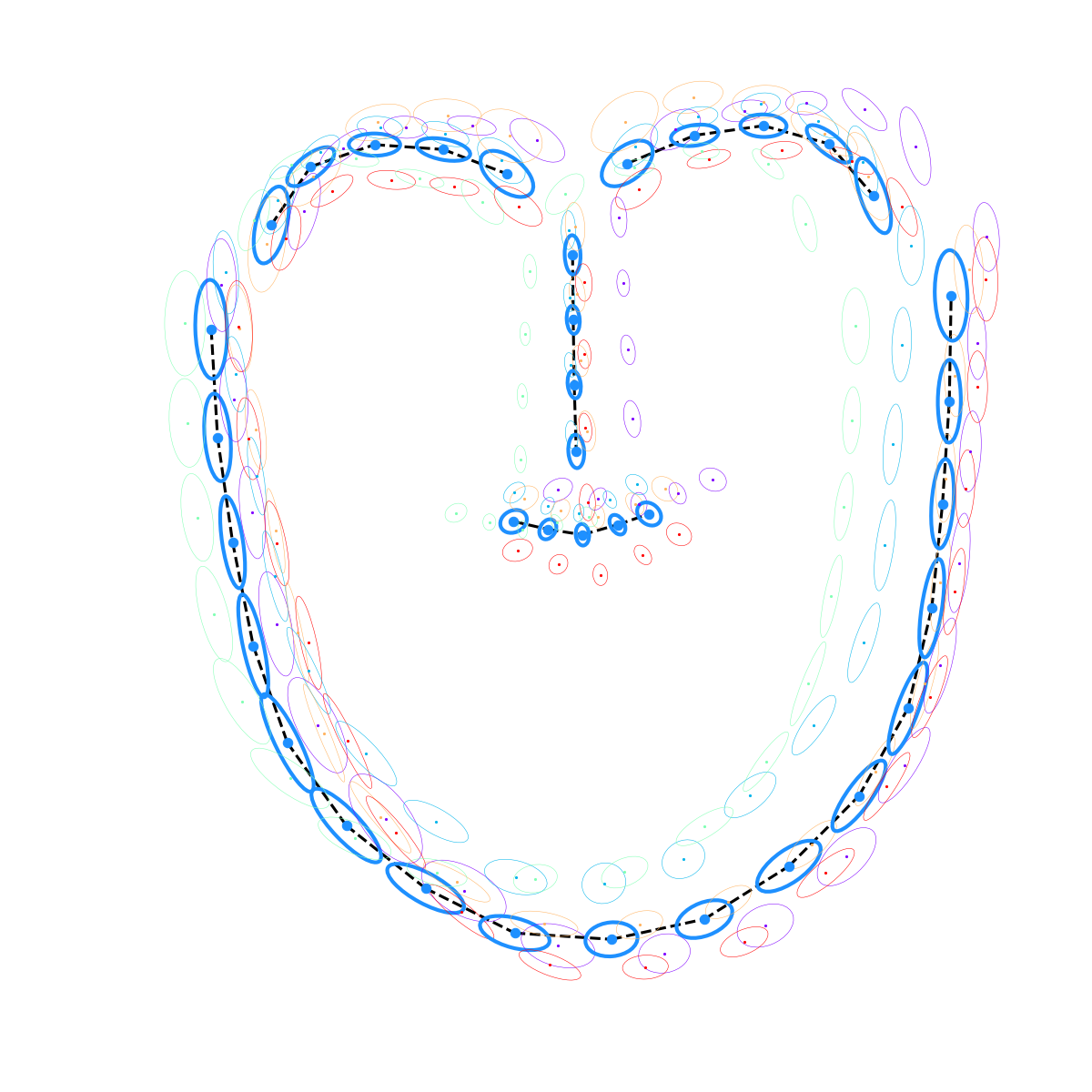}
        	\caption{Our model's uncertainty predictions for some individual frontal face images from the Multi-PIE dataset (various colors), overlaid with the mean uncertainty predictions across all frontal Multi-PIE faces (blue, copied from Figure~\ref{fig:multi_pie}).} 
            \label{fig:multi_pie_2}    
        \end{figure}

        To illustrate how the predicted uncertainties output by our method vary across different subjects from Multi-PIE, in Figure~\ref{fig:multi_pie_2} we overlay our model's mean uncertainty predictions (in blue, copied from right side of Figure~\ref{fig:multi_pie}) with our model's predicted uncertainties of some of the individual Multi-PIE face images (in various colors). To simplify the figure, we plot all landmarks except for the eyes, nose, and mouth.
    
        \subsubsection{Laplacian vs. Gaussian Likelihood}
        We have described two versions of our model: one whose loss function~\eqref{eq:lapl_simplified_loss} uses a $2$D Laplacian probability distribution~\eqref{eq:lapl_simplified_likelihood}, and another whose loss function~\eqref{eq:gauss_likelihood_loss} uses a $2$D Gaussian probability distribution~\eqref{eq:gauss_likelihood}. We now discuss the question of which of these two models performs better. 
        
        The numerical comparisons are shown in Table~\ref{tab:likelihood_compare}. The numbers in the first two columns of the table were also presented in the ablation studies table, Table~\ref{tab:abl_1}.
        
        \textbf{Comparing the Predicted Locations.} If we consider only the errors of the predicted landmark locations, the first two columns of Table~\ref{tab:likelihood_compare} show that the Laplacian model is slightly better: The Laplacian model has a smaller value of \nmebox {} and a larger value of \aucbox.
        
        \textbf{Comparing the Predicted Uncertainties.} 
        To compare the two models' predicted uncertainties as well as their predicted locations, we consider the probability distributions over landmark locations that are predicted by each model. We want to know which model's predicted probability distributions better explain the ground-truth locations of the landmarks in the test images. In other words, we want to know which model assigns a higher likelihood to the ground-truth landmark locations (\thatIs, which model yields a lower negative log-likelihood on the test data). We compute the negative log-likelihood of the ground-truth locations $\ground$ from the last hourglass using ~\eqref{eq:lapl_simplified_loss} for the Laplacian model and~\eqref{eq:gauss_likelihood_loss} for the Gaussian model. The results, in the last column of Table~\ref{tab:likelihood_compare}, show that the Laplacian model gives a lower negative log-likelihood. In other words, the ground-truth landmark locations have a higher likelihood under our Laplacian model. We conclude that the learned Laplacian model explains the human labels better than the learned Gaussian model.
        
        \begin{table}[!tb]
           \caption{Comparison of our model with Laplacian likelihood vs. with Gaussian likelihood, on \threehundredW~Test (Split $2$). \\ {}
           [Key:  ($\uparrow$) = higher is better;($\downarrow$) = lower is better ]}
           \label{tab:likelihood_compare}
           \centering
           \footnotesize
           \begin{tabular}{tc|cc|cm}
                \myTopRule
                Likelihood & \nmebox (\%) ($\downarrow$) & \aucbox (\%) ($\uparrow$) & NLL ($\downarrow$)\\
                \myTopRule
                Laplacian  & $2.10$ & $70.1$ & $0.51$ \\ 
                Gaussian   & $2.13$ & $69.8$ & $0.66$ \\  
                \myTopRule
           \end{tabular}
        \end{table}
       
        \begin{table}[!tb]
        	\caption{\nmeboxvis~comparisons on \ourdataset~dataset. [Key: \firstkey{Best}]}
        	\label{tab:nme_aflw_ours_nme_vis}
        	\vspace{-3pt}
        	\setlength{\tabcolsep}{0.075cm}
        	\centering
            \footnotesize
            \begin{tabular}{tcmcmc|c|c|cm}
        		\myTopRule
        		Metric $(\%)$ & Method &	All	& Frontal & Half-Profile& Profile\\
                \myTopRule
        		\multirow{2}{*}{\nmeboxvis($\downarrow$)}
                &DU-Net~\cite{tang2019towards} & $2.27$       & $1.91$       & $2.77$       & $3.10$       \\
        		&LUVLi (Ours)                  & \first{1.84} & \first{1.75} & \first{1.99} & \first{2.03} \\
        		\myTopRule
        		\multirow{2}{*}{\nmebox ($\downarrow$)}
                &DU-Net~\cite{tang2019towards} & $1.99$       & $1.89$       & $2.50$       & $1.92$       \\
        		&LUVLi (Ours)                  & \first{1.61} & \first{1.74} & \first{1.79} & \first{1.25} \\
        		\myTopRule
        	\end{tabular}
        \end{table} 
        
        \begin{table*}[ht]
        	\caption{\nmeocular~and \aucocular~comparison between our proposed method and the state-of-the-art landmark localization methods on the WFLW dataset. \\ {}
        	[Key: \firstkey{Best}, \secondkey{Second best}; (w/DA) = uses more data; (w/B) = uses boundary; ($\downarrow$) = smaller is better; ($\uparrow$) = larger is better] }
        	\label{tab:nme_wflw}
        	\centering
            \footnotesize
        	\begin{tabular}{cmcmc|c|c|c|c|c|c}
        		\myTopRule
        		Metric                & Method          & All	       & Head Pose     & Expression    & Illumination   & Make-up       & Occlusion     & Blur\\
                
                \myTopRule
        		\multirow{10}{*}{\nmeocular $(\%)$ ($\downarrow$)}
        		&CFSS~\cite{zhu2015face}                & $9.07$        & $21.36$       & $10.09$       & $8.30$        & $8.74$        & $11.76$       & $9.96$ \\
        		&DVLN~\cite{wu2017leveraging}           & $10.84$       & $46.93$       & $11.15$       & $7.31$        & $11.65$       & $16.30$       & $13.71$\\
        		&LAB (w/B)~\cite{wu2018look}            & $5.27$        & $10.24$       & $5.51$        & $5.23$        & $5.15$        & $6.79$        & $6.32$\\
        		&Wing~\cite{feng2018wing}               & $5.11$        & $8.75$        & $5.36$        & $4.93$        & $5.41$        & $6.37$        & $5.81$\\
        		&DeCaFA (w/DA)~\cite{dapogny2019decafa} & $4.62$        & $8.11$        & \second{4.65} & $4.41$        & $4.63$        & $5.74$        & $5.38$\\
        		&HR-Net~\cite{sun2019high}              & $4.60$        & $7.94$        & $4.85$        & $4.55$        & \second{4.29} & $5.44$        & $5.42$\\
        		&AVS~\cite{qian2019aggregation}         & $4.39$        & $8.42$        & $4.68$        & \first{4.24}  & $4.37$        & $5.60$        & \first{4.86}\\
        		&AWing~\cite{wang2019adaptive}          & \first{4.36}  & \first{7.38}  & \first{4.58}  & $4.32$        & \first{4.27}  & \first{5.19}  & $4.96$\\
        		\cline{2-9}
        		&LUVLi (Ours)                           & \second{4.37} & \second{7.56} & $4.77$        & \second{4.30} & $4.33$        & \second{5.29} & \second{4.94}\\
        		\cline{2-9}
        		
        		\myTopRule
        		\multirow{9}{*}{\aucocular ($\uparrow$)}
        		&CFSS~\cite{zhu2015face}                & $0.366$        & $0.063$        & $0.316$        & $0.385$        & $0.369$        & $0.269$        & $0.303$\\
        		&DVLN~\cite{wu2017leveraging}           & $0.456$        & $0.147$        & $0.389$        & $0.474$        & $0.449$        & $0.379$        & $0.397$\\
        		&LAB (w/B)~\cite{wu2018look}	        & $0.532$        & $0.235$        & $0.495$        & $0.543$        & $0.539$        & $0.449$        & $0.463$\\
        		&Wing~\cite{feng2018wing}               & $0.554$        & $0.310$        & $0.496$        & $0.541$        & $0.558$        & $0.489$        & $0.492$\\
        		&DeCaFA (w/DA)~\cite{dapogny2019decafa} & $0.563$        & $0.292$        & $0.546$        & $0.579$        & $0.575$        & $0.485$        & $0.494$\\
        		&AVS~\cite{qian2019aggregation}         & \first{0.591}  & \second{0.311} & \first{0.549}  & \first{0.609}  & \second{0.581} & \first{0.517}  & \first{0.551}\\
        		&AWing~\cite{wang2019adaptive}          & $0.572$        & \first{0.312}  & \second{0.515} & $0.578$        & $0.572$        & $0.502$        & $0.512$\\
        		\cline{2-9}
        		&LUVLi (Ours)                           & \second{0.577} & $0.310$        & \first{0.549}  & \second{0.584} & \first{0.588}  & \second{0.505} &	\second{0.525}\\
        		\cline{2-9}
        		
        		\myTopRule
        		\multirow{9}{*}{\frocular$(\%)$ ($\downarrow$)}
        		&CFSS~\cite{zhu2015face}               & $20.56$       & $66.26$        & $23.25$       & $17.34$       & $21.84$       & $32.88$       & $23.67$ \\
        		&DVLN~\cite{wu2017leveraging}          & $10.84$       & $46.93$        & $11.15$       & $7.31$        & $11.65$       & $16.30$       & $13.71$\\
        		&LAB (w/B)~\cite{wu2018look}	       & $7.56$        & $28.83$        & $6.37$        & $6.73$        & $7.77$        & $13.72$       & $10.74$\\
        		&Wing~\cite{feng2018wing}              & $6.00$        & $22.70$        & $4.78$        & $4.30$        & $7.77$        & $12.50$       & $7.76$\\
        		&DeCaFA(w/DA)~\cite{dapogny2019decafa} & $4.84$        & $21.40$        & $3.73$        & $3.22$        & $6.15$        & $9.26$        & $6.61$\\
        		&AVS~\cite{qian2019aggregation}        & $4.08$        & $18.10$        & $4.46$        & $2.72$        & $4.37$        & $7.74$        & $4.40$\\
        		&AWing~\cite{wang2019adaptive}         & \first{2.84}  & \first{13.50}  & \first{2.23}  & \second{2.58} & \first{2.91}  & \first{5.98}  & \second{3.75}\\
        		\cline{2-9}
        		&LUVLi (Ours)                          & \second{3.12} & \second{15.95} & \second{3.18} & \first{2.15}  & \second{3.40} & \second{6.39} & \first{3.23}\\
        		\myTopRule
        	\end{tabular}
        \end{table*}

    \subsection{WFLW Face Alignment}\label{subsec:wflw}
        
        \textbf{Data Splits and Implementation Details.}
        The training set consists of $7,\!500$ images, while the test set consists of $2,\!500$ images. In Table~\ref{tab:nme_wflw}, we report results on the entire test set (All), which we also reported in Table~\ref{tab:nme_wflw_short}. In Table~\ref{tab:nme_wflw}, we additionally report results on several subsets of the test set: large head pose ($326$ images), facial expression ($314$ images), illumination ($698$ images), make-up ($206$ images), occlusion ($736$ images), and blur ($773$ images). The images are cropped using the detector bounding boxes provided by~\cite{sun2019high} and resized to $256 \times 256$. 
    
        We first train the images with the heatmaps on proxy ground-truth heatmaps, then finetune using our proposed LUVLi loss. \nmeocular, \aucocular {}, and \frocular {} are used as evaluation metrics, as in~\cite{sun2019high, wang2019adaptive, dapogny2019decafa}. We report AUC and FR with cutoff $10\%$ as in~\cite{sun2019high, wang2019adaptive, dapogny2019decafa}.
        
        \textbf{Results of Facial Landmark Localization.}
        Table~\ref{tab:nme_wflw} compares our method's landmark localization results with those of other state-of-the-art methods on the WFLW dataset. Our method performs performs in the top two methods on all the metrics.  Importantly, all of the other methods only predict landmark locations--they do not predict the uncertainty of their estimated landmark locations. Not only does our method place in the top two on all three landmark localization metrics, but our method also accurately predicts its own uncertainty of landmark localization.

    \subsection{\ourdatasetHeading~Face Alignment}
    We next define a modified version of the evaluation metric NME that may be more appropriate for face images with extreme head pose. Whereas NME as defined in~\eqref{eq:nme} 
    divides by the total number of landmarks $N_p$, the modified NME instead divides by the number of visible landmarks. This metric, which we call $\mathrm{NME^{\text{vis}}}$, computes the mean across only the visible (unoccluded and externally occluded) landmarks:
    \begin{equation}
        \mathrm{NME^{\text{vis}}(\%)} = \frac{1}{\sum\limits_{j}\visgd}\sum_{j=1}^{N_p}\visgd \frac{{\|\ground - \bmuF\|}_2}{d} \times 100,
        \label{eq:nme_vis}
    \end{equation}
    If all of the facial landmarks are visible, then this reduces to our previous definition of NME~\eqref{eq:nme}.
    
    We define \nmeboxvis~as the special case of~$\mathrm{NME}^{\text{vis}}$ in which the normalization $d$ is set to the geometric mean of the width and height of the ground-truth bounding box $\bigl(\sqrt{w_\text{bbox} \cdot h_\text{bbox}}\,\bigr),$ as in $\mathrm{NME}_{\text{box}}$~\cite{zafeiriou2017menpo, bulat2017far, chen2019face}. Results for all head poses on \ourdataset~dataset using the metric \nmeboxvis~are shown in Table~\ref{tab:nme_aflw_ours_nme_vis}. We also repeat the \nmebox~numbers from Table~\ref{tab:nme_aflw_ours}. Clearly, the~\nmeboxvis~and \nmebox~numbers are very close for the frontal subsets but are different for half-profile and profile subsets. This is because half-profile and (especially) profile face images have fewer visible landmarks (more self-occluded landmarks), which causes the denominator in~\eqref{eq:nme_vis} to be smaller for these images.

    \subsection{Additional Qualitative Results}
\begin{figure*}[!htb]
            \setlength{\tabcolsep}{0.0cm}
            \begin{tabular}{cccccc}
                \includegraphics[width=0.166\linewidth]{fig/qualitative/300W_test/all_with_all_landmarks/indoor_003.png} &
                \includegraphics[width=0.166\linewidth]{fig/qualitative/300W_test/all_with_all_landmarks/outdoor_198.png} &
                \includegraphics[width=0.166\linewidth]{fig/qualitative/300W_test/all_with_all_landmarks/outdoor_174.png} &
                \includegraphics[width=0.166\linewidth]{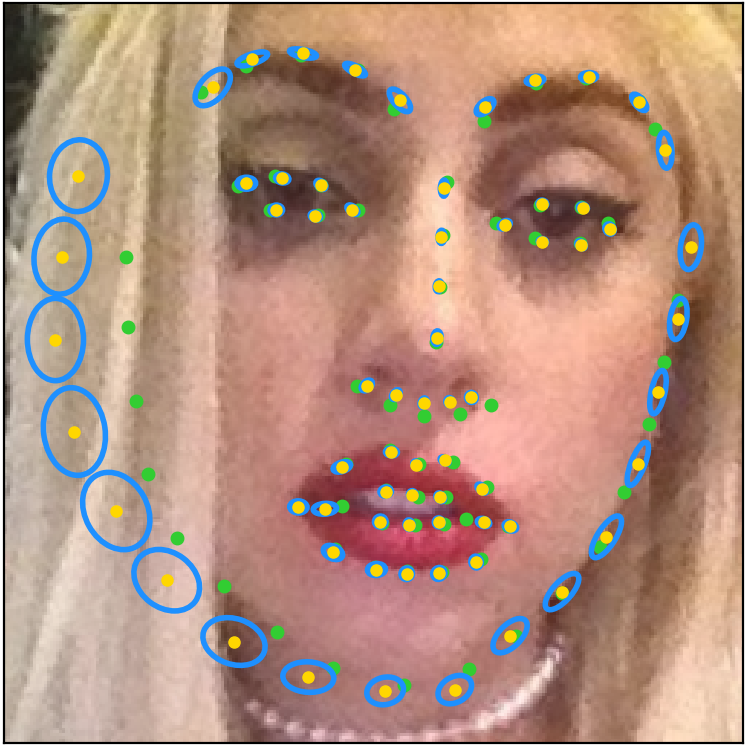} &
                \includegraphics[width=0.166\linewidth]{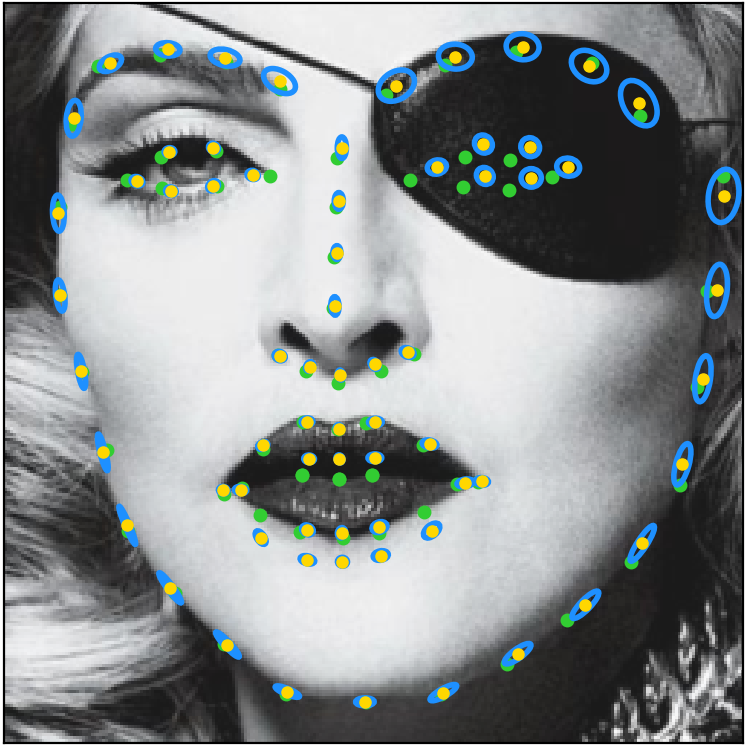} &
                \includegraphics[width=0.166\linewidth]{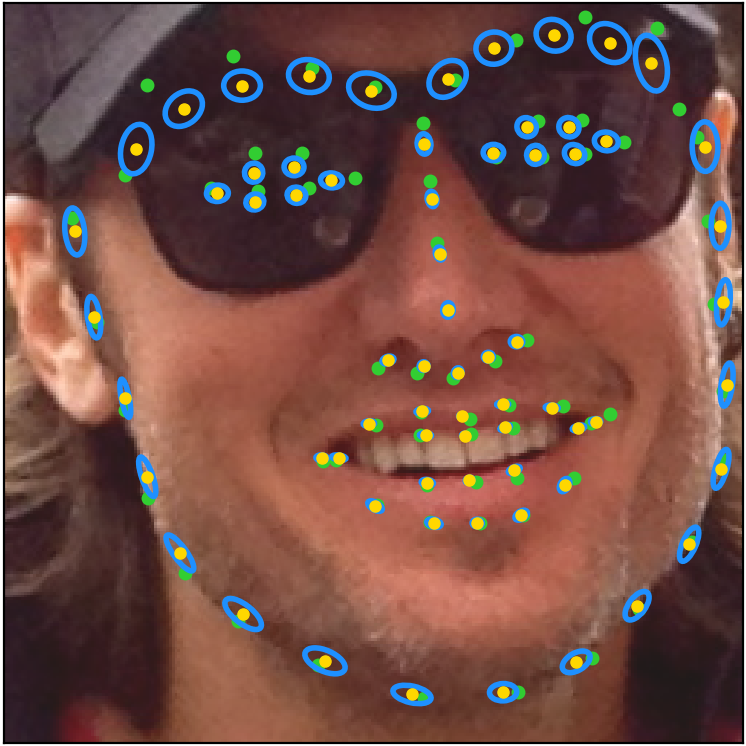} \\
                \includegraphics[width=0.166\linewidth]{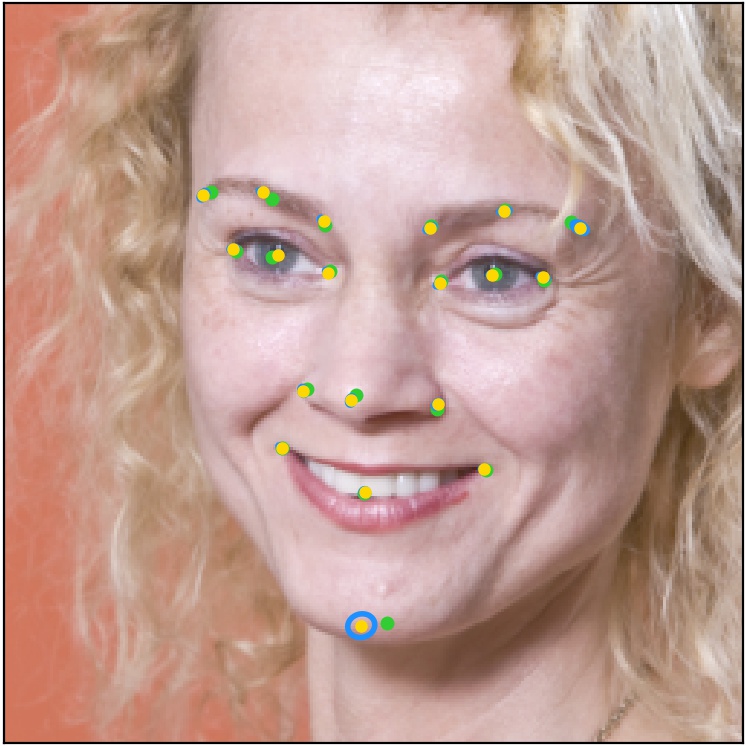} &
                \includegraphics[width=0.166\linewidth]{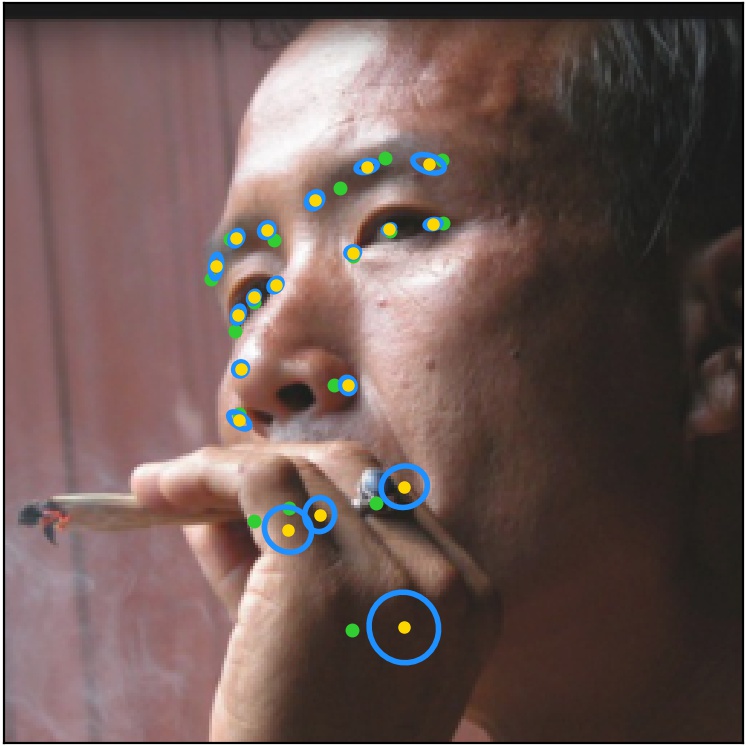} &
                \includegraphics[width=0.166\linewidth]{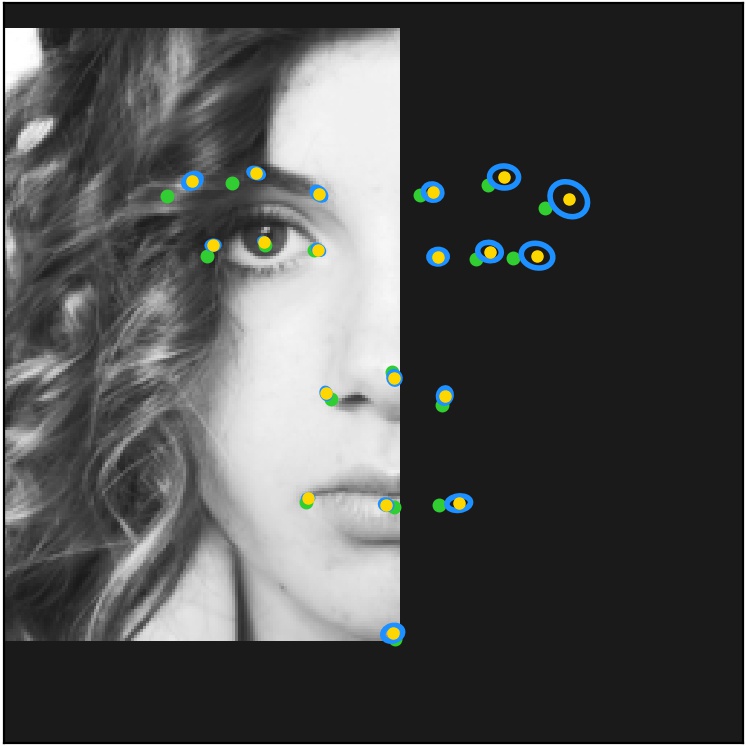} &
                \includegraphics[width=0.166\linewidth]{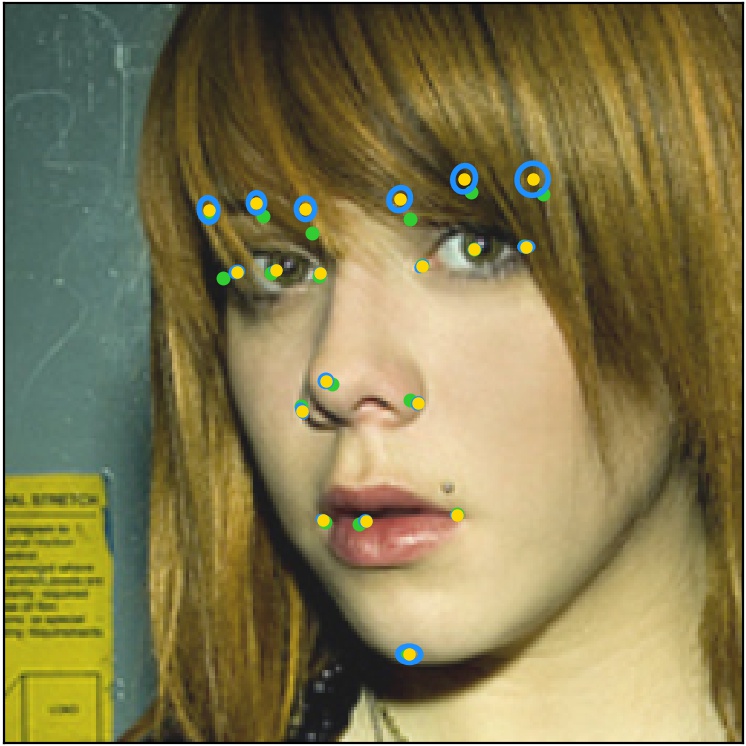} &
                \includegraphics[width=0.166\linewidth]{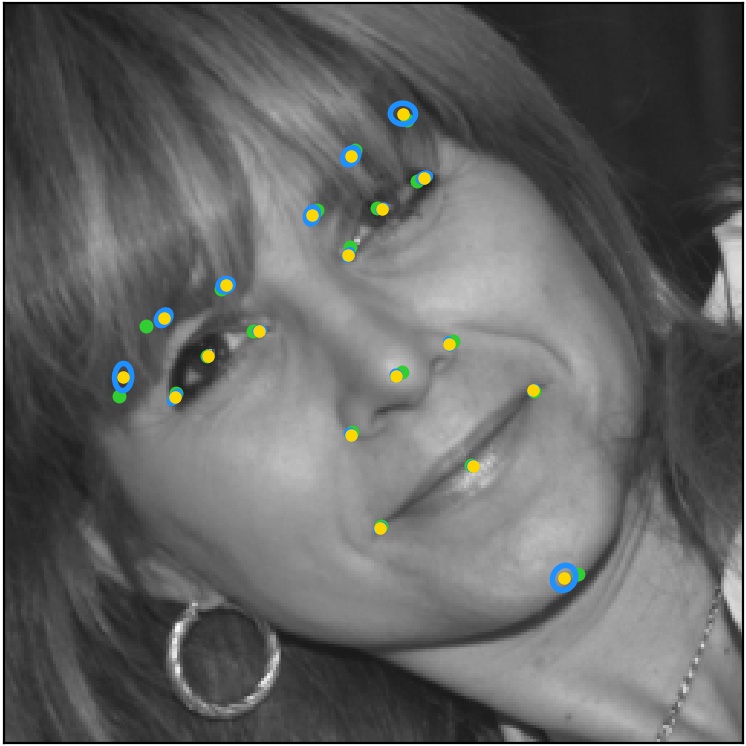} &
                \includegraphics[width=0.166\linewidth]{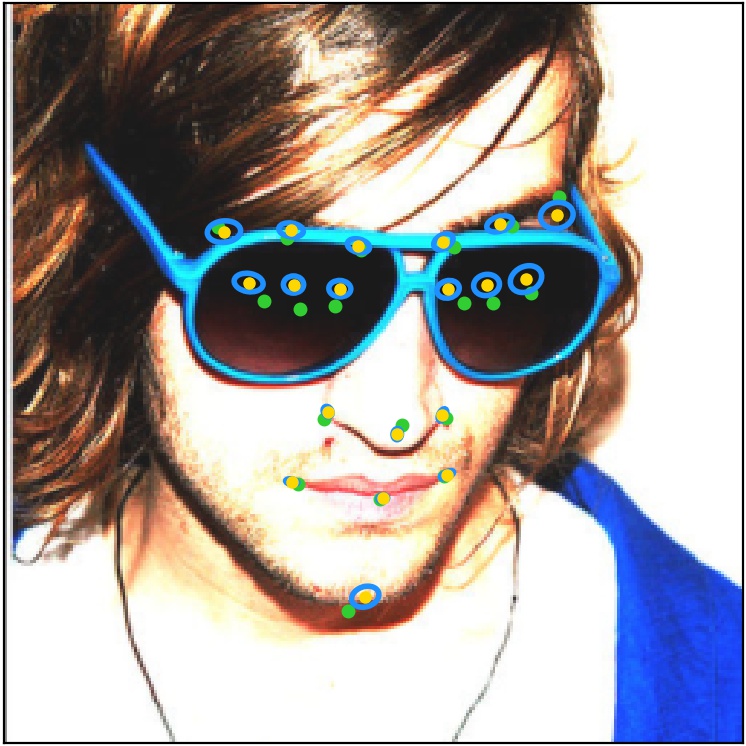}\\
                \includegraphics[width=0.166\linewidth]{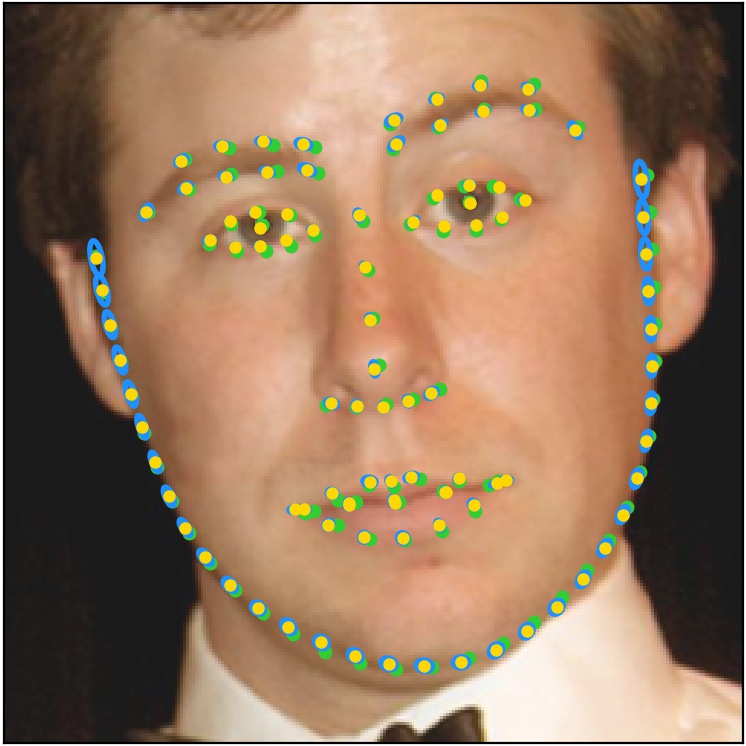} &
                \includegraphics[width=0.166\linewidth]{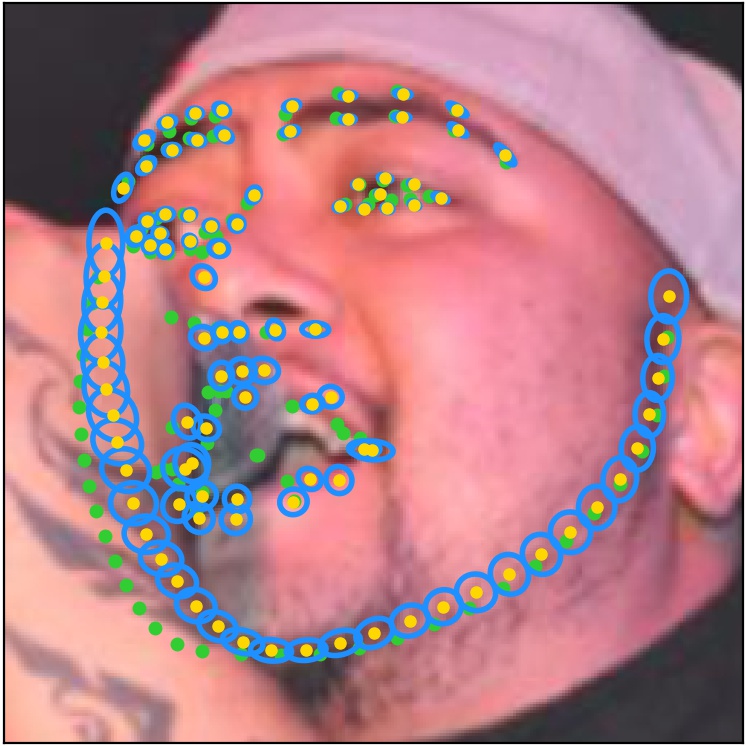} &
                \includegraphics[width=0.166\linewidth]{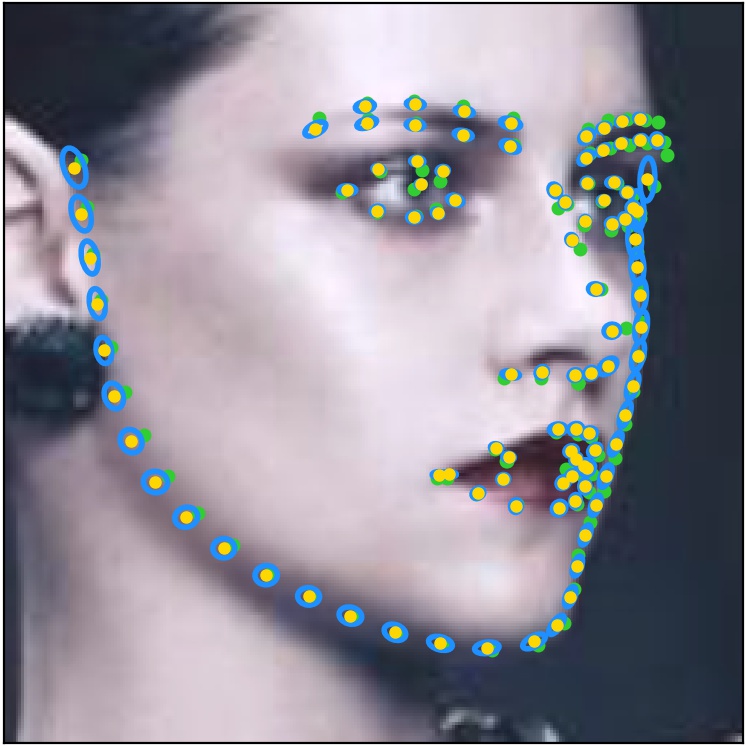} &
                \includegraphics[width=0.166\linewidth]{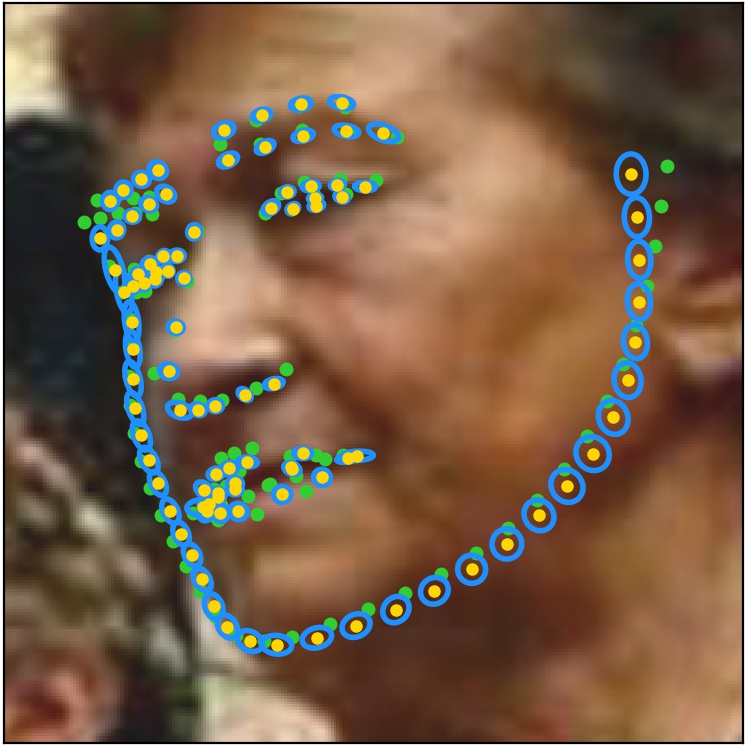} &
                \includegraphics[width=0.166\linewidth]{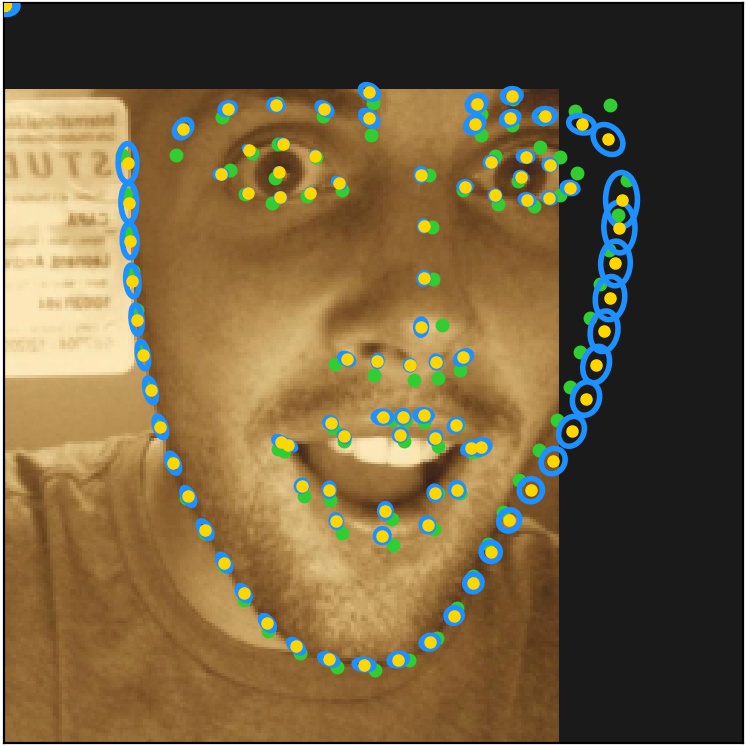} &
                \includegraphics[width=0.166\linewidth]{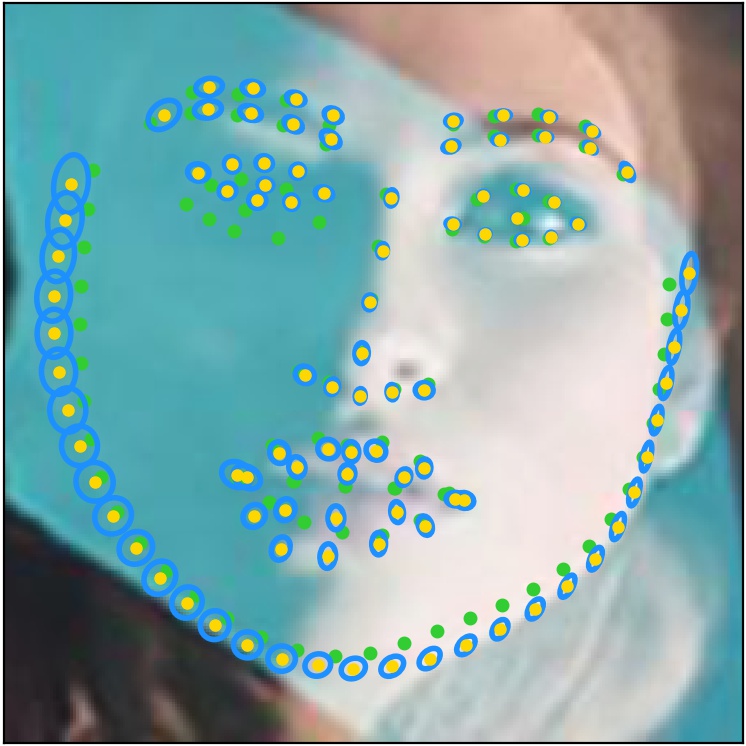}\\
                \includegraphics[width=0.166\linewidth]{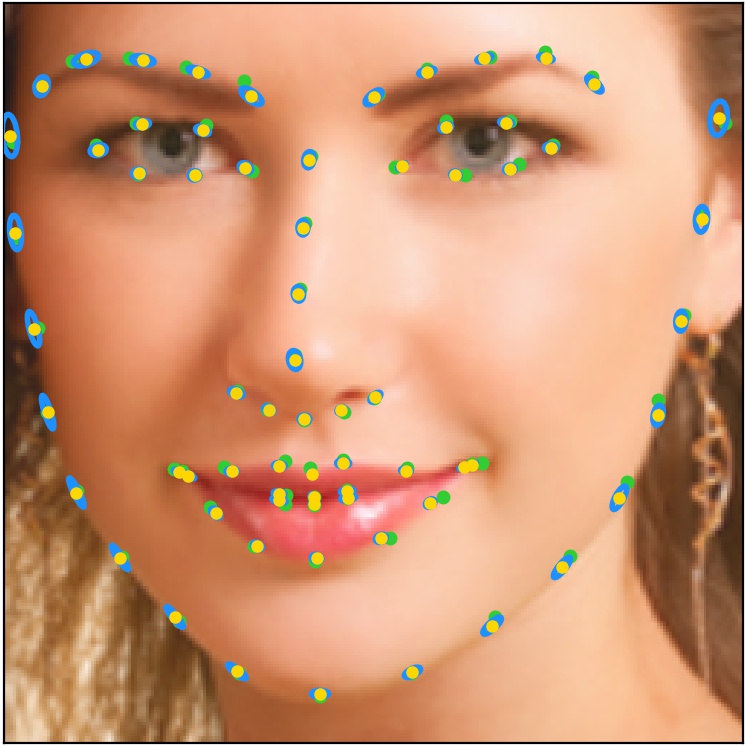} &
                \includegraphics[width=0.166\linewidth]{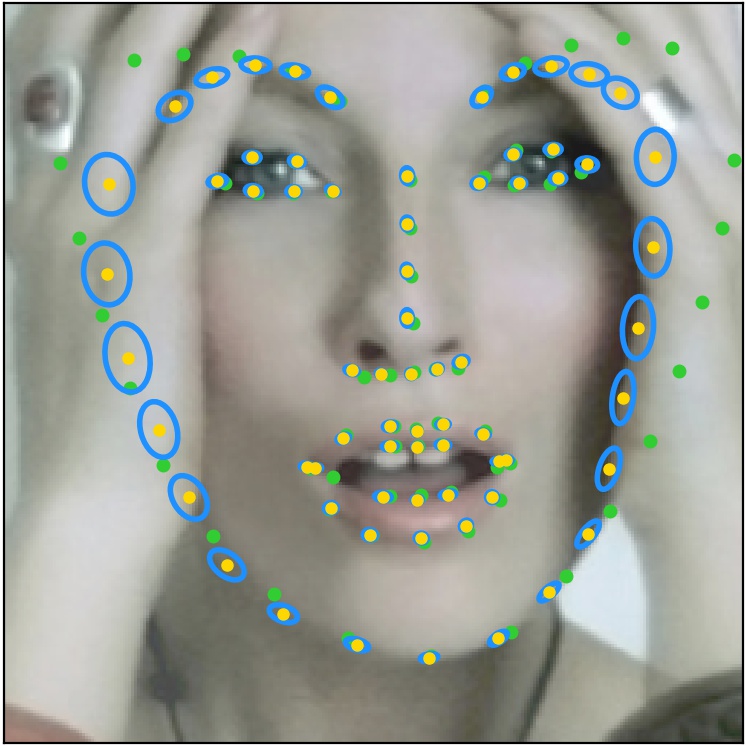}&
                \includegraphics[width=0.166\linewidth]{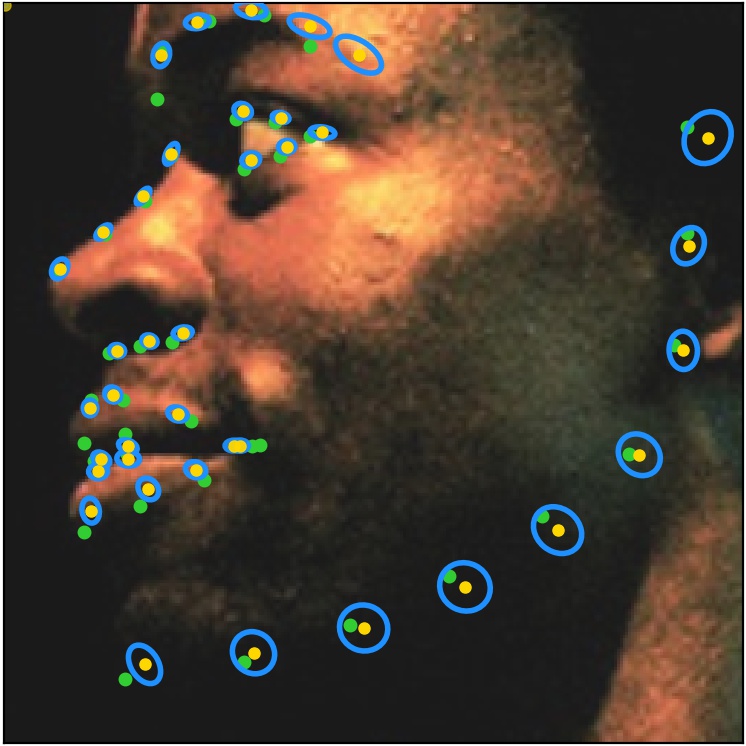} &
                \includegraphics[width=0.166\linewidth]{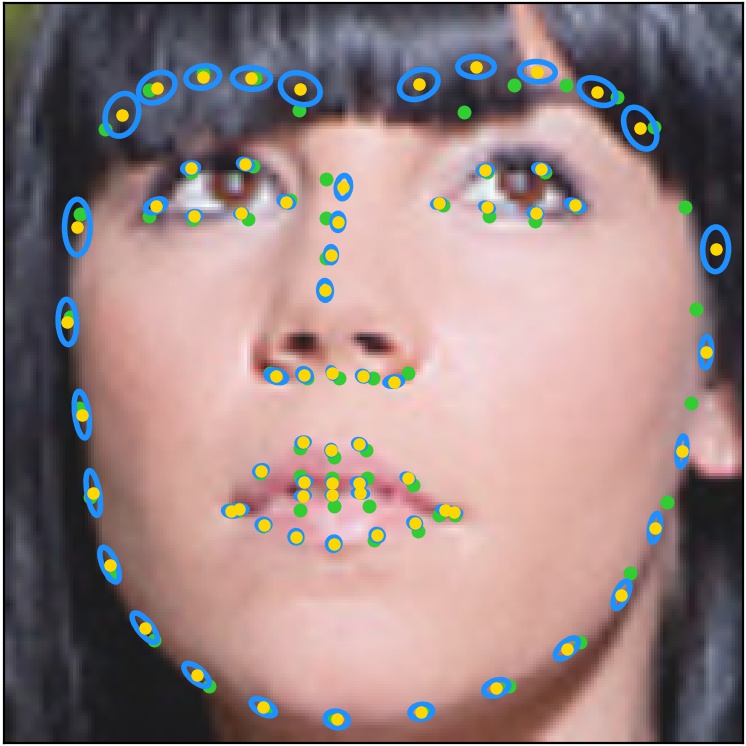} &
                \includegraphics[width=0.166\linewidth]{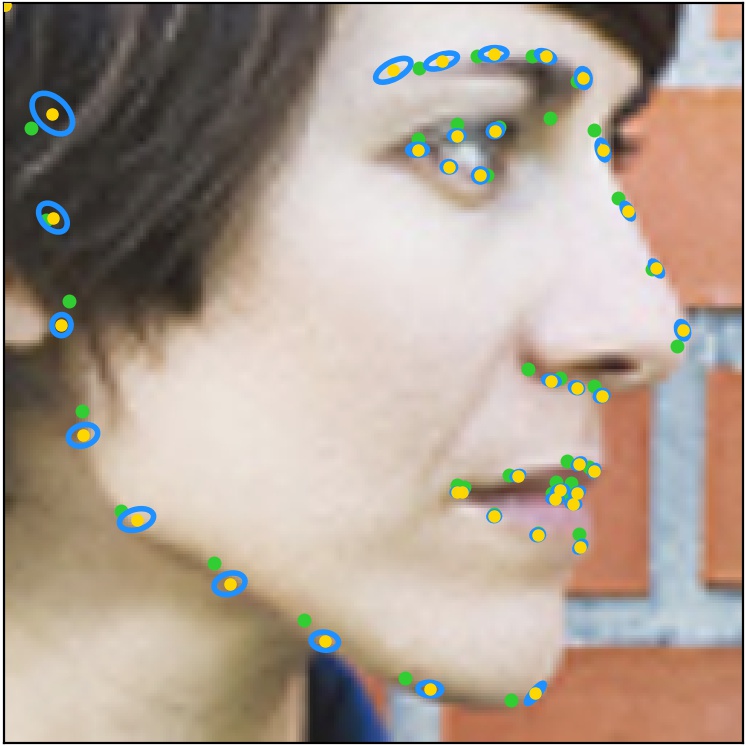} &
                \includegraphics[width=0.166\linewidth]{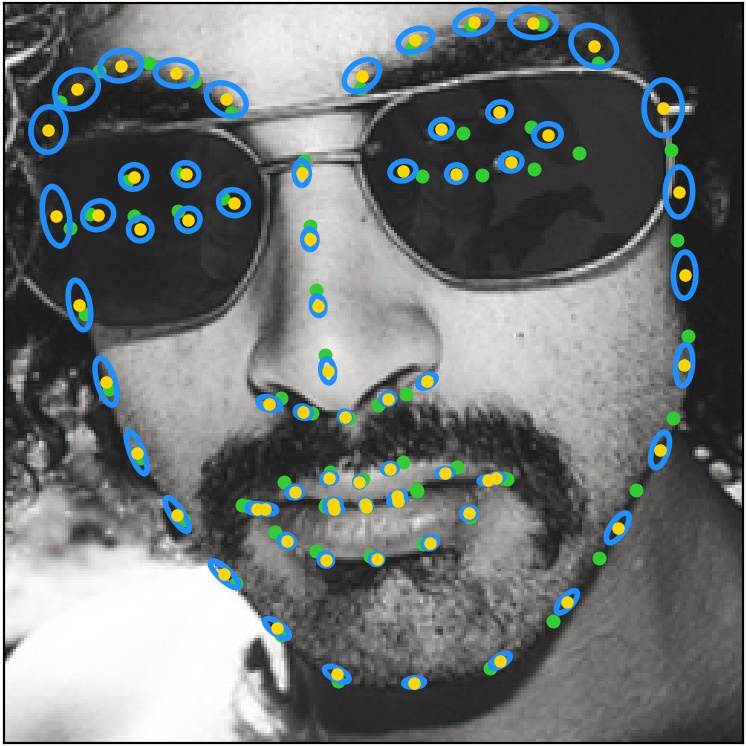} \\
            \end{tabular}
            \caption{Results of our LUVLi face alignment on example face images from four face datasets. {\em Top row:} \threehundredW. {\em Second row:} \aflwNineteen. {\em Third row:} WFLW. {\em Bottom row:}  \ourdataset. Ground-truth (green) and predicted (yellow) landmark locations are shown. The estimated uncertainty of the predicted location of each landmark is shown in blue (Error ellipse for Mahalanobis distance 1). In the \ourdataset~images (bottom row), the predicted visibility of each landmark controls its transparency. In particular, the predicted locations of landmarks with predicted visibility close to zero (such the points on the far side of the profile face in the third image of the bottom row) are 100\% transparent (not shown).} 
            \label{fig:more_qualitative}
        \end{figure*}

        In Figure~\ref{fig:more_qualitative}, we show example results on images from four datasets on which we tested.

    \subsection{Video Demo of LUVLi}
    We include a short demo video of our LUVLi model that was trained on our new \ourdataset~dataset. The video  demonstrates our method's ability to predict landmarks' visibility (\thatIs, whether they are self-occluded) as well as their locations and uncertainty. We take a simple face video of a person turning his head from frontal to profile pose and run our method on each frame independently. Overlaid on each frame of video, we plot each estimated landmark location in yellow, and plot the predicted uncertainty as a blue ellipse. To indicate the predicted visibility of each landmark, we modulate the transparency of the landmark (of the yellow dot and blue ellipse). Landmarks whose predicted visibility is close to 1 are shown as fully opaque, while landmarks whose predicted visibility is close to zero are fully transparent (are not shown). Landmarks with intermediate predicted visibilities are shown  as partially transparent. 
    
    In the video, notice that as the face approaches the profile pose, points on the far edge of the face begin to disappear, because the method correctly predicts that they are not visible (are self-occluded) when the face is in profile pose.

    \subsection{Examples from our \ourdatasetHeading~Dataset}
\begin{figure*}[!htb]
            \begin{tabular}{c@{\hskip 0.05cm}cc@{\hskip 0.05cm}cc@{\hskip 0.05cm}c}
                \includegraphics[width=0.18\linewidth, height=0.20\linewidth]{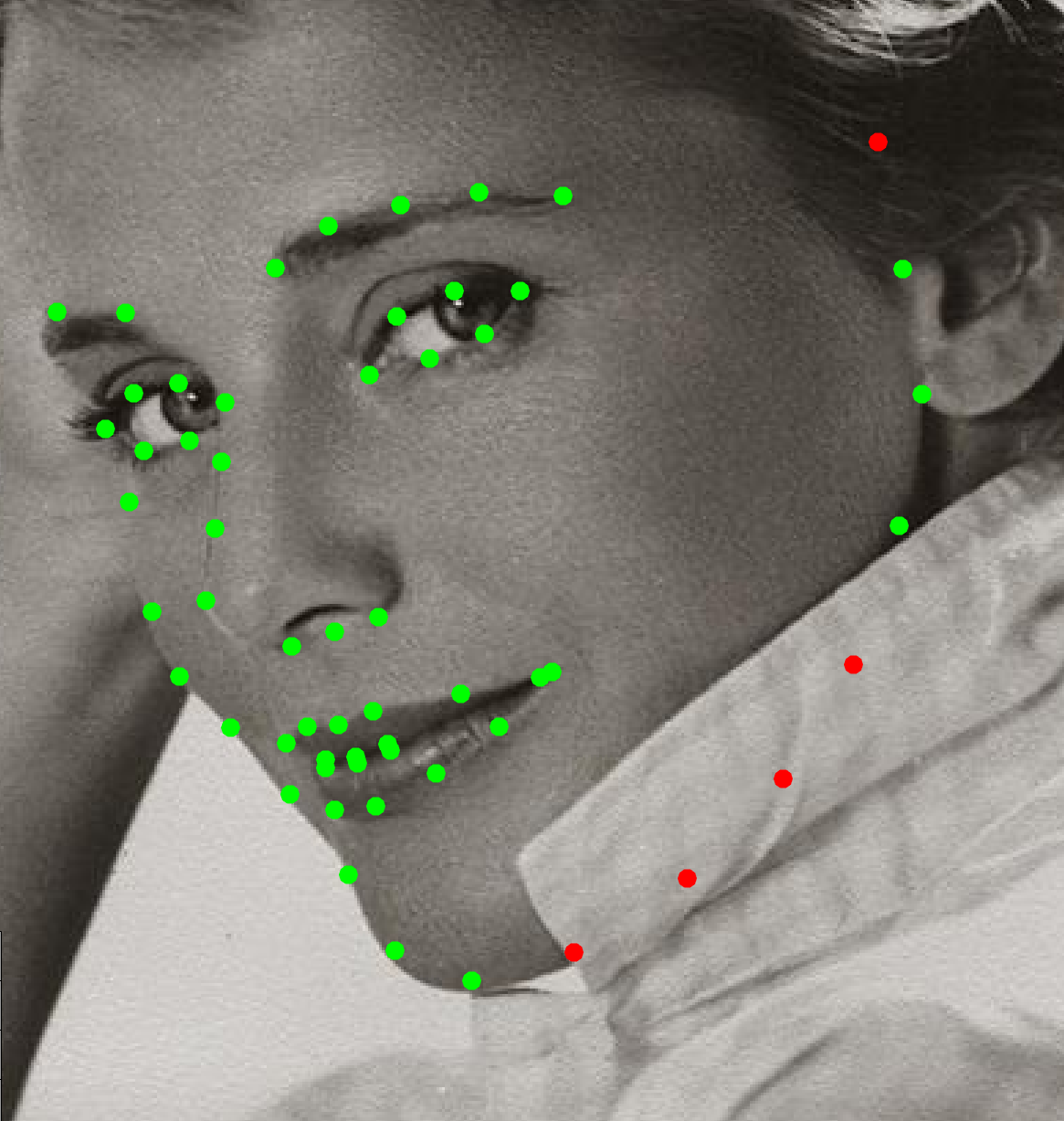} &
                \includegraphics[width=0.15\linewidth, height=0.17\linewidth]{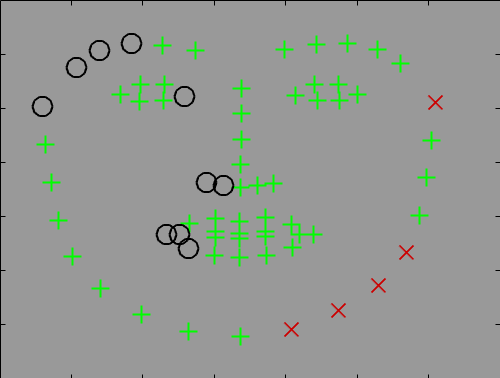} &
                \includegraphics[width=0.18\linewidth, height=0.20\linewidth]{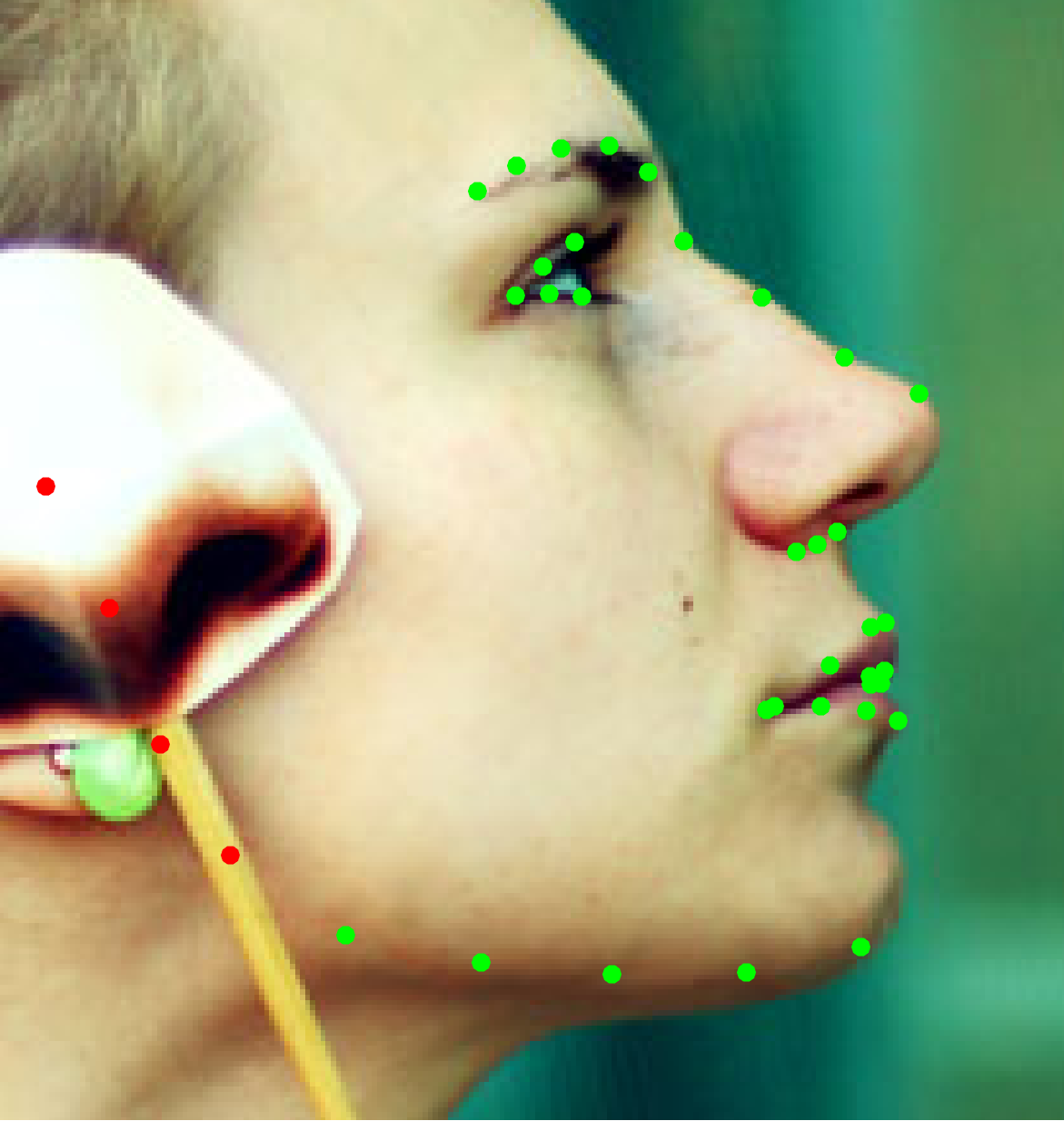} &
                \includegraphics[width=0.15\linewidth, height=0.17\linewidth]{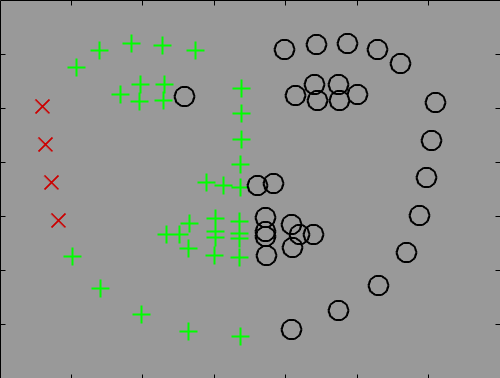} &
                \includegraphics[width=0.18\linewidth, height=0.20\linewidth]{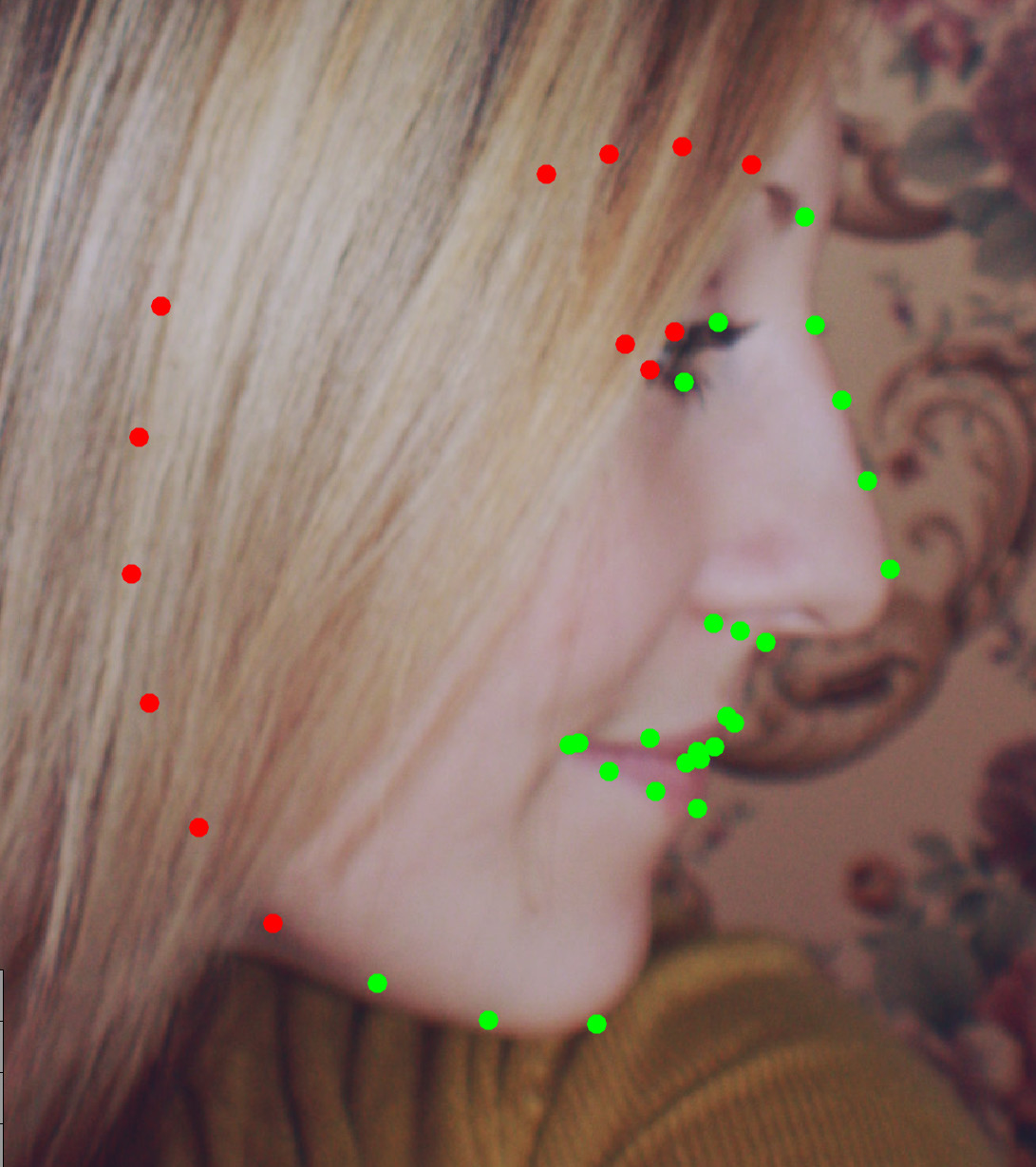} &
                \includegraphics[width=0.15\linewidth, height=0.17\linewidth]{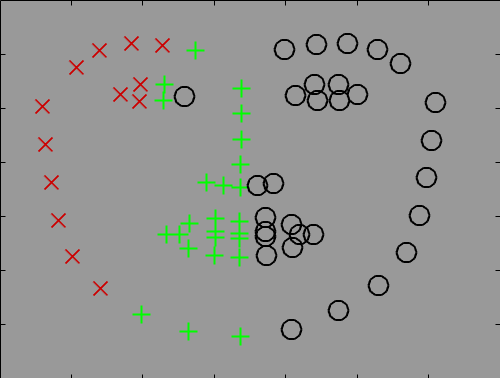} \\
                \includegraphics[width=0.18\linewidth, height=0.20\linewidth]{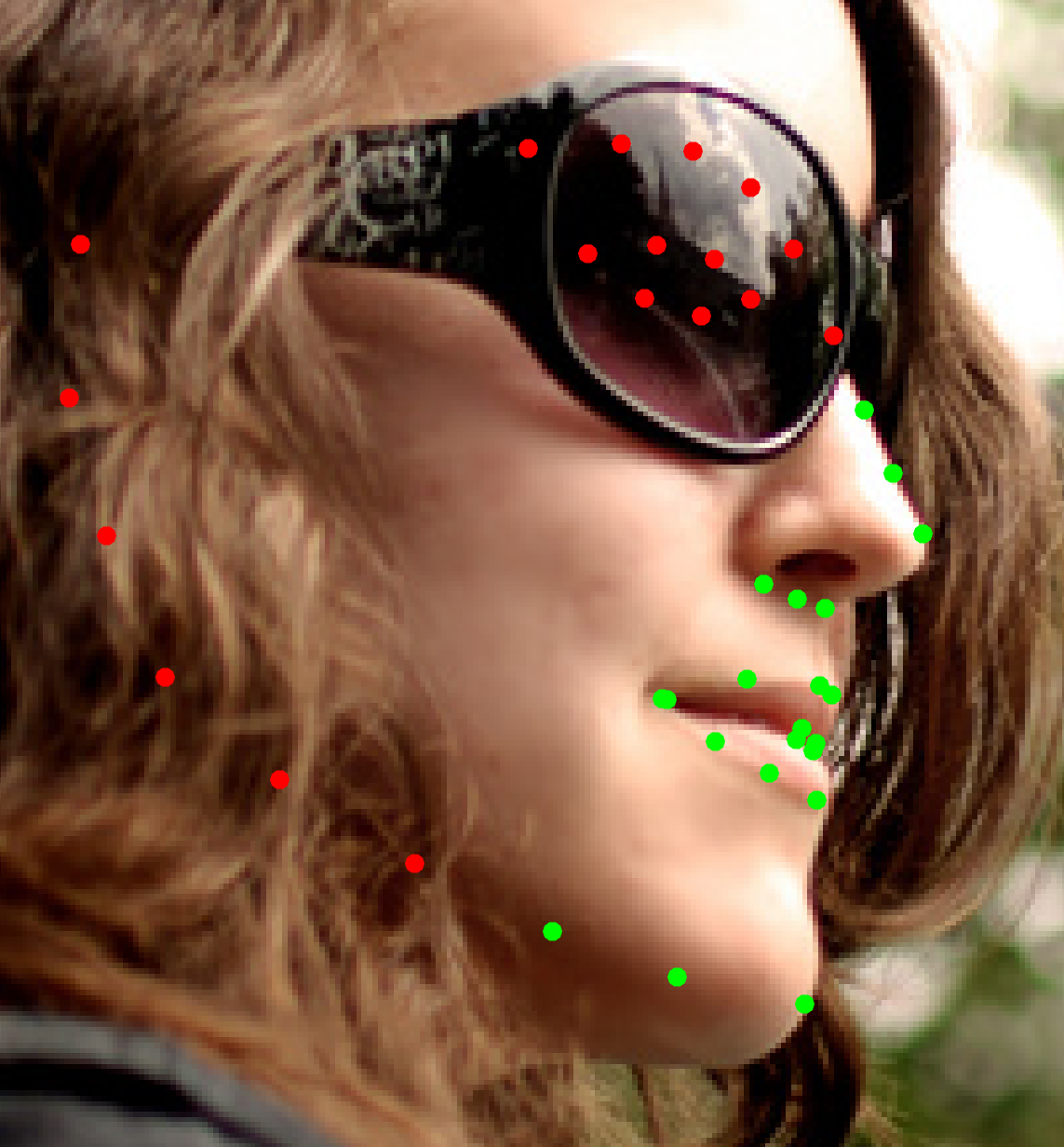} &
                \includegraphics[width=0.15\linewidth, height=0.17\linewidth]{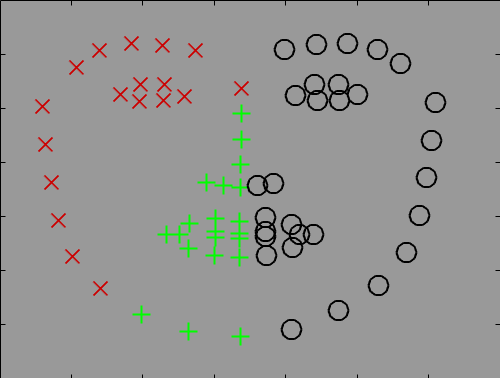} &
                \includegraphics[width=0.18\linewidth, height=0.20\linewidth]{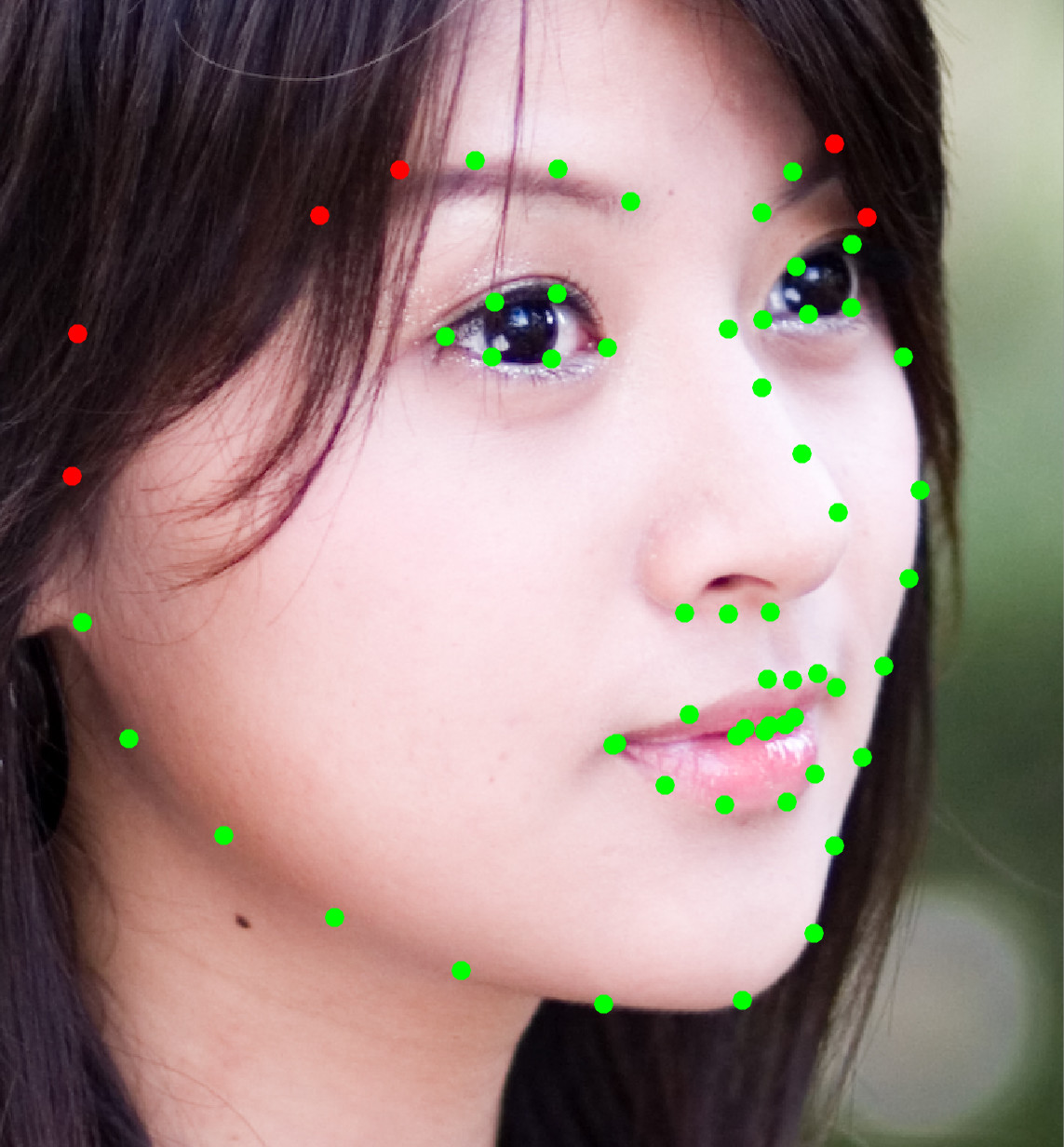} &
                \includegraphics[width=0.15\linewidth, height=0.17\linewidth]{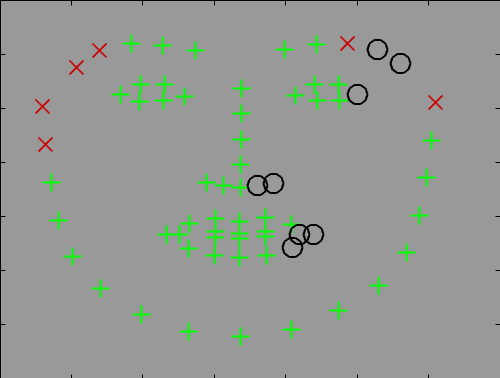} &
                \includegraphics[width=0.18\linewidth, height=0.20\linewidth]{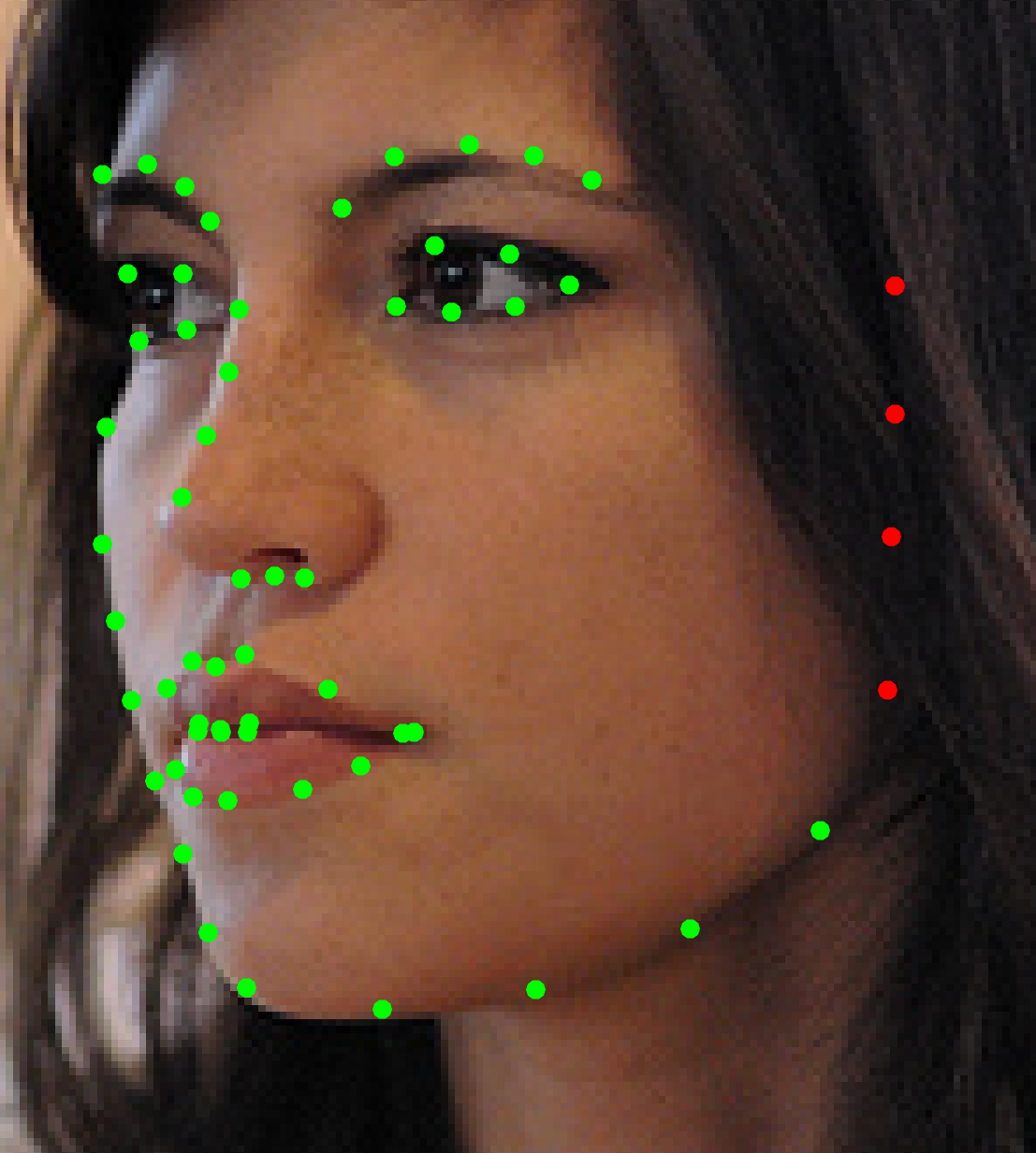} &
                \includegraphics[width=0.15\linewidth, height=0.17\linewidth]{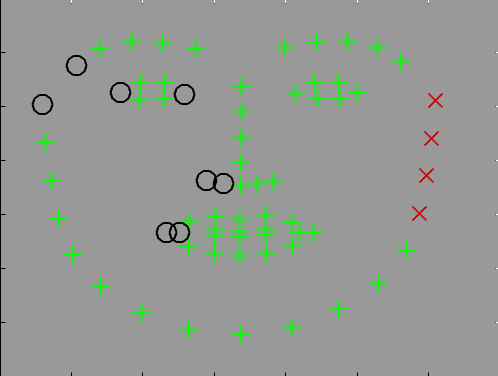} \\
                \includegraphics[width=0.18\linewidth, height=0.20\linewidth]{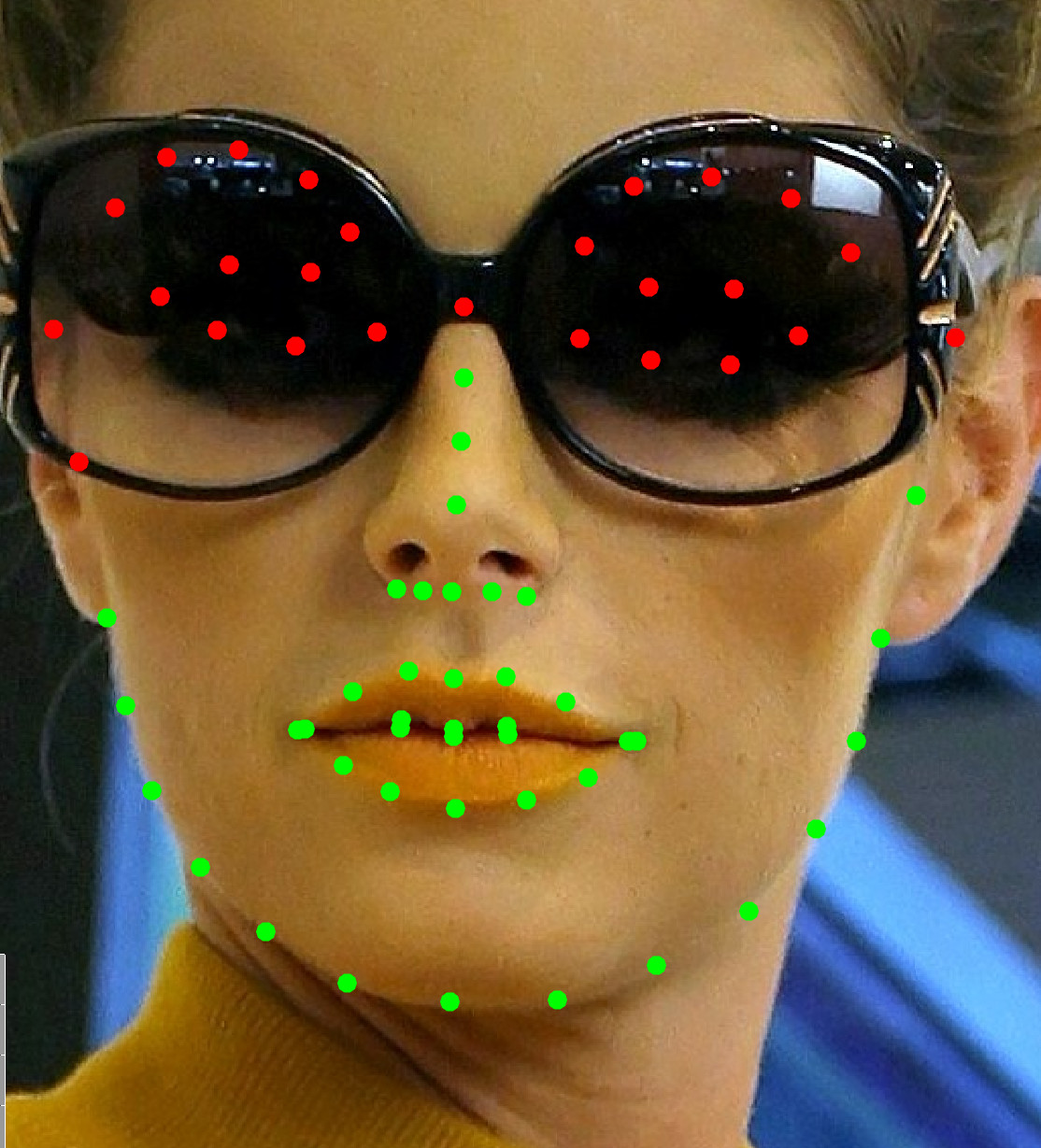} &
                \includegraphics[width=0.15\linewidth, height=0.17\linewidth]{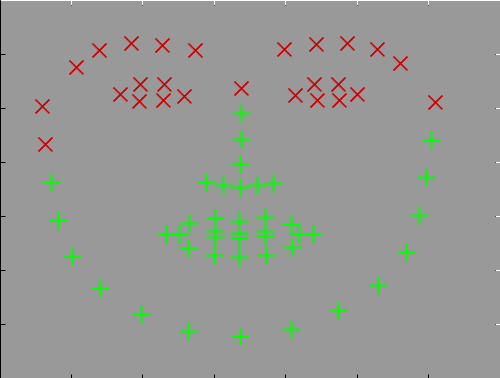} &
                \includegraphics[width=0.18\linewidth, height=0.20\linewidth]{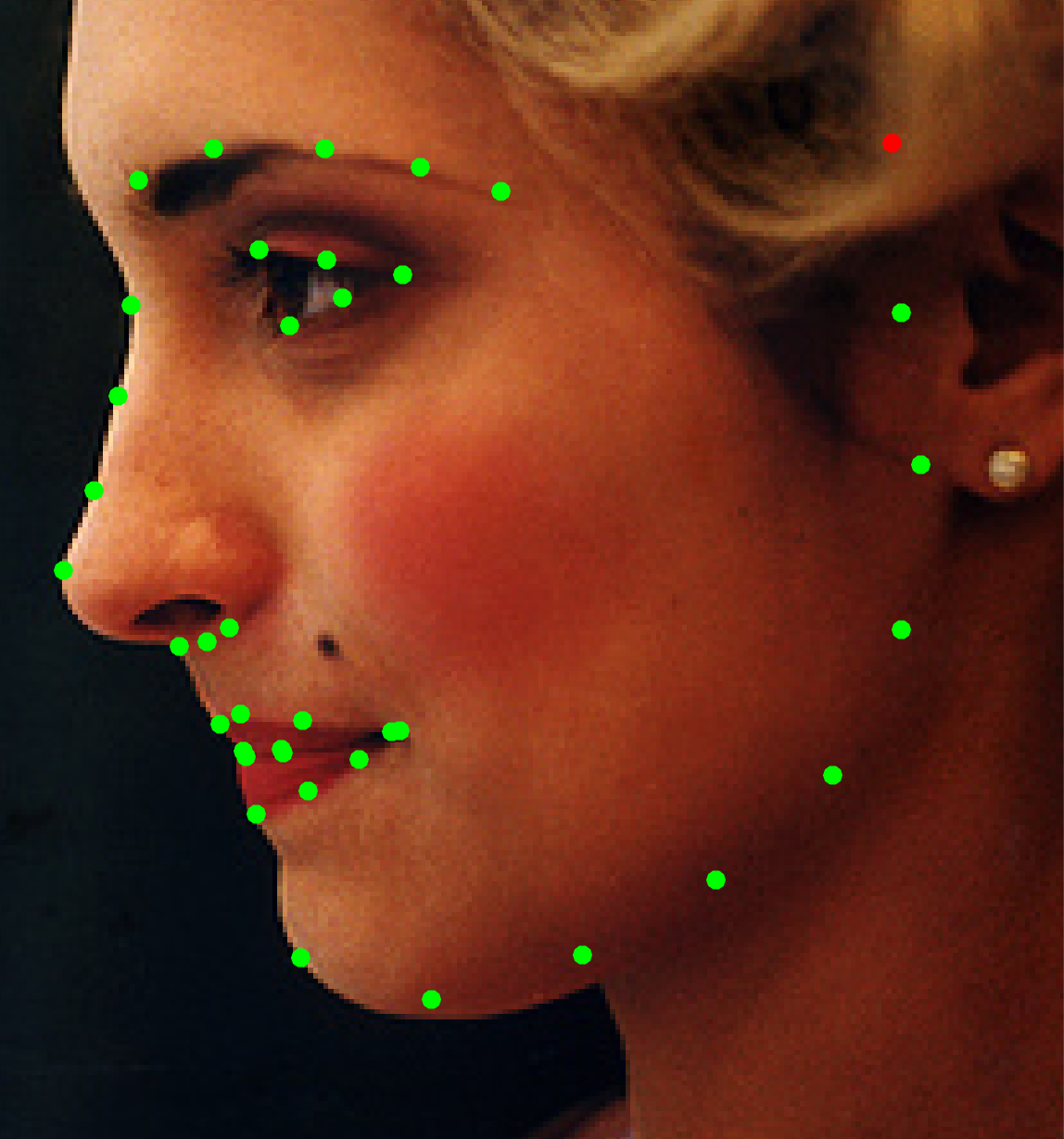} &
                \includegraphics[width=0.15\linewidth, height=0.17\linewidth]{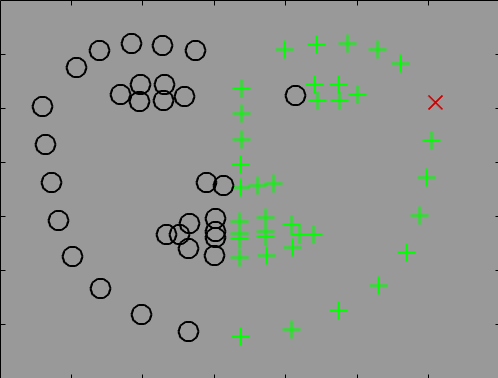} &
                \includegraphics[width=0.18\linewidth, height=0.20\linewidth]{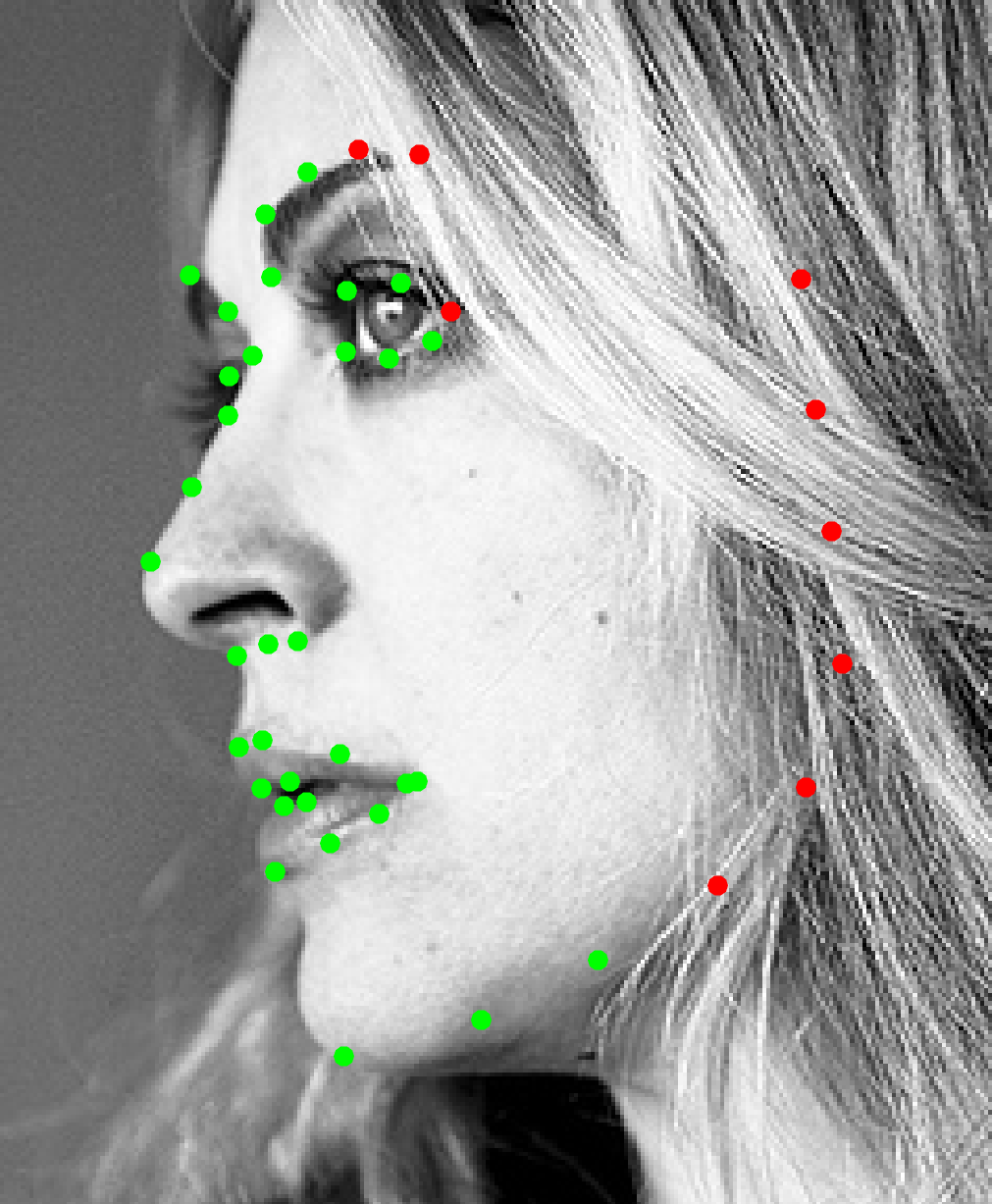} &
                \includegraphics[width=0.15\linewidth, height=0.17\linewidth]{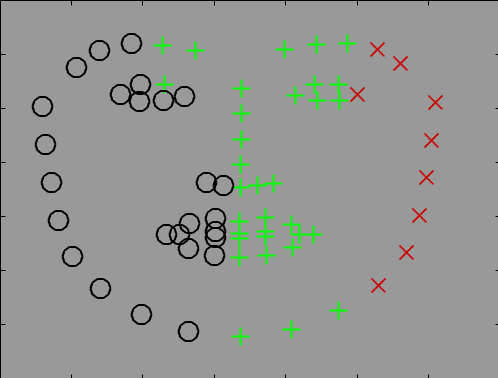} \\
                \includegraphics[width=0.18\linewidth, height=0.20\linewidth]{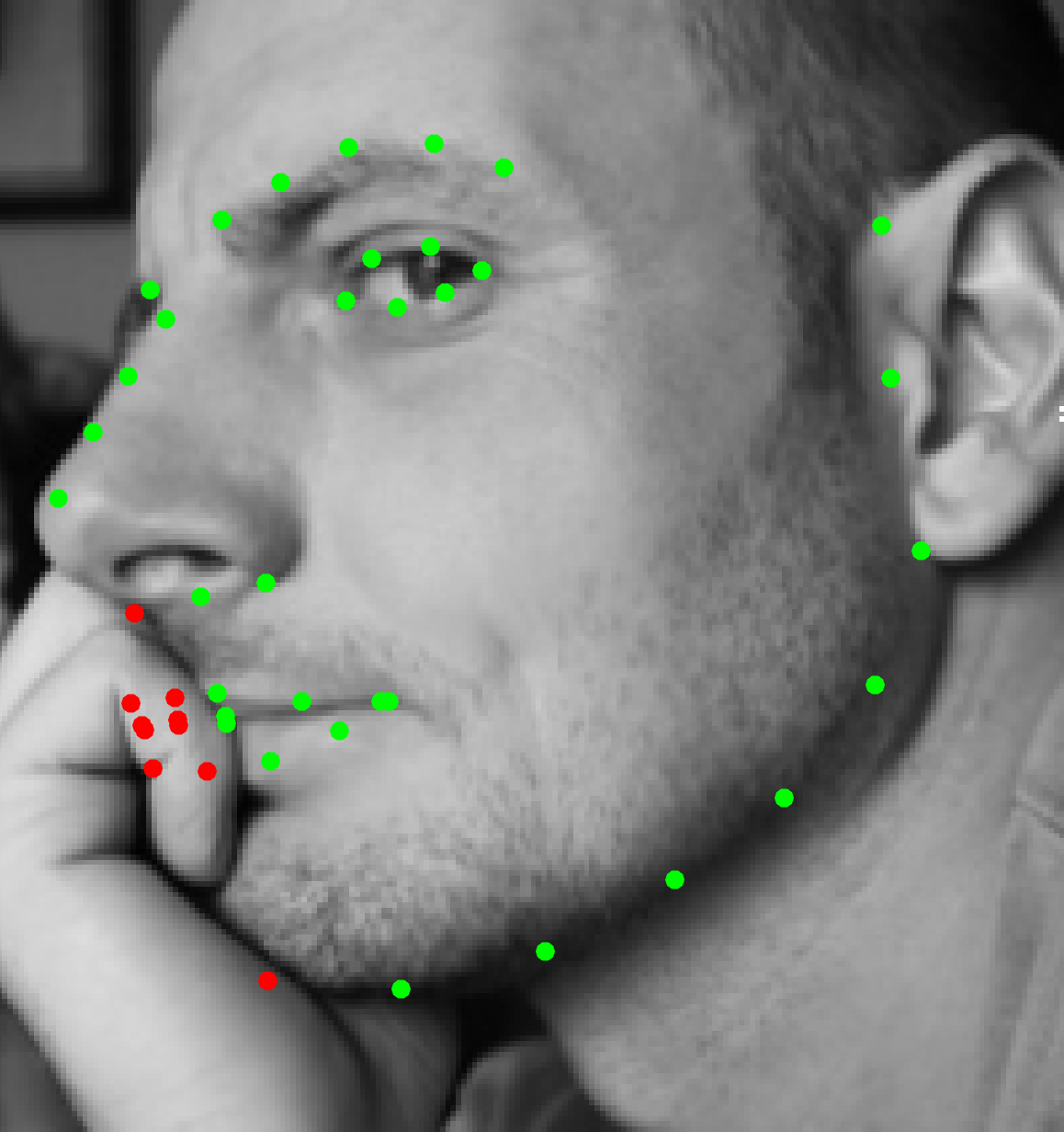} &
                \includegraphics[width=0.15\linewidth, height=0.17\linewidth]{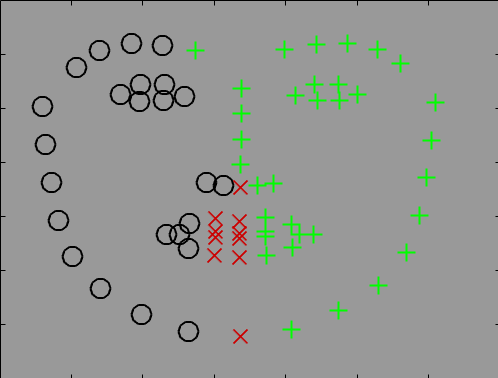} &
                \includegraphics[width=0.18\linewidth, height=0.20\linewidth]{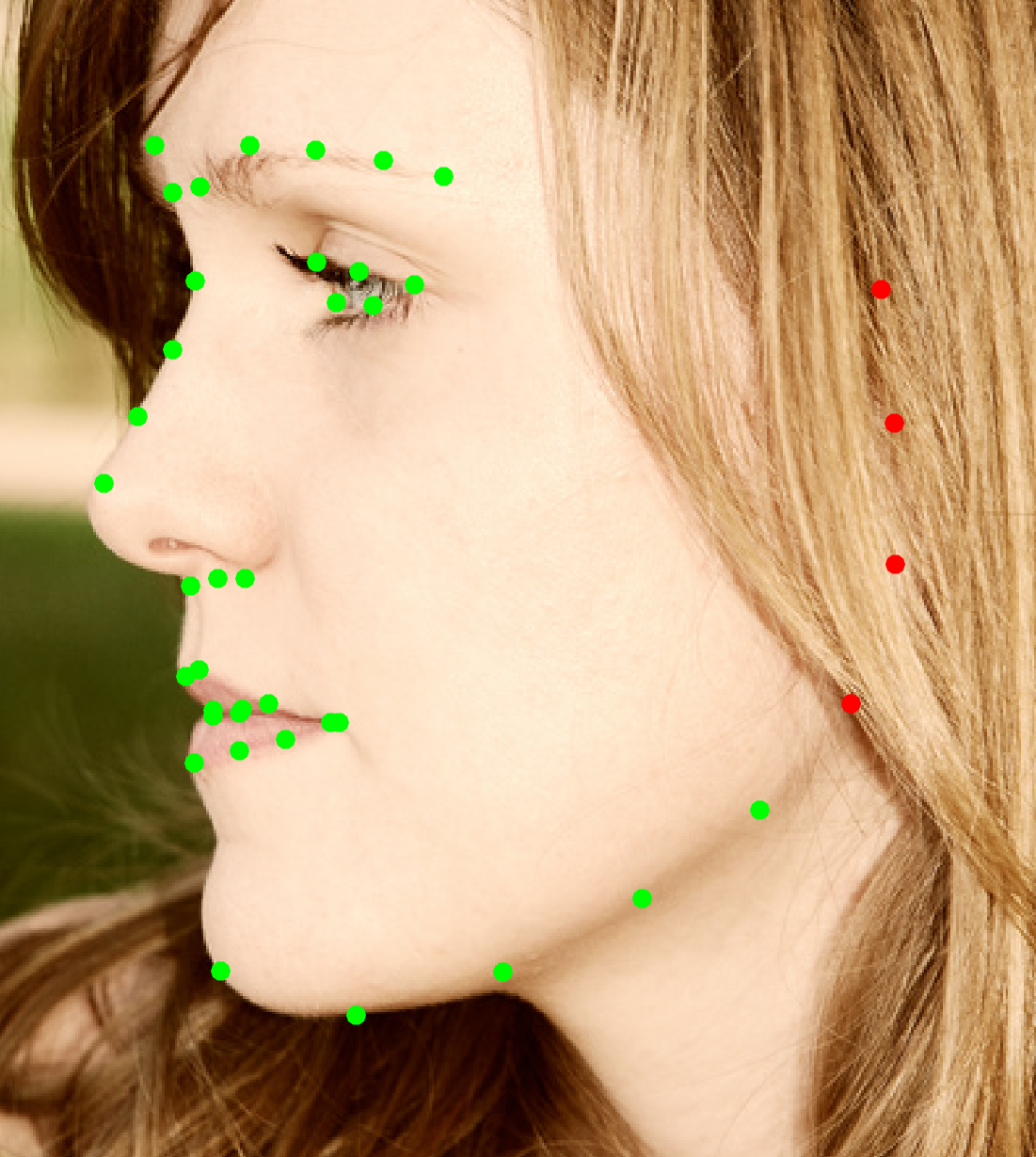} &
                \includegraphics[width=0.15\linewidth, height=0.17\linewidth]{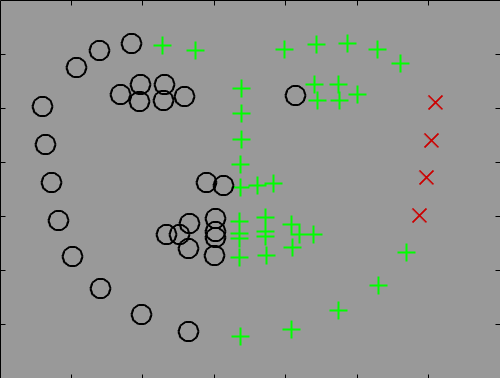} &
                \includegraphics[width=0.18\linewidth, height=0.20\linewidth]{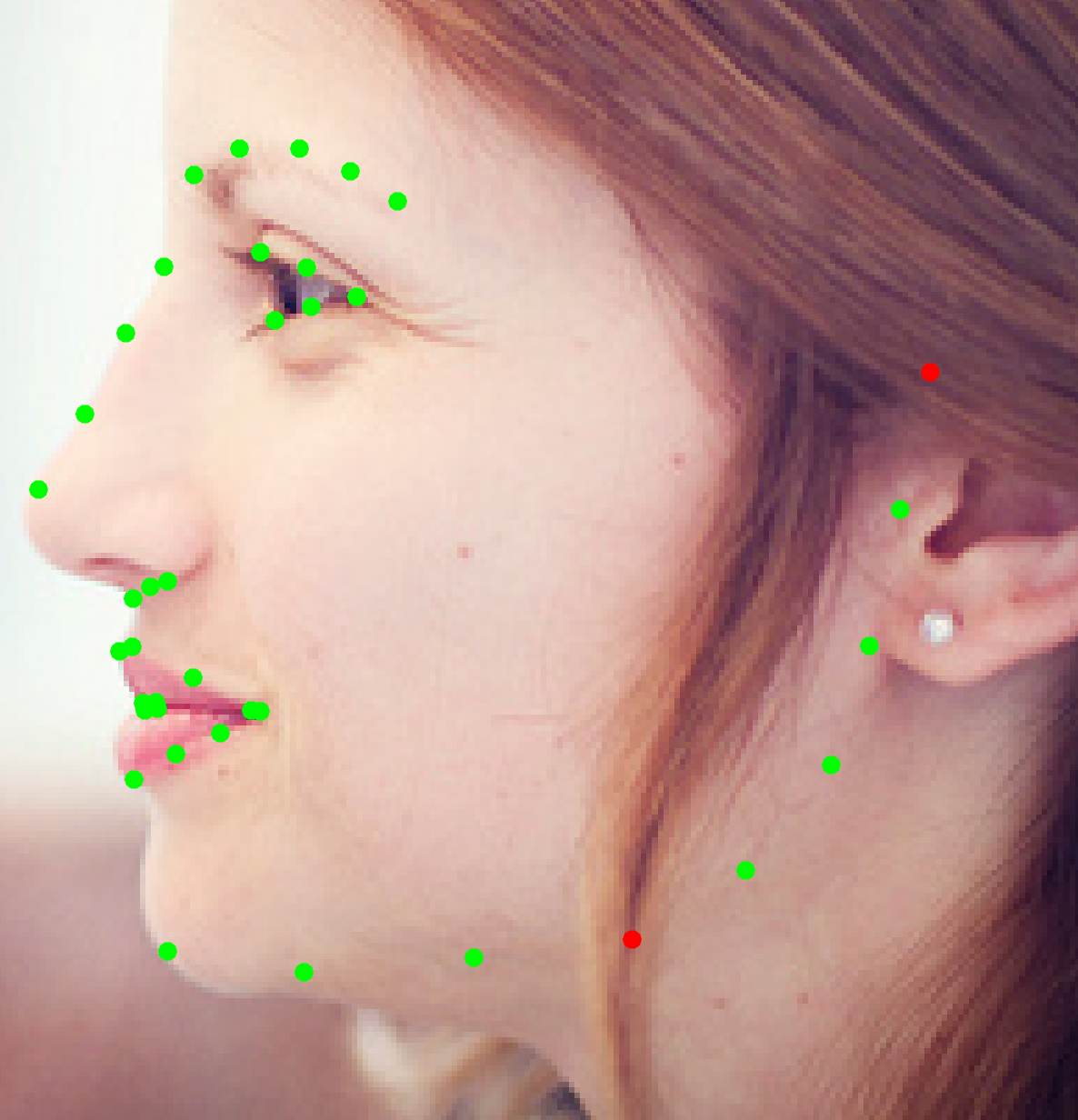} &
                \includegraphics[width=0.15\linewidth, height=0.17\linewidth]{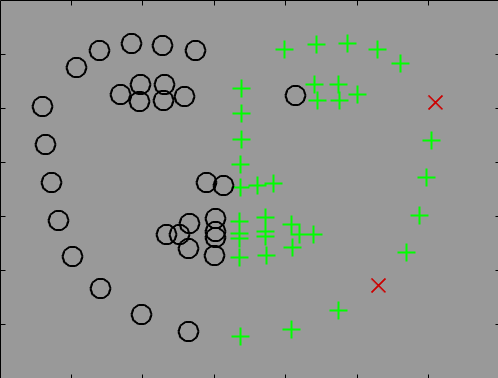} \\
                \includegraphics[width=0.18\linewidth, height=0.20\linewidth]{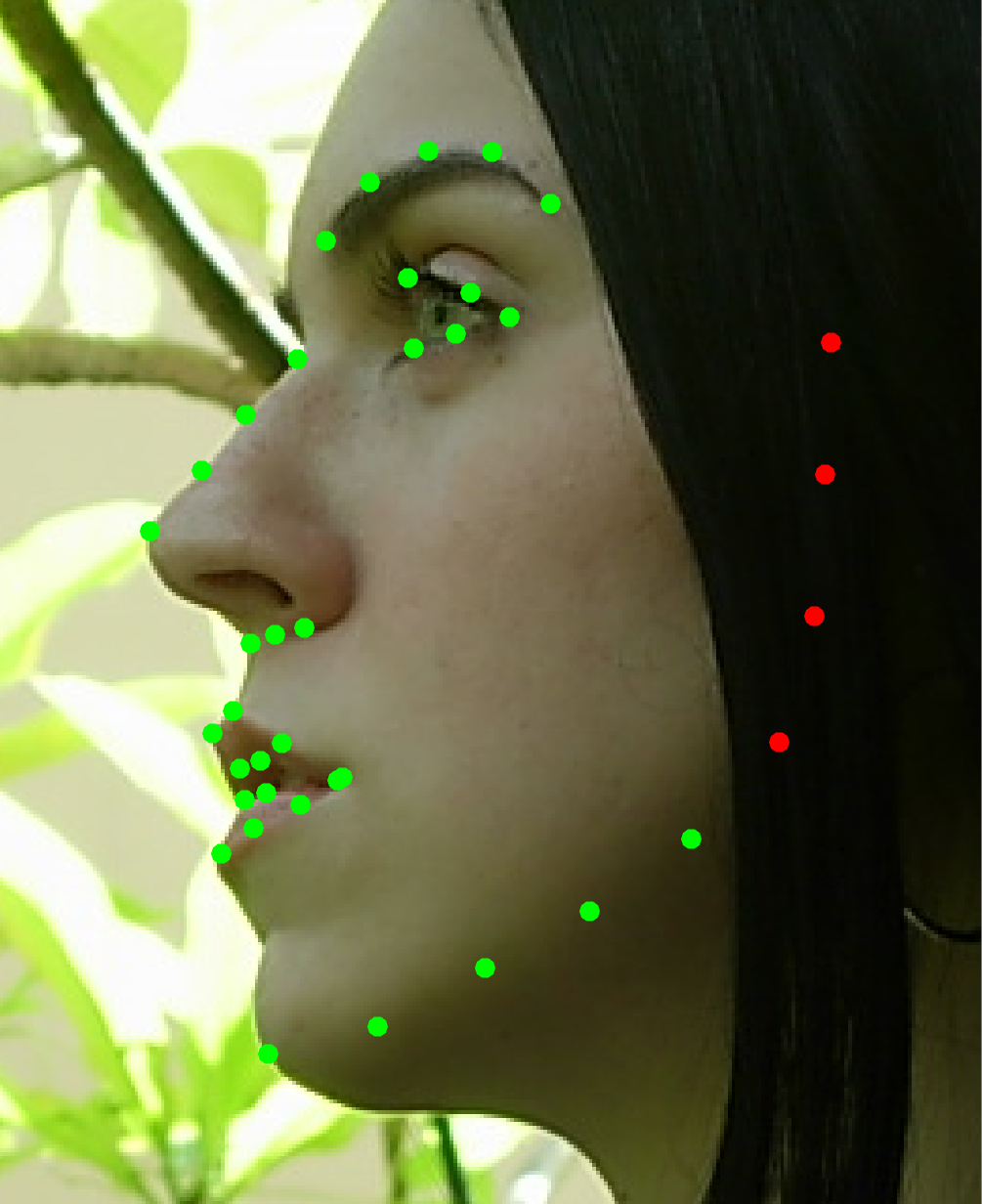} &
                \includegraphics[width=0.15\linewidth, height=0.17\linewidth]{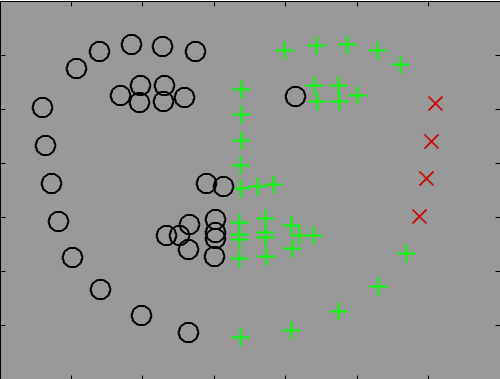} &
                \includegraphics[width=0.18\linewidth, height=0.20\linewidth]{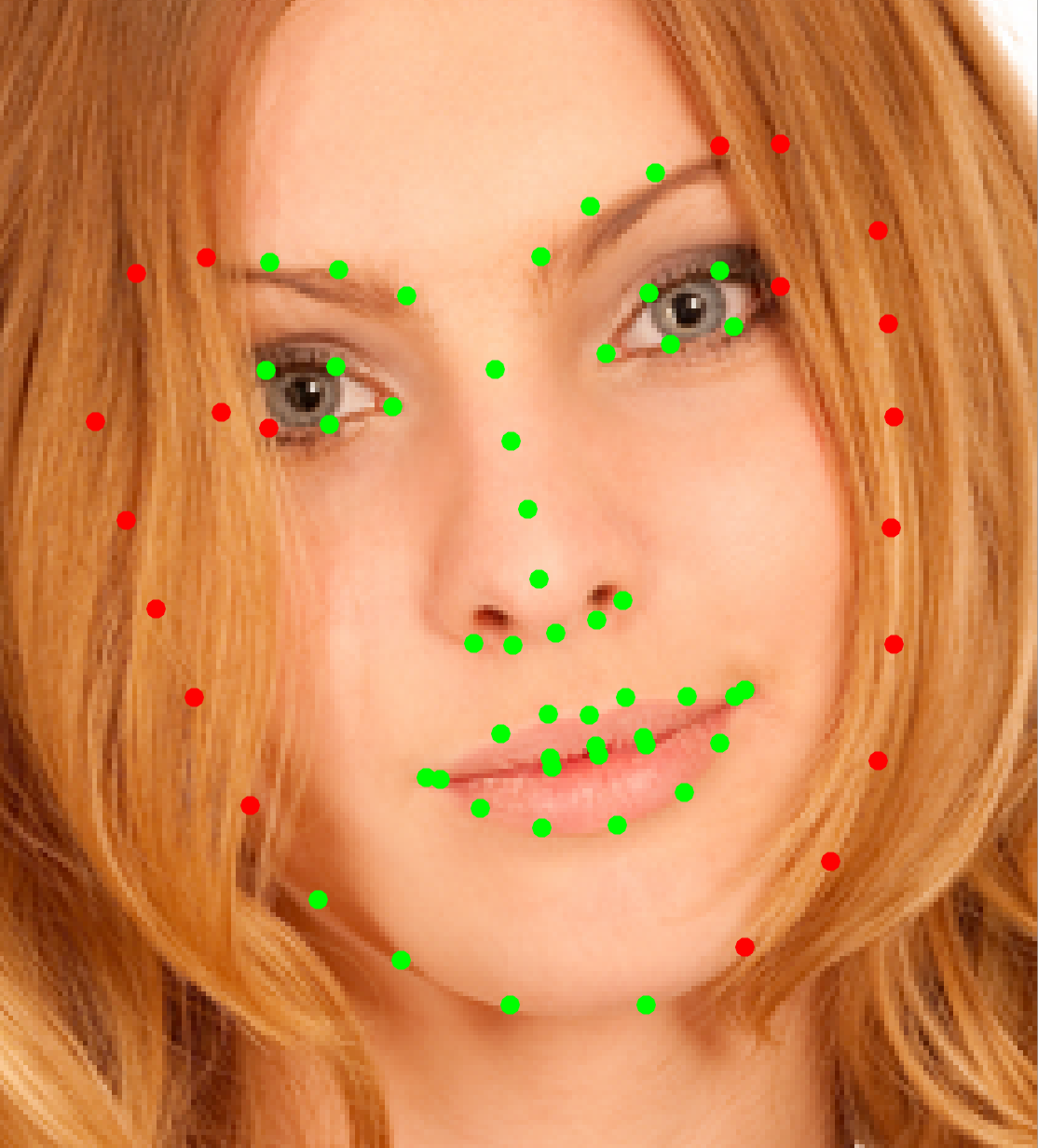} &
                \includegraphics[width=0.15\linewidth, height=0.17\linewidth]{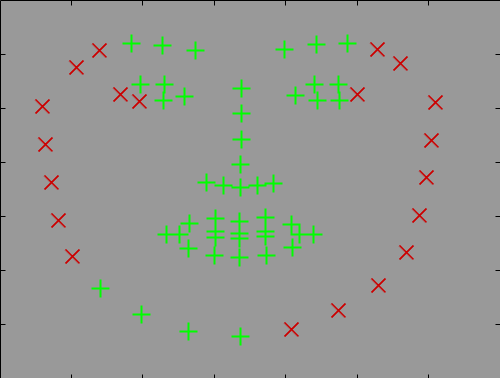} &
                \includegraphics[width=0.18\linewidth, height=0.20\linewidth]{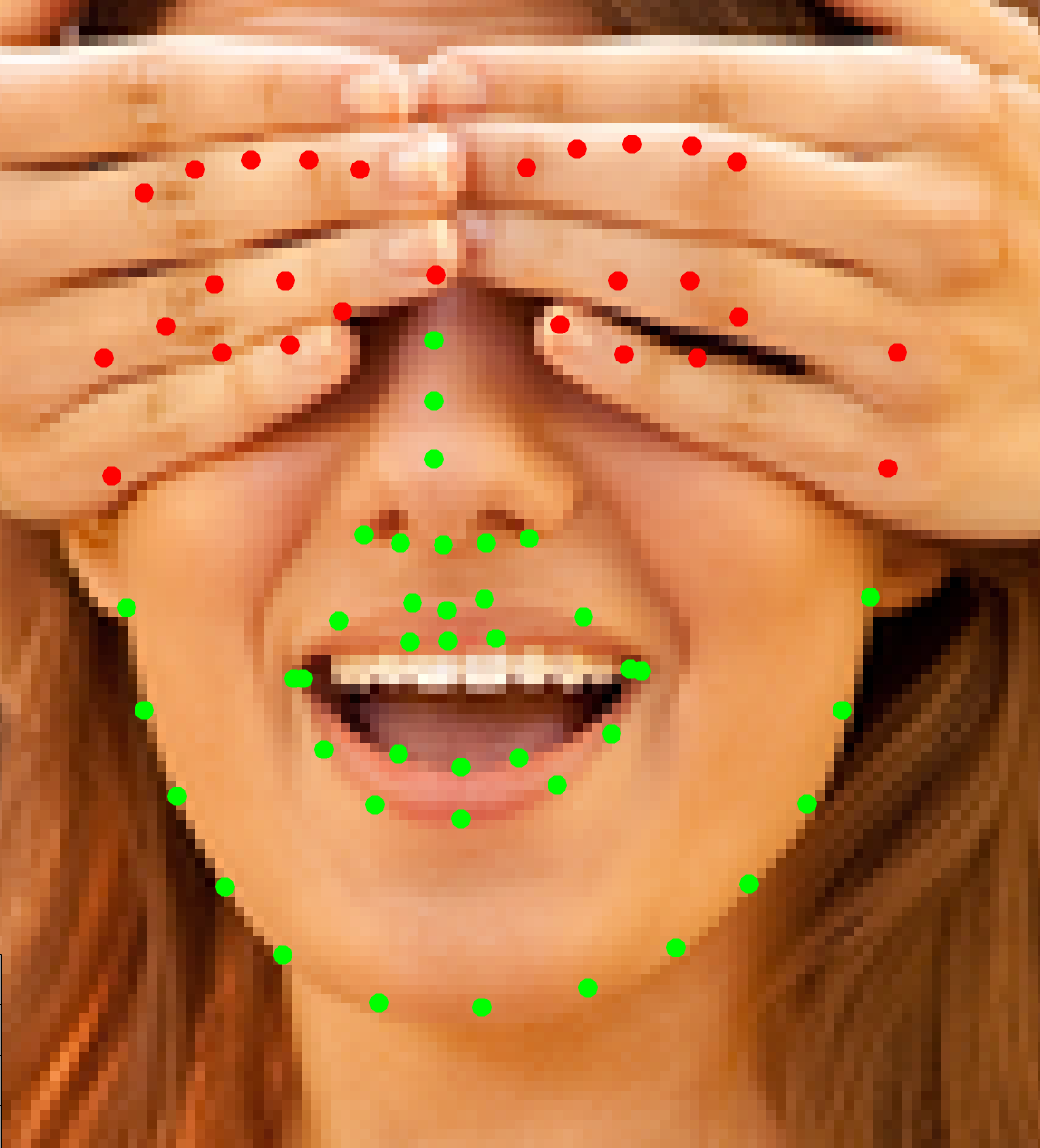}&
                \includegraphics[width=0.15\linewidth, height=0.17\linewidth]{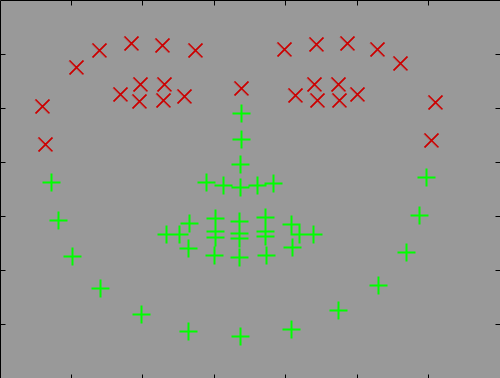} \\
            \end{tabular}
            \caption{Sample images from our \ourdataset~dataset with \textcolor{green}{\textbf{unoccluded}} landmarks shown in green, \textcolor{my_red}{\textbf{externally occluded}} landmarks shown in red, and self-occluded landmarks indicated by black circles in the face schematic on the right of each image.} 
            \label{fig:aflw_ours_samples}
        \end{figure*}

        Figure~\ref{fig:aflw_ours_samples} shows several sample images from our \ourdataset~dataset. The ground-truth labels are overlaid on the images. On each image, unoccluded landmarks are shown in green, externally occluded landmarks are shown in red, and self-occluded landmarks are indicated by black circles in the face schematic to the right of the image.
        
\section*{Acknowledgements}
    We would like to thank Lisha Chen from RPI for providing results from their method and Zhiqiang Tang and Shijie Geng from Rutgers University for providing their pre-trained models on \threehundredW~(Split $1$). We would also like to thank Adrian Bulat and Georgios Tzimiropoulos from the University of Nottingham for detailed discussions on getting bounding boxes for \threehundredW~(Split $2$). We also had very useful discussions with Peng Gao from Chinese University of Hong Kong on the loss functions and Moitreya Chatterjee from University of Illinois Urbana-Champaign. We are also grateful to Maitrey Mehta from the University of Utah who volunteered for the demo. We also acknowledge anonymous reviewers for their feedback that helped in shaping the final manuscript.

\end{document}